\documentclass[]{article}
\usepackage[nonatbib,preprint]{neurips_2023}
\usepackage[utf8]{inputenc} % allow utf-8 input
\usepackage[T1]{fontenc}    % use 8-bit T1 fonts     % hyperlinks
\usepackage{url}            % simple URL typesetting
\usepackage{booktabs}       % professional-quality tables
\usepackage{amsfonts}       % blackboard math symbols
\usepackage{nicefrac}       % compact symbols for 1/2, etc.
\usepackage{microtype}      % microtypography
\usepackage{xcolor}        % colors
\usepackage[colorlinks,citecolor=blue]{hyperref} 
\usepackage[capitalize,nameinlink]{cleveref}
\usepackage[maxbibnames=99]{biblatex}
\usepackage{xspace}
\usepackage{adjustbox}
\usepackage{multirow}
\usepackage{caption}
\usepackage{subcaption}
\usepackage{graphicx}
\usepackage{wrapfig}      % colors
\usepackage[normalem]{ulem}
\usepackage{prftree}
\usepackage{tikz}
\usepackage[capbesideposition=outside]{floatrow}

\title{Instruction Tuned Models are Quick Learners}

\author
{Himanshu Gupta$^{1\diamondsuit}$  \hspace{9pt}  Saurabh Arjun Sawant$^{1\diamondsuit}$ \hspace{9pt}  Swaroop Mishra$^{1\clubsuit}$  \hspace{9pt}  \hspace{9pt}\\ \textbf{Mutsumi Nakamura}$^{1}$ \hspace{9pt} \textbf{Arindam Mitra}$^{2}$ \hspace{9pt} \textbf{Santosh Mashetty}$^{1}$ \hspace{9pt} \textbf{Chitta Baral}$^{1}$\\
\small{$^{1}$Arizona State University \hspace{9pt} $^2$Microsoft Research}\\
\tt\small {\{hgupta35, ssawan13, srmishr1, mutsumi, cbaral\}}@asu.edu
}

\begin{document}

\maketitle

\begin{abstract}
Instruction tuning of language models has demonstrated the ability to enhance model generalization to unseen tasks via in-context learning using a few examples.
However, typical supervised learning still requires a plethora of downstream training data for finetuning.
Often in real-world situations, there is a scarcity of data available for finetuning, falling somewhere between few shot inference and fully supervised finetuning.
In this work, we demonstrate the sample efficiency of instruction tuned models over various tasks by estimating the minimal downstream training data required by them to perform transfer learning and match the performance of state-of-the-art (SOTA) supervised models.
We conduct experiments on 119 tasks from Super Natural Instructions (SuperNI) in both the single task learning (STL) and multi task learning (MTL) settings.
Our findings reveal that, in the STL setting, instruction tuned models equipped with 25\% of the downstream train data surpass the SOTA performance on the downstream tasks.
In the MTL setting, an instruction tuned model trained on only 6\% of downstream training data achieve SOTA, while using 100\% of the training data results in a 3.69\% points improvement (ROUGE-L 74.68) over the previous SOTA.
We conduct an analysis on T5 vs Tk-Instruct by developing several baselines to demonstrate that instruction tuning aids in increasing both sample efficiency and transfer learning.
% We also find that there is an increase of 4\% in both settings if pre-finetuning is done with instructions.
Additionally, we observe a consistent $\sim4\%$ performance increase in both settings when pre-finetuning is performed with instructions.
Finally, we conduct a categorical study and find that contrary to previous results, tasks in the question rewriting and title generation categories suffer from instruction tuning.\footnote{
$\diamondsuit$ Co-first authors $\clubsuit$ Currently in Google Brain \\
Baseline approach, data splits and scripts are freely available at \url{https://github.com/srsawant34/efficient_instruction_learning}}
\end{abstract}

\section{Introduction}

% \begin{figure*}[t!]
% 	\centering
% 	\includegraphics[width= \linewidth, height= 8 cm]{Pictures/new_teaser_2.png}
% 	\caption{ The Figure demonstrates the difference between the Few-Shot inference (without any finetuning), conventional fully supervised finetuning and our proposed analysis. 
%  Conventional few shot inference using Tk-Instruct directly evaluates unseen tasks via few shot inference. This yields a score of 54.3. 
%  Supervised SOTA uses 100\% of train samples and finetunes T5-3B to get a SOTA score of 70.99. 
%  \textbf{Our findings:} We instruction tune Tk-Instruct using just 6\% of train data available to get a score of 70.40 showcasing \textbf{quick learning} ability of the instruction tuned model.
%  We also develop a baseline where we finetune T5-3B using just 25\% of the data. 
%  The instruction tuned model is more efficient with respect to both the baseline and the SOTA.
%  % first row write inference on test set add a plus symbol 
%  % s. proposed instruction tuning. \textbf{Multi-Task Finetune:} We finetune T5-3B in MTL setup using 25\% of the eval set (1000 samples of each task) of Super Natural Instructions to get a 68.10 ROUGE-L score.
%  % \textbf{Multitask Instruction tuning:} We instruction tune on Tk-Instruct T5-3B using 6\% of the same data (200 samples of each task) to get a higher score of 70.40 ROUGE-L score showcasing \textbf{quick learning} of the instruction tuned model.}
%   % \vspace{-1mm}
% 	\label{fig:teaser}
%  % \vspace{-6mm}
% \end{figure*}

\begin{figure*}[t!]
	\centering 
	\small
	\includegraphics[width=\linewidth]{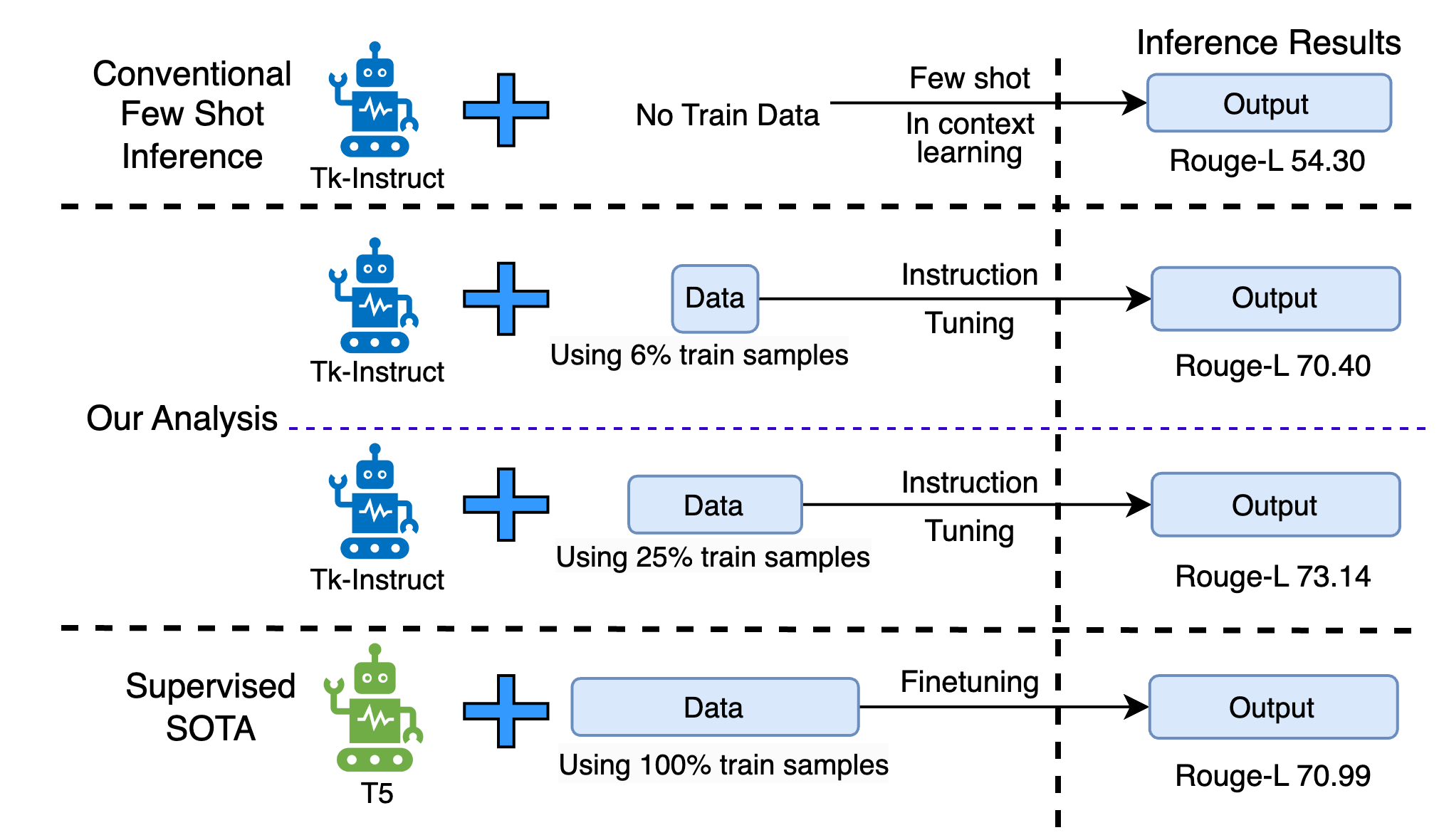}
	\caption{Showcasing the difference between the few shot inference, fully supervised finetuning, and our proposed analysis. 
 % The first row demonstrates conventional few shot inference using Tk-Instruct and evaluates unseen tasks, yielding a score of 54.30. 
 The first row represents conventional few shot inference using Tk-Instruct which results in a score of 54.30.
 The fourth row indicates supervised SOTA that uses 100\% of downstream train data to finetune T5-3B to get a SOTA score of 70.99. 
\textbf{Our findings} demonstrate the quick learning ability of the instruction tuned model. 
Using only 6\% of downstream train data, Tk-Instruct achieved a score of 70.40. Surpassing SOTA by 2 points with 25\% of downstream train data, our results highlight the MTL setting.}
	\label{fig:teaser}
\end{figure*}

%  The instruction tuned model is more efficient with respect to both the baseline and the SOTA.
%  % s. proposed instruction tuning. \textbf{Multi-Task Finetune:} We finetune T5-3B in MTL setup using 25\% of the eval set (1000 samples of each task) of Super Natural Instructions to get a 68.10 ROUGE-L score.
%  % \textbf{Multitask Instruction tuning:} We instruction tune on Tk-Instruct T5-3B using 6\% of the same data (200 samples of each task) to get a higher score of 70.40 ROUGE-L score showcasing \textbf{quick learning} of the instruction tuned model.}

Large language models (LLM) have achieved remarkable performances on several benchmark evaluation suites such as SuperGLUE \cite{wang2019superglue}, BIG-Bench Hard (BBH) \cite{suzgun2022challenging}, and HELM \cite{liang2022holistic}. 
 Research on LLMs has explored their abilities to follow instructions \cite{wei2022finetuned,mishra2022cross,wang-etal-2022-super} and has developed specialized models for the same (Flan, Instruct-GPT, Tk-Instruct, T0) \cite{wei2022finetuned,ouyang2022training,sanhmultitask}.
Recent studies in the instruction paradigm demonstrate the generalizability of models that are instruction tuned on training tasks and evaluated using few shot inference \cite{wang-etal-2022-super,wei2022finetuned}, as shown in the first row of Fig. \ref{fig:teaser}.
Despite this, SOTA performance is obtained by fully supervised finetuning on all available downstream training data, as shown in the 4th row of Fig. \ref{fig:teaser}.
In real-world situations, there is usually a limited amount of data available for finetuning, which is somewhere between few shot inference and fully supervised finetuning. % / full supervision / few shot and fully supervised inference  / the fully supervised setting
Given this context, we pose the question - if we use a small amount of the data from these downstream tasks, how quickly could the model learn in the instruction paradigm?

To answer this, we evaluate the minimal downstream training data required by instruction tuned models to perform transfer learning and match the performance of supervised SOTA models.
We experiment on unseen tasks of Super Natural Instructions (SuperNI) \cite{wang-etal-2022-super}, comprising of 119 tasks. 
We experiment with single-task learning (STL), i.e. training 119 task-specific models, and multi-task learning (MTL), where a single model is trained to solve all 119 tasks.
We use Tk-Instruct 3B (T5-3B, instruction tuned on 757 tasks of SuperNI) as the instruction tuned model \cite{wang-etal-2022-super} and use T5-3B \cite{raffel2020exploring} as our non-instruction tuned model. 
We find that in the STL setting, we achieve competitive results with just 5.91\% of the training data (68.34 ROUGE-L) and surpass the supervised SOTA score when using only 25.33\% of the entire dataset (71.71 ROUGE-L).
 In the MTL setting, when using 6\% of the train split, we match the SOTA performance (70.40 ROUGE-L), as shown in the 2nd row of Fig. 1.  
 We outperform SOTA by roughly 2\% (73.14 ROUGE-L) when using 25\% of the train split (3rd row of Fig. 1.) and 3.69\% when using 100\% of the train split. 
 To the best of our knowledge, we are first to explore the space of sample efficiency in instruction tuned large language models in both STL and MTL setups.
Details about the experimental setup are described in \S \ref{experiments}, and results are described in \S \ref{results}.

We analyze the impact of instructions by investigating sample efficiency across diverse ranges, by developing multiple baselines to simulate low resource settings pertaining to training data availability.
Our findings highlight sample efficiency achieved through instruction tuning, reaching up to 75\%, even in limited training data. 
% These results provide insights into the potential of instruction-based approaches in addressing data limitations and enhancing model performance.
% We also highlight the effect of instruction tuning as a pre-finetuning step.
% For this, we develop two baselines by pre-finetuning a T5-3B on the train set of SuperNI (100 samples each of 757 tasks) and were later finetuned on the downstream training set using 200 samples (6\% of total downstream training data) in both STL and MTL setup. 
% We instruction tune Tk-instruct on 200 samples to find that it outperforms baselines by 3\% and 5\% (in STL and MTL setup, respectively), showcasing the effect of instructions during pre-finetuning in terms of transfer learning as well.
We delve into the impact of instruction tuning as an initial pre-finetuning step. 
We develop two baselines (for both STL and MTL setups) employing pre-finetuning without instructions. 
These baselines undergo further finetuning on the downstream training set.
% Both the STL and MTL setups are considered in this evaluation.
Our findings demonstrate an increase in the performance of Tk-Instruct over the baselines by 3\% and 5\% in the STL and MTL setups, respectively. 
This highlights the impact of instructions during pre-finetuning in terms of facilitating transfer learning. 
We finally perform a category-wise analysis to investigate the impact of instruction tuning on different task categories. 
Our findings reveal that tasks falling under the textual entailment category demonstrate the most substantial improvements through instruction tuning. 
On the other hand, tasks related to question rewriting and title generation exhibit challenges and limitations when subjected to instruction tuning.

% This category-wise analysis provides valuable insights into the efficacy and limitations of instruction tuning across different task types. Understanding these variations contributes to a more nuanced understanding of the applicability and potential benefits of instruction tuning for specific task categories.

% Finally, we also conduct a category wise analysis to find that tasks of the textual entailment category had the best effects of instruction tuning, whereas tasks of question rewriting and title generation category suffer from instruction tuning. 

% All the studies are conducted in both STL and MTL setups. 

% To answer the above-mentioned questions,  In this work, we do a comparative study of instruction tuned models against large language models. 
% Our findings indicate that instruction tuning an instruction tuned model with a fraction of the data achieves the same performance as state-of-the-art language models. 
% The experimentation is done over unseen tasks of supernatural instructions (119 tasks). 
% The quick 

\noindent\textbf{Contributions:} (a) we show that an instruction tuned model using just 6\% of downstream train data matches the performance of a supervised SOTA model. (b) we find that the instruction tuned models perform up to 3\% better than the SOTA when instruction tuned with 100\% of the data. (c) to investigate scenarios with significantly limited downstream train data, we conduct a comprehensive analysis by constructing multiple baselines. (d) we show the impact of our method on various categories of tasks.

\section{Related Work}

Multi-task learning using LLMs (Language Models) has consistently shown performance benefits over task-specific learning \cite{mishra2022cross,ye2021crossfit,lin2022unsupervised,chen2022improving,yang2022cross,chenweighted,zhang2021hierarchical,chung2022scaling}. 
Instruction-based learning has emerged as a promising paradigm in LLMs \cite{malkiel-wolf-2021-maximal,shao2021concept2robot,ouyang2022training,kang2022instruction,honovich2022instruction,liu2022few,schick2022true,menon2022clues,Anderson2022VisionEI,su2022selective}, with recent studies exploring various aspects such as dialogue generation \cite{Gupta2022InstructDialIZ}, multimodality \cite{Xu2022MultiInstructIM}, chain of thought \cite{Wei2022ChainOT}, distributed training \cite{Jang2023ExploringTB}, and federated learning \cite{Zhang2023TowardsBT}.
Moreover, the effectiveness of Prompts and Instructions has been demonstrated in low-resource settings \cite{le2021many,puri2022many}, and different variants of prompting, including Scratchpad \cite{nye2021show}, Majority Voting \cite{wang2022self}, Reframing \cite{mishra-etal-2022-reframing}, Least-to-Most Prompting \cite{zhou2022least}, and Question Decomposition \cite{khot2020text, patel2022question}, have proven effective across various tasks.
Instruction-based techniques have also shown efficacy in different applications, such as NER \cite{wang2022instructionner}, program synthesis \cite{kuznia2022less}, style transfer \cite{reif2021recipe}, tabular question answering \cite{luo2022biotabqa}, relation extraction \cite{chen2021adaprompt}, and biomedical applications \cite{parmar-etal-2022-boxbart}.
However, the majority of the existing works have primarily focused on zero/few-shot inference scenarios \cite{Ivison2022HINTHI,gu2022learning}.

Our experiments are meant to simulate a real-world setting where few shot inference is not always necessary and there are some samples available for training. 
While there have been several studies on sample efficiency in other domains like reinforcement learning \cite{Yang2022TowardsAR,Guo2022ImprovingTS,Zhang2021OnTC,Yarats2019ImprovingSE,Yang2022SampleEO,Lagani2021HebbianSL}, to the best of our knowledge, no other work has explored the sample efficiency of instruction tuned models in a generalized fashion.
We also provide detailed analysis and task-specific insights with respect to various instruction tuning methods.
% giving additional details about where the proposed setup does and doesn't do well.
% A comparative analysis with existing work is shown in Table \ref{tab:comparison}.
% No previous work explored the limitations of task-specific insights in detail. 

\section{Instruction Tuning}

% In this section, we formally define the problem setup for single-task models, and instruction-based generalization across tasks. Along with that, we describe the models that used to achieve the generalization.

% \subsection{Modeling Paradigms}

% We assume that input and output instances pair $(X^t, Y^t)$ samples are given task the $t$. 
% Each sample of the task is described in terms of its instruction ${Inst}^t$. 

For each given task $t$, we assume that there are input and output instance pairs $(X^t, Y^t)$. Each sample of the task is described by its instruction ${inst}$.

\textbf{Single-task Learning (STL)} Traditional supervised models learn a mapping function ($f_M$) between input ($x$) and output ($y$) by using a training set of input/output pairs, $(x,y) \in ( {X^{t}_{train}}, {Y^{t}_{train}})$, for a given task $t$. The model is then evaluated on the test set for the same task, $({X^{t}_{test}}, {X^{t}_{test}})$. In the STL setup, $t$ models are trained for $t$ tasks in an individual fashion. 
% This type of learning is referred to as single-task learning. 

\textbf{Multi-task Learning (MTL)}
In this setup, the training data for all tasks are combined together. 
The goal of MTL models is to learn a mapping function ($f_M$) between the input ($x$) and output ($y$), such that $f_M(x) = y$, where $(x,y) \in ( {X^{t}_{train}}, {Y^{t}_{train}})$ for all $t$ tasks in a combined way. 
This model is then evaluated on task-specific instances $(x,y) \in ({X^{t}_{test}}, {Y^{t}_{test}})$. 
In contrast to single-task models, a single model is used to solve various tasks in this setup, which allows for generalization. 
% This setup is referred to as MTL (multi-task learning).

\begin{figure*}[t!]
	\centering
	\includegraphics[width=\linewidth, height= 6.5 cm]{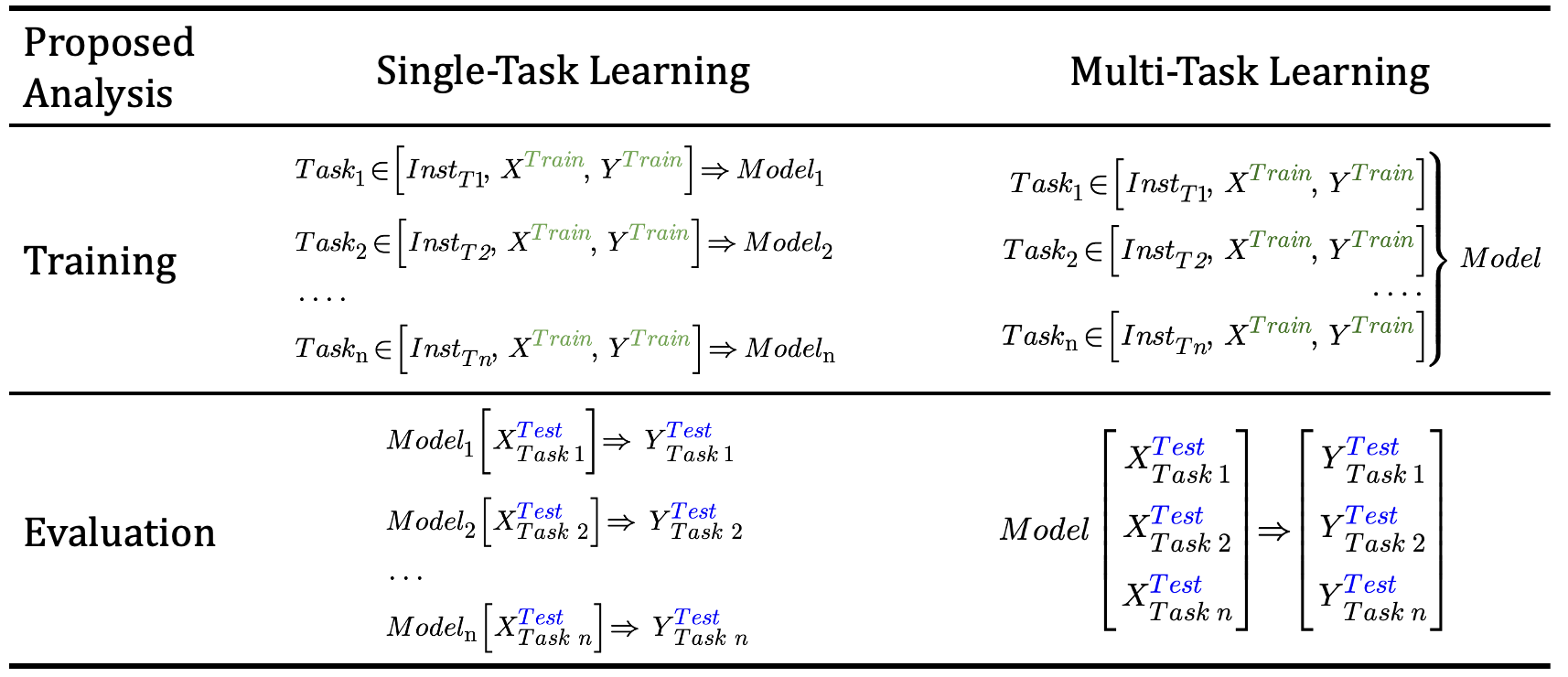}
	\caption{ Formulation of the proposed analysis. 
		In the Single-task learning (STL) setting, $f_{inst}$ is  instruction tuned individually on each downstream dataset. 
		In the Multi-task learning (MTL) setup, all downstream training tasks are combined together, and $f_{inst}$ is instruction tuned on all of them. 
  In both setups, the number of input samples from the downstream train data is varied. }
	\label{fig:flowchart}
	% 	\vspace{-0.2cm}
\end{figure*}

\textbf{Modelling Instruction Tuning}
In this setup, the mapping function takes an instruction ${inst}_t$ along with the input sample to give output as $y$; $f_M({inst},x) = y$.
Instruction tuning can be achieved in both Single-task and Multi-task setups. 
% An instruction-equipped sample consists of the following elements: \textbf{Definition} refers to the comprehensive explanation of task at hand and the specific instructions provided to the model to complete the given task successfully. 
% \textbf{Examples} consist of input/output pairs for a particular task instance.
% We follow the approach established in SuperNI to append two examples in the instruction prompts.
\textbf{Definition}: The term "Definition" pertains to the detailed explanation of the task at hand along with specific instructions provided, enabling the model to successful completion of the given task.
\textbf{Examples}: "Examples" refer to the input/output pairs associated with a particular instance of the task.
In line with the approach introduced in SuperNI, we incorporate two examples within the instruction prompts.
% \textbf{Instances} include input/output pairs for training samples taken from task datasets. 
% More details and examples of instruction tuned modeling can be found in the Appendix. 

\subsection{Proposed Analysis}

% We instruction tune 
% achieve both single-task and multi-task learning to show sample efficiency in both setups. 
We introduce two datasets for our task: $x_{pre-finetune}$ and $x_{train}$. 
These datasets are utilized as the pre-finetuning and downstream training datasets, respectively.
By pre-finetuning an LLM $f_M$ using instructions $({inst},x_{pre-finetune})$, we get instruction tuned model $f_{inst}$. 
$f_{inst}$ is now instruction tuned on the downstream train data $f_{inst}({inst}, x_{train}) = y$. 
For each experiment, different number of downstream train samples are used. 
% This starts with 0.3\%, 3\%, 5.91\%, 25.33\%, and 100\% of the downstream training set (10,100, 200, 1000, and all samples of each task). 
The instruction prompts change according to each downstream task.
% We aim to show that using a fraction of instruction-equipped data with an instruction tuned model can get the same accuracy as the model trained using a complete train set.   
For \textbf{STL} setup, $f_{inst}$ is individually instruction tuned on all tasks of the downstream train data; $t$ models are finetuned for $t$ tasks.
Each experiment will consist of $t$ models instruction tuned with a different number of training samples (Column 1 of Fig. \ref{fig:flowchart}).  
For \textbf{MTL} setup, one dataset is prepared by combining all the tasks of the downstream train dataset together. $f_{inst}$ is instruction tuned on the combined dataset. 
Similar to the last setting, the experiment will have a different number of training samples to highlight sample efficiency.

\subsection{Baselines}
To show a detailed analysis of instruction tuned modelling, pre-finetuning and cross-task generalization, we develop different baselines across both setups. 

\subsubsection{STL baselines}
We develop three baselines to compare the proposed modelling paradigm comprehensively. 
For the first baseline, we pre-finetune the model $f_M$ using $x_{pre-finetune}$ without instructions to get $f_{M1}$. 
$f_{M1}$ is now individually finetuned on all $t$ tasks of $x_{train}$ using 5.91\% of downstream train data to get $f_{STL-baseline-1}$. 
 For the second baseline ($f_{STL-baseline-2}$), we individually finetune the $f_M$ model on all $t$ tasks with 25.33\% of the downstream train set (1000 samples of each task) of $x_{train}$. 
 We develop the third baseline ($f_{STL-baseline-3}$) by individually finetuning the model $f_M$ on $t$ tasks of $x_{train}$ and use 100\% of the downstream training set. Third baseline serves as the supervised SOTA.
 The rationale for all three baselines is two-fold: First, to compare the baselines with the proposed model with fewer training samples. 
 % For example, the first baseline can be compared with the instruction tuned model trained over 3\% of the train set.
Second, the baseline is also used to compare the instruction tuned model trained with the same number of samples to observe the relative improvement in performance. 
Both advantages can be explained through the following example: $f_{STL-baseline-1}$ can highlight the effect of pre-finetuning, demonstrate sample efficiency and can be compared for performance improvement when the same number of samples are used.
 % we show the advantage of instruction prompts for transfer learning as well as it can be directly compared with  $f_{Inst}$  that was instruction tuned using the same number of samples.
% Using the First baseline, we show the advantage of using instructions in pre-finetuning as compared to the traditional approach.
% For example, the first baseline can be compared with the instruction tuned model trained over 5.91\% of the train set to observe the increase of instruction tuned models.
% usin the first baseline, we show the advantage of instruction prompts for transfer learning as well as it can be directly compared with  $f_{Inst}$ \hg{redundant following line, and restructure the sentence} that was instruction tuned using the same number of samples.
% Using the First baseline, we show the advantage of using instructions in pre-finetuning as compared to the traditional approach.

\begin{table*}[t!]
    \centering
    \small
    \resizebox{\linewidth}{!}
    {
    	\begin{tabular}{l|rrrrrrr|r}
    		\hline
    		\textbf{} &
    		\multicolumn{1}{c}{\textbf{\begin{tabular}[c]{@{}c@{}}Answerability \\ Classification\end{tabular}}} &
    		\multicolumn{1}{c}{\textbf{\begin{tabular}[c]{@{}c@{}}Coreference \\ Resolution\end{tabular}}} &
    		\multicolumn{1}{c}{\textbf{\begin{tabular}[c]{@{}c@{}}Data \\ to Text\end{tabular}}} &
    		\multicolumn{1}{c}{\textbf{\begin{tabular}[c]{@{}c@{}}Question \\ Rewriting\end{tabular}}} &
    		\multicolumn{1}{c}{\textbf{\begin{tabular}[c]{@{}c@{}}Textual \\ Entailment\end{tabular}}} &
    		\multicolumn{1}{c}{\textbf{\begin{tabular}[c]{@{}c@{}}Title \\ Generation\end{tabular}}} &
    		\multicolumn{1}{c|}{\textbf{\begin{tabular}[c]{@{}c@{}}Other \\ Categories\end{tabular}}} &
    		\multicolumn{1}{c}{\textbf{\begin{tabular}[c]{@{}c@{}}Grand Total\\ and percentage\end{tabular}}} \\ \hline
    		\textbf{\# of tasks} & 13    & 14    & 9     & 11    & 24    & 18    & 30    & \textbf{119 (100\%)}     \\
    		\textbf{100}         & 1300  & 1370  & 826   & 949   & 2376  & 1784  & 2994  & \textbf{11.5K (3.09\%)}  \\
    		\textbf{200}         & 2600  & 2441  & 1626  & 1849  & 4457  & 3484  & 5708  & \textbf{22.1K (5.91\%)}  \\
    		\textbf{1000}        & 11529 & 8831  & 8026  & 9049  & 19019 & 16514 & 21988 & \textbf{94.9K (25.33\%)} \\
    		\textbf{All}         & 43871 & 35560 & 36815 & 41391 & 78802 & 78357 & 59949 & \textbf{374.7K (100\%)}  \\ \hline
    	\end{tabular}
    }
    \caption{Category wise statistics of the downstream train data used.We note that since all the tasks have unequal samples, the total samples in each category will be different than \# of Tasks*Number of samples. Rows 100, 200, 1000, and all samples represent the sum of the  number of samples chosen during each experiment.}
    \label{tab:category_stats_main}
\end{table*}

\subsubsection{MTL baselines}
We develop two baselines to compare with the proposed MTL instruction tuned modelling paradigm. 
For the first baseline, we finetune $f_M$ on 25.33\% of the downstream train set $x_{train}$ in MTL setup to get $f_{MTL-baseline-1}$. 
For developing the second baseline, we pre-finetune the model $f_M$ using $x_{pre-finetune}$ without instructions to get $f_{M1}$. $f_{M1}$ model is finetuned in MTL setup on all $t$ tasks using  5.91\% of the downstream train data (200 samples of each task) of $x_{train}$. 
The rationale for creating two MTL baselines is similar to STL baselines; to show sample efficiency, highlight performance improvement when using the same number of samples, and showcase the effect of instructions in pre-finetuning.

\section{Experiments}
\label{experiments}

\subsection{Dataset}

\begin{wraptable}{r}{0.5\textwidth}
 % \vspace{-4.5mm}
\centering
\fontsize{9pt}{\baselineskip}\selectfont % font size
\renewcommand\tabcolsep{1pt} % column space
\renewcommand\arraystretch{1} % line space
\resizebox{\textwidth}{!}
    {
        \begin{tabular}{lr}
\hline
\textbf{Statistics}                    & \multicolumn{1}{l}{\textbf{}} \\ \hline
\# Total Tasks                         & 119                           \\
\# Total instances in train set     & 374745                        \\
\# Total instances in Test Set         & 11810                         \\
\# Total instances in pre-finetuning set  & 75700                         \\ \hline
Avg len of Train data w instructions   & 364.97                        \\
Avg len of Train data w/o instructions & 89.03                         \\ \hline
\end{tabular}
    }
    \caption{Statistics of the SuperNI dataset.
    Train and Test set refers to Downstream data.}
    \label{tab:train_stats}
     % \vspace{-6.0mm}
\end{wraptable}

We use seen tasks of SuperNI as the pre-finetuning set consisting of 757 tasks with 100 samples for each task.
We used the unseen task set of SuperNI as the downstream train data, consisting of 119 tasks that could be classified into 11 categories. 
Their statistics are presented in Table \ref{tab:train_stats}. 
The Dataset is classified into 11 categories of NLP tasks as shown in Table \ref{tab:category_stats}. 
For the sake of clarity, we have clubbed seven categories with fewer tasks into the \textit{others} category. 
Table \ref{tab:category_stats_main} gives detailed statistics across each category. 
Since not all the tasks have exactly the same number of samples,  we choose the maximum number available if the number of samples is below the threshold. 
For example, task 1388 (Table \ref{tab:dataset_stats} of Appendix \S \ref{detailed_results}) has a total of 191 samples. 
 We use 191 samples when finetuning with 200, 1000, and the entire dataset.

\begin{wraptable}{r}{0.5\textwidth}
% \vspace{-3.5mm}
 % \vspace{-4.5mm}
\centering
\fontsize{9pt}{\baselineskip}\selectfont % font size
\renewcommand\tabcolsep{1pt} % column space
\renewcommand\arraystretch{1} % line space

        \begin{tabular}{l|r}
        \hline
        \textbf{Category}            & \textbf{Count of Category} \\ \hline
        Textual Entailment           & 24                         \\
        Title Generation             & 18                         \\
        Coreference Resolution       & 14                         \\
        Answerability Classification & 13                         \\
        Question Rewriting           & 11                         \\
        Data to Text                 & 9                          \\
        Word Analogy                 & 8                          \\
        Cause Effect Classification  & 7                          \\
        Dialogue Act Recognition     & 7                          \\
        Keyword Tagging              & 5                          \\
        Overlap Extraction           & 2                          \\
        Grammar Error Correction     & 1                          \\ \hline
        \textbf{Grand Total}         & \textbf{119}               \\ \hline
        \end{tabular}
    \caption{Training sample statistics}
    \label{tab:category_stats}
     % \vspace{-6.0mm}
\end{wraptable}

We refer the reader to Appendix \S \ref{detailed_results}, total number of training samples in each task (Table \ref{tab:dataset_stats}), and the number of samples used for development of baselines and proposed analysis (Table \ref{tab:train_statistics}).

\paragraph{Dataset Split:} Hundred samples per task have been used for testing, and ten samples have been used for validation. 
These samples are not a part of the training set, and we have ensured that there is no data leakage. 
Each task contains the same number of samples for the test (100 samples) and validation set (10 samples). 
% \footnote{Dataset, the script to generate it are freely available at \url{https://github.com/srsawant34/efficient_instruction_learning}}

\subsection{Baselines} 

 We use $x_{pre-finetune}$ and $x_{train}$ in both STL and MTL setups. For creating baselines, both datasets are not equipped with instructions; they contain just input and output for conventional finetuning.

\subsubsection{Single Task Learning Baselines}

\textbf{STL Baseline 1:} A T5-3B model undergoes pre-finetuning using $x_{pre-finetune}$, which consists of 757 tasks from SuperNI. Subsequently, the model undergoes further finetuning on 200 samples per task from the downstream train data of SuperNI ($x_{train}$), resulting in 119 models.

\textbf{STL Baseline 2:} Each task from the downstream train data ($x_{train}$) is used to finetune a T5-3B model ($f_{M}$) with 1000 samples per task, resulting in 119 distinct models.

\textbf{STL Baseline 3 (Supervised SOTA):} A T5-3B model ($f_{M}$) is finetuned for each task using all available samples from the downstream train data.

\subsubsection{Multi Task Learning Baselines}

\textbf{MTL Baseline 1:} Similar to STL Baseline 1, we prefinetune a T5-3B $x_{pre-finetune}$ for each task to get the model $f_{M1}$. 
 $f_{M1}$ is now finetuned on 200 samples per task from the downstream train data ($x_{train}$) of SuperNI (in an MTL fashion). 
% Baseline 1 is used to compare with the proposed model (MTL Setup) finetuned with 100 (50\% fewer train samples) and 200 samples of instructional data. 

\textbf{MTL Baseline 2:} T5-3B model is now finetuned on 1000 samples per task from the downstream train data ($x_{train}$) of SuperNI (in an MTL fashion). 
% Baseline 2 is used to compare with the proposed model (MTL setup) finetuned with 1000 samples and all training samples of instructional data.

% \begin{wrapfigure}{r}{0.5\textwidth}
% \centering 
% 	\small
% 	\includegraphics[width=0.98\linewidth]{Pictures/stl_overall.png}
% 	\caption{Overall results in the Single-Task Learning setup (STL). The \textcolor{red}{red}, \textcolor{yellow}{Yellow}, and \textcolor{green}{Green} dots represent STL Baselines 1, 2, and 3, respectively. The horizontal dashed line is marked on the line graph to highlight the difference in train data required between the proposed approach and baselines. The x-axis is in logarithmic scale.
% }
% 	\label{fig:stl_overall}
%  \vspace{-6mm}
% \end{wrapfigure}

% \begin{wrapfigure}{r}{0.5\textwidth}
% \centering
%  % \vspace{-4.5mm}
%   \includegraphics[width=1.0\linewidth]{Pictures/stl_overall.png}
% 	\caption{Overall results in the STL setup. The \textcolor{red}{red}, \textcolor{yellow}{Yellow}, and \textcolor{green}{Green} dots represent STL Baselines 1, 2, and 3, respectively. The horizontal dashed line is marked on the line graph to highlight the difference in train data required between the proposed approach and baselines. The x-axis is in logarithmic scale.}
%   % \vspace{-1.0mm}
% 	\label{fig:stl_overall}
%   % \vspace{-6.0mm}
% \end{wrapfigure}

\subsection{Models and Evaluation Metrics}

\textbf{Models:} We use Tk-Instruct 3B as the instruction tuned model. 
% For the STL setup, we individually instruction tune Tk-instruct 
% of all tasks with different samples.
% For the STL setup, we combine the same splits as above and combine them in 1 trainset to evaluate performance.
For STL setup 952 models (119x8) were trained and 9 models were trained for MTL setup resulting in a total of \textbf{961} models for our analysis.
% A total of \textbf{961} models were trained for the entire experiment. 
All the models were trained on 6x Nvidia A100 40GB GPUs.

\textbf{Evaluation metric:} We consider all the tasks in the dataset as text generation problems and use the ROUGE-L score \cite{lin-2004-rouge} to evaluate the generated outputs.

% \begin{figure}[t!]
% 	\centering 
% 	\small
% 	\includegraphics[width=0.98\linewidth]{Pictures/stl_overall.png}
% 	\caption{
% 		Overall results in the Single-Task Learning setup (STL). The \textcolor{red}{red}, \textcolor{yellow}{Yellow}, and \textcolor{green}{Green} dots represent STL Baselines 1, 2, and 3, respectively. The horizontal dashed line is marked on the line graph to highlight the difference in train data required between the proposed approach and baselines. The x-axis is in logarithmic scale.
% 	}
% 	\label{fig:stl_overall}
% \end{figure}

% \begin{figure}[h!]
% 	\centering 
% 	\small
% 	\includegraphics[width=0.98\linewidth]{Pictures/mtl_overall.png}
% 	\caption{
% 		Proposed model overall results in Multi-Task Learning setup. The  Red and Yellow dots represent MTL Baselines 1 and 2, respectively. The score gap between the proposed approach and the baseline widens as compared to the STL setup.  
% 	}
% 	\label{fig:mtl_overall}
% \end{figure}

\section{Results}
\label{results}

\begin{wrapfigure}{r}{0.5\textwidth}
  \centering
  % \vspace{-4.5mm}
  \includegraphics[width=1.0\linewidth]{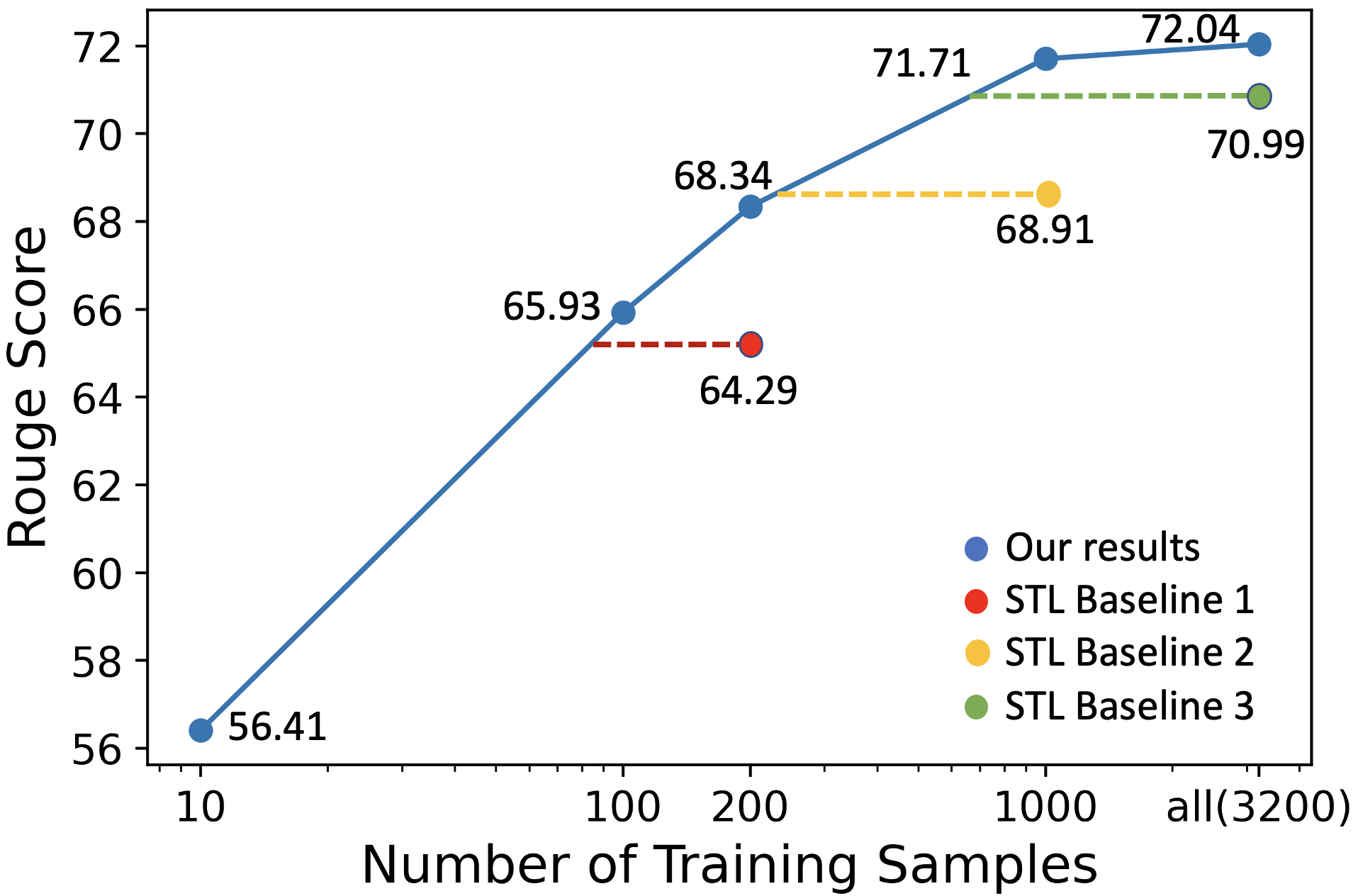}
  \caption{
  Results in the STL setup. 
  % The \textcolor{red}{red}, \textcolor{yellow}{Yellow}, and \textcolor{green}{Green} dots represent STL Baselines 1, 2, 
and 3, respectively. 
  The horizontal dashed line is marked on the graph to highlight the difference in train data required between the proposed approach and baselines. x-axis is in logarithmic scale. }
  % \vspace{-4.0mm}
     % \vspace{-6mm}
  \label{fig:stl_overall}
\end{wrapfigure}

The results are presented in two parts, STL and MTL results.
Each section contains overall results, Category wise results, and a comparison with the baselines that were defined earlier.

\subsection{Single Task Learning setup (STL)}

\begin{figure*}[t!]
	\centering 
	\small
	\includegraphics[width=0.98\linewidth]{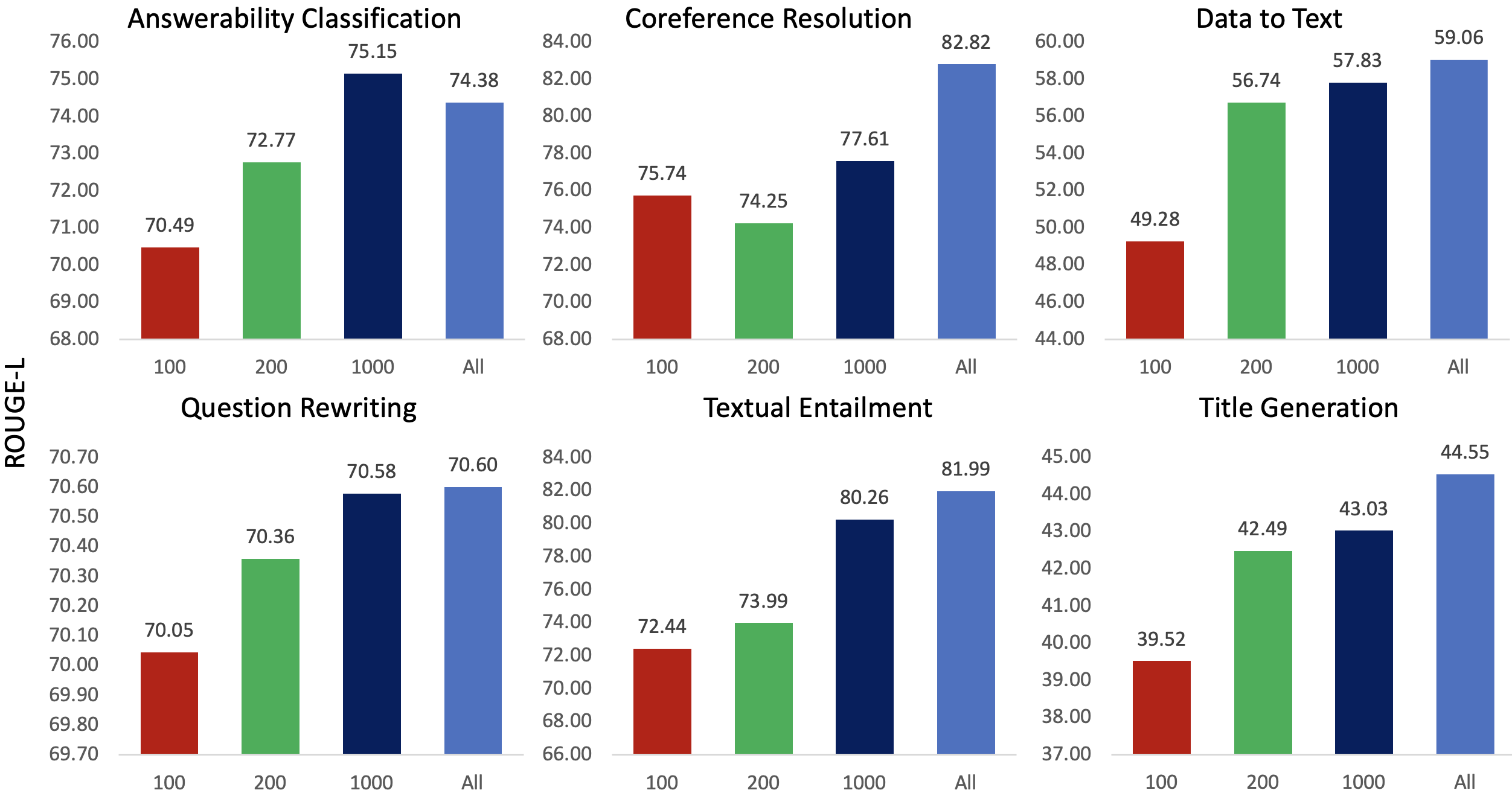}
	\caption{
		Histogram showing category wise results of the proposed approach in \textbf{STL} setting. 
		The x-axis shows avg number of training samples. the y-axis shows the rouge-L scores. 
		Most categories follow a conventional trend of performance increase as the number of training samples increases. 
  This trend has an exception in two places. 
  First:  Answerability classification score drops when all samples are used after 1000 (75.15 to 74.38). 
  Second: Coreference Resolution score drops when 200 samples are used after 100 (75.74 to 74.25).}
	\label{fig:stl_categories}
\end{figure*}

Fig. \ref{fig:stl_overall} shows the rouge score of instruction tuned models when training with different numbers of samples. 
We see that there is an overall increasing trend as the number of samples increases. 
From the figure we observe that a max ROUGE-L score of 72.04 is obtained when all samples are used. 
Category wise scores of the baselines are present in Table \ref{tab:stl_baselines}.
Figure \ref{fig:stl_categories} shows the category wise results of the instruction tuned models. 
% Almost all the categories and baselines follow an increasing trend. 
From the figure, we see that except for the Answerability Classification category, all the categories have an increasing trend. 
Detailed results regarding each category, baseline, and instruction tuning of the MTL setup can be found in Table \ref{tab:1m_categories}, Table \ref{tab:1m_baselines}, and Table \ref{tab:1m_scores} in Appendix \S \ref{detailed_results}.

% Table \ref{tab:1m_baselines} (Appendix \S \ref{detailed_results}) gives a detailed score of all tasks in each baseline. 
% % A detailed score of each baseline is present in Table \ref{tab:1m_baselines}.
% Detailed results about each task are present in Table \ref{tab:1m_scores}.
% Table \ref{tab:1m_categories} (Appendix \S \ref{detailed_results}) shows the category wise results of all three STL baselines.

\begin{table*}[t!]
    \centering
    \small
    \resizebox{\linewidth}{!}
    {
       \begin{tabular}{lrrrrrrr}
\hline
\multicolumn{1}{c}{\textbf{}} & \multicolumn{1}{c}{\textbf{\begin{tabular}[c]{@{}c@{}}Answerability \\ Classification\end{tabular}}} & \multicolumn{1}{c}{\textbf{\begin{tabular}[c]{@{}c@{}}Coreference \\ Resolution\end{tabular}}} & \multicolumn{1}{c}{\textbf{Data to Text}} & \multicolumn{1}{c}{\textbf{\begin{tabular}[c]{@{}c@{}}Question \\ Rewriting\end{tabular}}} & \multicolumn{1}{c}{\textbf{\begin{tabular}[c]{@{}c@{}}Textual \\ Entailment\end{tabular}}} & \multicolumn{1}{c}{\textbf{\begin{tabular}[c]{@{}c@{}}Title \\ Generation\end{tabular}}} & \multicolumn{1}{c}{\textbf{\begin{tabular}[c]{@{}c@{}}Other \\ Categories\end{tabular}}} \\ \hline
STL Baseline 1                & 70.38                                                                                                & 70.01                                                                                          & 49.75                                     & 68.12                                                                                      & 71.00                                                                                      & 44.71                                                                                    & 72.31                                                                                    \\
STL Baseline 2                & 78.77                                                                                                & 68.61                                                                                          & 50.89                                     & 71.00                                                                                      & 77.78                                                                                      & 46.78                                                                                    & 75.62                                                                                    \\
STL Baseline 3                & 80.36                                                                                                & 74.40                                                                                          & 52.84                                     & 71.11                                                                                      & 80.42                                                                                      & 48.35                                                                                    & 76.82                                                                                    \\ \hline
\end{tabular}
    }
    \caption{STL category wise scores for all three baselines. 
    We see that all the baselines follow a linearly increasing trend as the number of samples increases with baselines 1, 2, and 3 (200, 1000, and all samples used, respectively). 
    However, little improvement is observed in the Question rewriting category w.r.t baseline 2 and 3's ROUGE-L score (71.00 and 71.11, respectively).  
    Another deviation from the standard trend was observed in the Coreference Resolution category where STL baseline 1 had a higher score as compared to STL baseline 2 (70.01 and 68.61 respectively).}
    \label{tab:stl_baselines}
\end{table*}

\textbf{50\% efficient w.r.t STL baseline 1:} STL Baseline 1 is denoted by the red point in Figure \ref{fig:stl_overall}. 
The score with 100 samples is 65.93, and the score with baseline 1 is 64.29.
% We get a score of 68.34 when instruction tune with 200 samples observing a 4\% gain, showcasing 50\% sample efficiency. 
% When trained on 100 samples, the instruction tuned model uses roughly 50\% of data when compared to data used by baseline 1. 
% This demonstrates the effect of instructions in initial instruction tuning.

\textbf{Competitive performance using 6\% data:} 
STL Baseline 2 is denoted by the yellow point in Figure \ref{fig:stl_overall}). 
% was trained using 1000 samples. 
The instruction tuned model uses roughly 23.33\% of training samples when trained on 6\% data compared to STL baseline 2 which uses 25.33\% data.
The score with 200 samples/task is 68.34, while the score with baseline 2 is 68.91. 
% We get a competitive score (less than 1\% gap in \hg{scores word seems redundant} scores) by using just 23\% of samples in the instruction tuned setting compared to baseline two. 
% Even though the Baseline 2 score is slightly higher than the instruction tuned model, using just 23\% of the data to get a competitive score highlights the sample efficiency. 
In comparison with STL baseline 1 with the instruction tuned model (both trained using the 6\% data), an increase of $\sim 3\%$ is observed.

\textbf{Surpassing SOTA with 25\% data:} 
% STL Baseline 3 (denoted by the green point in Fig. \ref{fig:stl_overall}).
The instruction tuned model uses 25.33\% of the data compared to STL baseline 3 and gets a score of 71.71, compared to the 70.99 score of the baseline. 
% If we compare baseline 2 with the instruction tuned model (both trained using 1000 samples), we see a performance 
Comparison with baseline 2 (both trained using 25.33\% of the data) yields an increment of 3\%. 
When all samples are used, there is a further increase of 1.04\% (72.04 vs 70.99).

\textbf{Category wise effect of instruction tuning:} 
We observe that Answerability Classification and Title Generation categories scores decrease from instruction tuning as compared to baselines. 
The best scores from instruction tuning are 75.15 and 44.55 respectively which are significantly lower than the best baseline scores of 80.36 and 48.35. 
The categories that benefit from instruction tuning compared to the baselines are Coreference Resolution (82.82 vs 74.40) and Data to Text (59.06 vs 52.84).

\subsection{Multi task setup}

\begin{wrapfigure}{r}{0.5\textwidth}
\centering
 % \vspace{-4.5mm}
  \includegraphics[width=1.0\linewidth]{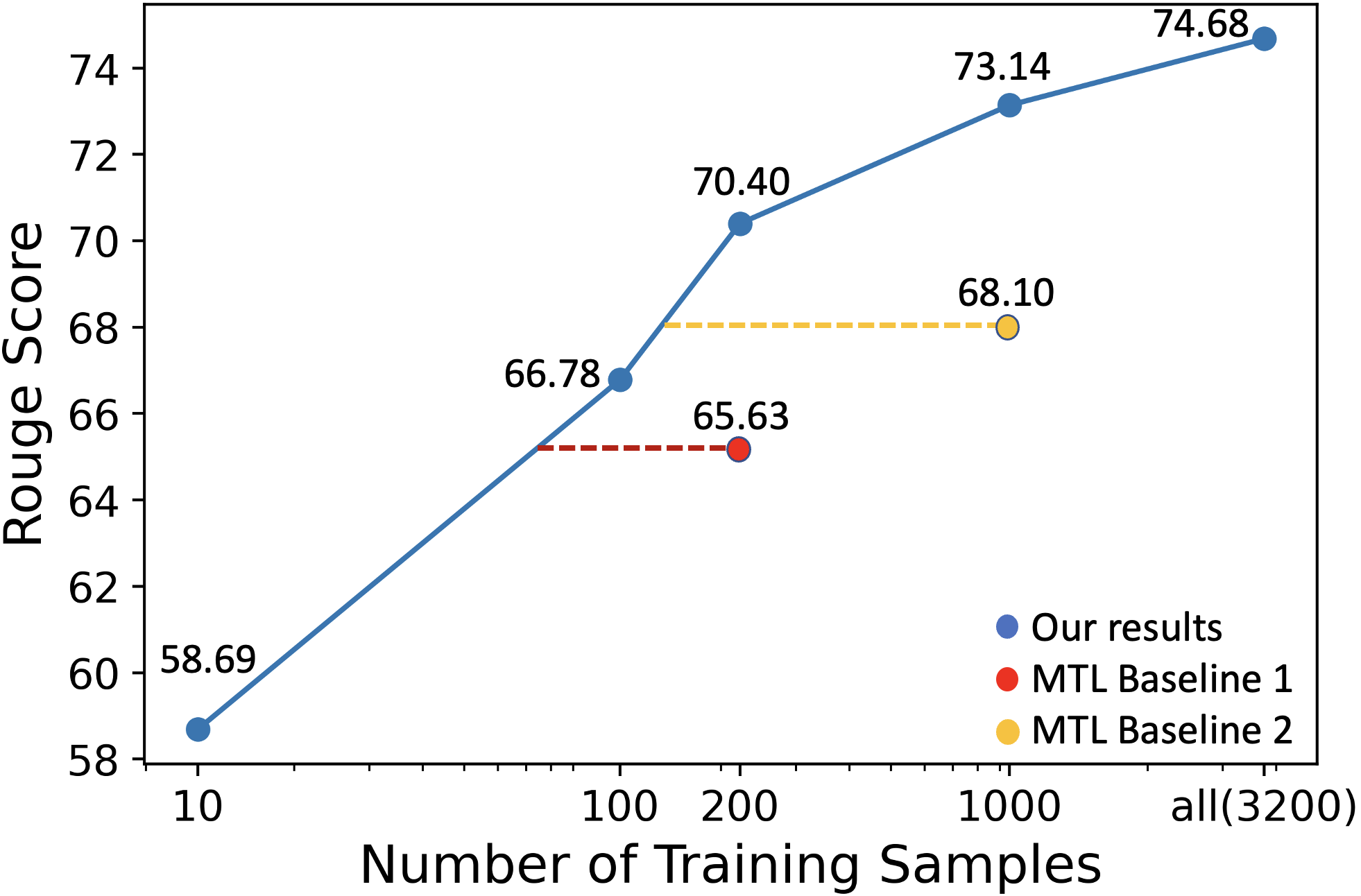}
	\caption{Proposed model overall results in MTL setup. The  Red and Yellow dots represent MTL Baselines 1 and 2, respectively. The score gap between the proposed approach and the baseline widens as compared to the STL setup. }
	\label{fig:mtl_overall}
  % \vspace{-3.0mm}
\end{wrapfigure}

\begin{figure*}[t!]
	\centering 
	\small
	\includegraphics[width=0.98\linewidth]{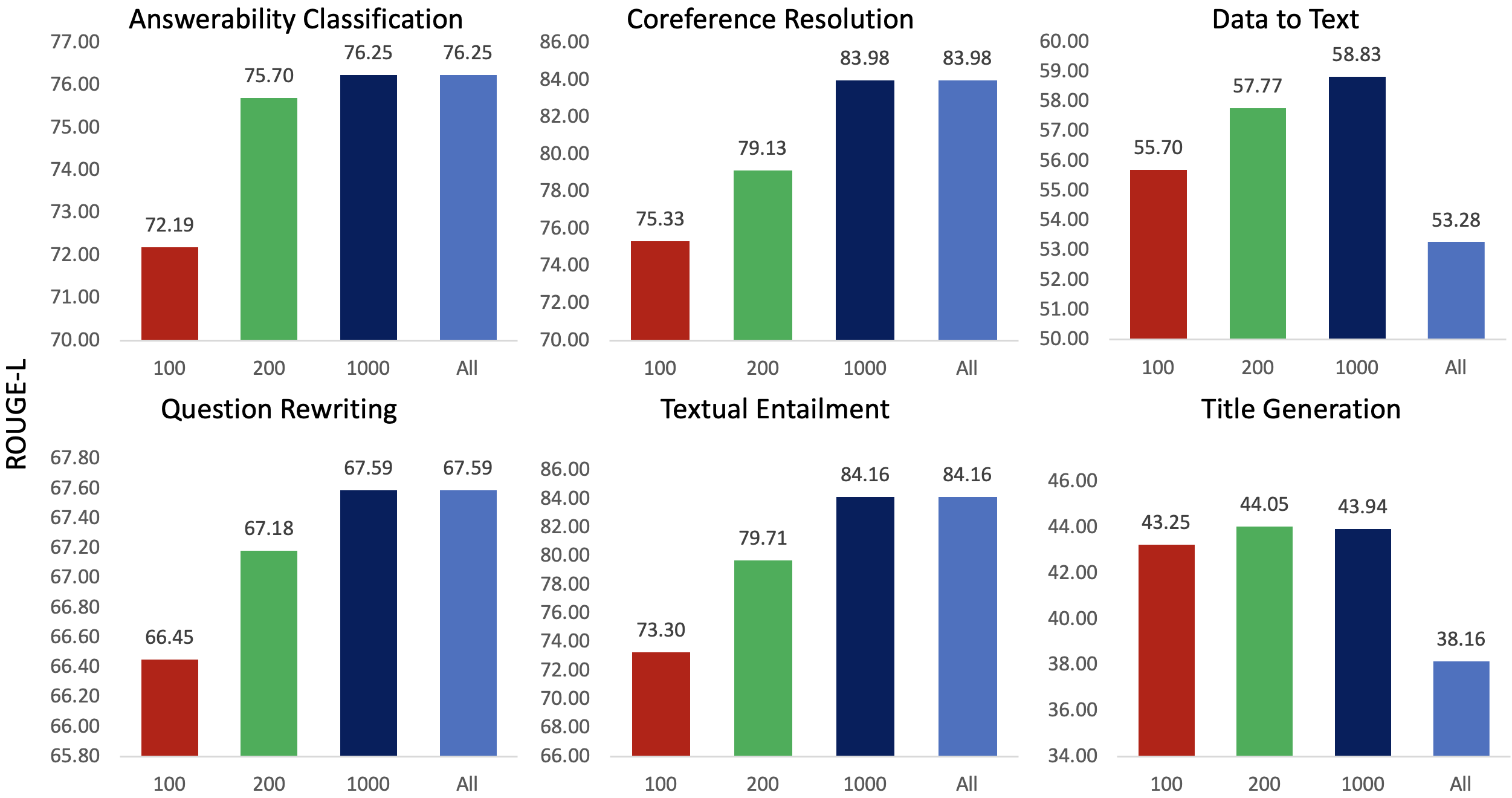}
	\caption{
	Histogram showing category wise results of the proposed approach in the MTL setting. 
	Similar to STL category wise scores, a linear trend is followed but has two exceptions. First:  Data to text  score drops when all samples are used after 1000 (58.83 to 53.28). Second:  Title Generation score drops when all samples are used after 1000  (43.94 to 38.16)
	}
	\label{fig:mtl_categories}
\end{figure*}

Figure \ref{fig:mtl_overall} shows the overall ROUGE-L score of instruction tuned models in the MTL setup.
% We see that there is an overall increasing trend as the number of samples increases. 
A max ROUGE-L score of 74.68 is obtained when all samples are used, surpassing the SOTA of 70.99. 
Figure \ref{fig:mtl_categories} shows the category wise results of the instruction tuned models and Table \ref{tab:mtl_baselines} showcases the baseline results. 
Detailed results regarding each category, baseline, and instruction tuning of the MTL setup can be found in Table \ref{tab:mtl_categories}, Table \ref{tab:mlt_baselines}, and Table \ref{tab:mlt_scores} respectively in Appendix \S \ref{detailed_results}.
% Detailed Category wise results are present in Table \ref{tab:mtl_categories}.
% Table \ref{tab:mtl_baselines} in Appendix shows the category wise results of both MTL baselines and Table \ref{tab:mlt_scores} shows all the scores of each task with different training samples. 
% Four of the six categories follow a uniformly increasing trend as the number of samples increases. 
% Contrasting to the usual trend, the "Data to text" category's ROUGE-L score drops slightly when all samples are used. The "Title Generation" category's score drops a bit sharply when all samples are used. 

% \begin{wrapfigure}{r}{0.5\textwidth}
% \centering 
% 	\small
% 	\includegraphics[width=0.98\linewidth]{Pictures/mtl_overall.png}
% 	\caption{
% 		Proposed model overall results in Multi-Task Learning setup. The  Red and Yellow dots represent MTL Baselines 1 and 2, respectively. The score gap between the proposed approach and the baseline widens as compared to the STL setup.  
% 	}
% 	\label{fig:mtl_overall}
% \end{wrapfigure}

\paragraph{50\% efficient compared to MTL Baseline 1:} 
MTL baseline 1 was trained on roughly 6\% of downstream train samples. 
% The instruction tuned model uses roughly 50\% of the data when trained on 3\% samples when compared to data used by baseline 1. 
The score with 3\% downstream data samples is 66.78, while the score with baseline 1 is 65.63. 
If we compare the instruction tuned model trained 6\% of downstream train samples, there is an increase of roughly 5\% points as it reaches a 70.40 rouge score.
% This also highlights the effect of instructions in initial instruction tuning. 

\begin{table*}[t!]
    \centering
    \small
    \resizebox{\linewidth}{!}
    {
       \begin{tabular}{lrrrrrrr}
\hline
\multicolumn{1}{c}{\textbf{}} & \multicolumn{1}{c}{\textbf{\begin{tabular}[c]{@{}c@{}}Answerability \\ Classification\end{tabular}}} & \multicolumn{1}{c}{\textbf{\begin{tabular}[c]{@{}c@{}}Coreference \\ Resolution\end{tabular}}} & \multicolumn{1}{c}{\textbf{Data to Text}} & \multicolumn{1}{c}{\textbf{\begin{tabular}[c]{@{}c@{}}Question \\ Rewriting\end{tabular}}} & \multicolumn{1}{c}{\textbf{\begin{tabular}[c]{@{}c@{}}Textual \\ Entailment\end{tabular}}} & \multicolumn{1}{c}{\textbf{\begin{tabular}[c]{@{}c@{}}Title \\ Generation\end{tabular}}} & \multicolumn{1}{c}{\textbf{\begin{tabular}[c]{@{}c@{}}Other \\ Categories\end{tabular}}} \\ \hline
MTL Baseline 1                & 75.35                                                                                                & 75.87                                                                                          & 55.85                                     & 75.64                                                                                      & 65.95                                                                                      & 62.19                                                                                    & 45.20                                                                                    \\
MTL Baseline 2                & 76.89                                                                                                & 81.93                                                                                          & 54.73                                     & 79.80                                                                                      & 67.89                                                                                      & 67.18                                                                                    & 42.19                                                                                    \\
 \hline
\end{tabular}
    }
    \caption{
    	MTL baseline category-wise scores. All categories follow an increasing trend as conventional thinking would suggest. The trend is however broken in the Data to Text category and other categories.  
    }
    \label{tab:mtl_baselines}
\end{table*}

\paragraph{Surpassing SOTA with 6\% train data:} MTL baseline two is denoted by the yellow point in Figure \ref{fig:stl_overall},  and was trained using 25\% of downstream train samples. 
The instruction tuned model uses $\sim 76\%$ fewer samples  when trained on 6\% of downstream train samples and gets a score of 70.40, while the score with baseline 2 is 68.10. 
The instruction tuned approach, trained on the same samples as baseline 2, improves by roughly 5\% (73.14 vs. 68.10). 
When all samples are used, a score of 74.68 is obtained, surpassing SOTA by 3\%.

\textbf{Category wise effect of instruction tuning:} 
Contrasting results to STL settings are observed as Question Rewriting and Title Generation categories experience a significant drop (12\% and 23\% points respectively) as compared to the best baseline scores. 
There is a significant improvement observed in the Textual Entailment category as the best score improves to 84.16 from 67.89 as compared to the baseline score. 

\subsection{Analysis}

\subsubsection{Category Wise Analysis} 
We analyze the performance across each category in both settings. 
In the STL setting, we find that the tasks belonging to the coreference resolution and data to text category have a high increase in ROUGE-L score with instruction tuning as compared to baseline approaches (78.23 vs. 71.00 ROUGE-L in coreference resolution and 57.88 vs. 51.16 in Data to Text). 
Question rewriting performed nearly the same (70.51 vs. 70.07 ROUGE-L) while answerability classification and title generation's score decreased w.r.t baseline (74.10 vs. 76.50 ROUGE-L in answerability classification and 43.36 vs. 46.61 in title generation). 
In the \textbf{MTL} setup, similar findings are observed but across different categories. 
We find that the tasks belonging to the textual entailment category have the highest increase with instruction tuning compared to the baseline (81.93 vs. 66.91 ROUGE-L).
Answerability classification performed nearly the same (75.97 vs. 76.11 ROUGE-L) while question rewriting and title generation's score decreased w.r.t baseline (67.38 vs. 77.71 ROUGE-L in question rewriting and 43.99 vs. 64.68 in title generation).

\textbf{MTL consistently outperforming STL:} We have performed multiple experiments across instruction tuned modelling settings while keeping the number of training samples the same across different settings. 
% The experiment was done by using 10, 100, 200, 1000, and the complete train sets. 
Across each training setup, 
% when a different number of training samples are used, the MTL setup consistently performs better than the STL setup. 
there is an increase of 1-2\% ROUGE-L in MTL setup as compared to STL.
Through both settings and all the experiments, it was evident that instruction tuned models perform better in the multi-task setup as compared to the single-task setup.

\textbf{Sample Efficiency:} Instruction tuned models showcase sample efficiency across both MTL and STL setups. 
Using multiple baselines, sample efficiency of roughly 50, 75, and 80\% are achieved across different spaces in both STL and MTL setups. 
We also see that when all samples are used in an instruction tuned setting, the overall performance beats SOTA.

\textbf{Effect of Instructions in pre-finetuning:} STL Baseline 1 and MTL Baseline 1 were pre-finetuned with 757 tasks of the SupperNI dataset but without instructions. 
They were later finetuned on 119 tasks downstream train data using 6\% in STL and MTL fashion. 
Instructions have a significant effect in pretraining as the instruction tuned model outperformed these baselines by 4 and 5\%, respectively, when trained with the same number of samples.

\section{Conclusion, Limitations and Future work}

% We take forward the instruction paradigm by incorporating a small portion of training data that is usually available for downstream tasks. 
% We instruction tune models on the small-scale training sets of downstream tasks and observe strong performance benefits for the Tk-instruct model on SuperNatural Instructions. 
% Our findings may indicate that instruction tuning helps a model to learn a task quickly with limited data.
% We hope our work will serve as the first step to make the instruction paradigm more realistic and future work can build on top of our work to elicit strong performance benefits across various benchmarks. 

In this study, we have taken a significant step forward in advancing the instruction paradigm by incorporating a small portion of training data commonly available for downstream tasks. 
By instruction tuning models on the small-scale training sets of downstream tasks, we have observed notable performance benefits for the Tk-instruct model on SuperNI. 
These findings suggest that instruction tuning can effectively assist a model in quickly learning a task even with limited data. 
While our work presents promising results, there are several limitations that need to be acknowledged.

Firstly, our experiments were limited to using T5-3B and its instruction tuned variant. 
We were unable to conduct experiments using T5-11B and its instruction tuned variant due to resource constraints. 
Therefore, further investigation using larger models and datasets would provide a more comprehensive understanding of the instruction tuning approach.
% Additionally, the evaluation of sample efficiency remains a challenging task. Although adding instructions in the form of definitions and examples to a dataset requires less human effort compared to incorporating chain-of-thought reasoning explanations to each sample, assessing sample efficiency in this setup presents difficulties that are yet to be resolved.
Additionally, our experiments focused solely on SuperNI, and we trained a total of 950+ models in the process. 
However, to achieve a more robust evaluation, it is necessary to extend our investigation to larger benchmarks such as BigBench \cite{Srivastava2022BeyondTI}. 
Nonetheless, evaluating these bigger benchmarks would require substantial time and resources.

In conclusion, our work represents an initial step towards making the instruction paradigm more realistic. 
Future research can build upon our findings to enhance performance across various benchmarks. 
Addressing the limitations mentioned above and conducting experiments with more extensive resources will contribute to a deeper understanding of instruction tuning and its potential for improving model performance in downstream tasks.

\printbibliography

@article{liang2022holistic,
  title={Holistic evaluation of language models},
  author={Liang, Percy and Bommasani, Rishi and Lee, Tony and Tsipras, Dimitris and Soylu, Dilara and Yasunaga, Michihiro and Zhang, Yian and Narayanan, Deepak and Wu, Yuhuai and Kumar, Ananya and others},
  journal={arXiv preprint arXiv:2211.09110},
  year={2022}
}

@article{wang2019superglue,
  title={Superglue: A stickier benchmark for general-purpose language understanding systems},
  author={Wang, Alex and Pruksachatkun, Yada and Nangia, Nikita and Singh, Amanpreet and Michael, Julian and Hill, Felix and Levy, Omer and Bowman, Samuel},
  journal={Advances in neural information processing systems},
  volume={32},
  year={2019}
}

@article{suzgun2022challenging,
  title={Challenging BIG-Bench tasks and whether chain-of-thought can solve them},
  author={Suzgun, Mirac and Scales, Nathan and Sch{\"a}rli, Nathanael and Gehrmann, Sebastian and Tay, Yi and Chung, Hyung Won and Chowdhery, Aakanksha and Le, Quoc V and Chi, Ed H and Zhou, Denny and others},
  journal={arXiv preprint arXiv:2210.09261},
  year={2022}
}

@inproceedings{mishra2022cross,
  title={Cross-Task Generalization via Natural Language Crowdsourcing Instructions},
  author={Mishra, Swaroop and Khashabi, Daniel and Baral, Chitta and Hajishirzi, Hannaneh},
  booktitle={Proceedings of the 60th Annual Meeting of the Association for Computational Linguistics (Volume 1: Long Papers)},
  pages={3470--3487},
  year={2022}
}

@inproceedings{ye2021crossfit,
  title={CrossFit: A Few-shot Learning Challenge for Cross-task Generalization in NLP},
  author={Ye, Qinyuan and Lin, Bill Yuchen and Ren, Xiang},
  booktitle={Proceedings of the 2021 Conference on Empirical Methods in Natural Language Processing},
  pages={7163--7189},
  year={2021}
}

@article{lin2022unsupervised,
  title={Unsupervised cross-task generalization via retrieval augmentation},
  author={Lin, Bill Yuchen and Tan, Kangmin and Miller, Chris and Tian, Beiwen and Ren, Xiang},
  journal={arXiv preprint arXiv:2204.07937},
  year={2022}
}

@article{chen2022improving,
  title={Improving Cross-task Generalization of Unified Table-to-text Models with Compositional Task Configurations},
  author={Chen, Jifan and Zhang, Yuhao and Liu, Lan and Dong, Rui and Chen, Xinchi and Ng, Patrick and Wang, William Yang and Huang, Zhiheng},
  journal={arXiv preprint arXiv:2212.08780},
  year={2022}
}

@inproceedings{yang2022cross,
  title={Cross-task knowledge distillation in multi-task recommendation},
  author={Yang, Chenxiao and Pan, Junwei and Gao, Xiaofeng and Jiang, Tingyu and Liu, Dapeng and Chen, Guihai},
  booktitle={Proceedings of the AAAI Conference on Artificial Intelligence},
  volume={36},
  pages={4318--4326},
  year={2022}
}

@inproceedings{chenweighted,
  title={Weighted Training for Cross-Task Learning},
  author={Chen, Shuxiao and Crammer, Koby and He, Hangfeng and Roth, Dan and Su, Weijie J},
  booktitle={International Conference on Learning Representations},
year={2021}
}

@inproceedings{zhang2021hierarchical,
  title={Hierarchical Task Learning from Language Instructions with Unified Transformers and Self-Monitoring},
  author={Zhang, Yichi and Chai, Joyce},
  booktitle={Findings of the Association for Computational Linguistics: ACL-IJCNLP 2021},
  pages={4202--4213},
  year={2021}
}

@article{gu2022learning,
  title={Learning Instructions with Unlabeled Data for Zero-Shot Cross-Task Generalization},
  author={Gu, Yuxian and Ke, Pei and Zhu, Xiaoyan and Huang, Minlie},
  journal={arXiv preprint arXiv:2210.09175},
  year={2022}
}

@article{shao2021concept2robot,
  title={Concept2robot: Learning manipulation concepts from instructions and human demonstrations},
  author={Shao, Lin and Migimatsu, Toki and Zhang, Qiang and Yang, Karen and Bohg, Jeannette},
  journal={The International Journal of Robotics Research},
  volume={40},
  number={12-14},
  pages={1419--1434},
  year={2021},
  publisher={SAGE Publications Sage UK: London, England}
}

@article{ouyang2022training,
  title={Training language models to follow instructions with human feedback},
  author={Ouyang, Long and Wu, Jeffrey and Jiang, Xu and Almeida, Diogo and Wainwright, Carroll and Mishkin, Pamela and Zhang, Chong and Agarwal, Sandhini and Slama, Katarina and Ray, Alex and others},
  journal={Advances in Neural Information Processing Systems},
  volume={35},
  pages={27730--27744},
  year={2022}
}

@article{kang2022instruction,
  title={Instruction-based learning: A review},
  author={Kang, Weixi and Hern{\'a}ndez, S{\`o}nia Pineda and Wang, Junxin and Malvaso, Antonio},
  journal={Neuropsychologia},
  pages={108142},
  year={2022},
  publisher={Elsevier}
}

@article{chung2022scaling,
  title={Scaling instruction-finetuned language models},
  author={Chung, Hyung Won and Hou, Le and Longpre, Shayne and Zoph, Barret and Tay, Yi and Fedus, William and Li, Eric and Wang, Xuezhi and Dehghani, Mostafa and Brahma, Siddhartha and others},
  journal={arXiv preprint arXiv:2210.11416},
  year={2022}
}

@article{honovich2022instruction,
  title={Instruction Induction: From Few Examples to Natural Language Task Descriptions},
  author={Honovich, Or and Shaham, Uri and Bowman, Samuel R and Levy, Omer},
  journal={arXiv preprint arXiv:2205.10782},
  year={2022}
}

@article{liu2022few,
  title={Few-shot parameter-efficient fine-tuning is better and cheaper than in-context learning},
  author={Liu, Haokun and Tam, Derek and Muqeeth, Mohammed and Mohta, Jay and Huang, Tenghao and Bansal, Mohit and Raffel, Colin A},
  journal={Advances in Neural Information Processing Systems},
  volume={35},
  pages={1950--1965},
  year={2022}
}

@article{schick2022true,
  title={True Few-Shot Learning with Prompts—A Real-World Perspective},
  author={Schick, Timo and Sch{\"u}tze, Hinrich},
  journal={Transactions of the Association for Computational Linguistics},
  volume={10},
  pages={716--731},
  year={2022},
  publisher={MIT Press}
}

@inproceedings{menon2022clues,
  title={CLUES: A Benchmark for Learning Classifiers using Natural Language Explanations},
  author={Menon, Rakesh and Ghosh, Sayan and Srivastava, Shashank},
  booktitle={Proceedings of the 60th Annual Meeting of the Association for Computational Linguistics (Volume 1: Long Papers)},
  pages={6523--6546},
  year={2022}
}

@inproceedings{Anderson2022VisionEI,
  title={Vision Encoders in Visual Question Answering},
  author={Ryan R. Anderson},
  year={2022}
}

@article{su2022selective,
  title={Selective annotation makes language models better few-shot learners},
  author={Su, Hongjin and Kasai, Jungo and Wu, Chen Henry and Shi, Weijia and Wang, Tianlu and Xin, Jiayi and Zhang, Rui and Ostendorf, Mari and Zettlemoyer, Luke and Smith, Noah A and others},
  journal={arXiv preprint arXiv:2209.01975},
  year={2022}
}

@inproceedings{wang-etal-2022-super,
    title = "Super-{N}atural{I}nstructions: Generalization via Declarative Instructions on 1600+ {NLP} Tasks",
    author = "Wang, Yizhong  and
      Mishra, Swaroop  and
      Alipoormolabashi, Pegah  and
      Kordi, Yeganeh  and
      Mirzaei, Amirreza  and
      Naik, Atharva  and
      Ashok, Arjun  and
      Dhanasekaran, Arut Selvan  and
      Arunkumar, Anjana  and
      Stap, David  and
      Pathak, Eshaan  and
      Karamanolakis, Giannis  and
      Lai, Haizhi  and
      Purohit, Ishan  and
      Mondal, Ishani  and
      Anderson, Jacob  and
      Kuznia, Kirby  and
      Doshi, Krima  and
      Pal, Kuntal Kumar  and
      Patel, Maitreya  and
      Moradshahi, Mehrad  and
      Parmar, Mihir  and
      Purohit, Mirali  and
      Varshney, Neeraj  and
      Kaza, Phani Rohitha  and
      Verma, Pulkit  and
      Puri, Ravsehaj Singh  and
      Karia, Rushang  and
      Doshi, Savan  and
      Sampat, Shailaja Keyur  and
      Mishra, Siddhartha  and
      Reddy A, Sujan  and
      Patro, Sumanta  and
      Dixit, Tanay  and
      Shen, Xudong",
    booktitle = "Proceedings of the 2022 Conference on Empirical Methods in Natural Language Processing",
    month = dec,
    year = "2022",
    address = "Abu Dhabi, United Arab Emirates",
    publisher = "Association for Computational Linguistics",
    url = "https://aclanthology.org/2022.emnlp-main.340",
    pages = "5085--5109",
}

@article{wang2022self,
  title={Self-Instruct: Aligning Language Model with Self Generated Instructions},
  author={Wang, Yizhong and Kordi, Yeganeh and Mishra, Swaroop and Liu, Alisa and Smith, Noah A and Khashabi, Daniel and Hajishirzi, Hannaneh},
  journal={arXiv preprint arXiv:2212.10560},
  year={2022}
}

@inproceedings{sanhmultitask,
  title={Multitask Prompted Training Enables Zero-Shot Task Generalization},
  author={Sanh, Victor and Webson, Albert and Raffel, Colin and Bach, Stephen and Sutawika, Lintang and Alyafeai, Zaid and Chaffin, Antoine and Stiegler, Arnaud and Raja, Arun and Dey, Manan and others},
  booktitle={International Conference on Learning Representations},
  year={2021}
}

@article{raffel2020exploring,
  title={Exploring the limits of transfer learning with a unified text-to-text transformer.},
  author={Raffel, Colin and Shazeer, Noam and Roberts, Adam and Lee, Katherine and Narang, Sharan and Matena, Michael and Zhou, Yanqi and Li, Wei and Liu, Peter J and others},
  journal={J. Mach. Learn. Res.},
  volume={21},
  number={140},
  pages={1--67},
  year={2020}
}

@article{Srivastava2022BeyondTI,
  title={Beyond the Imitation Game: Quantifying and extrapolating the capabilities of language models},
  author={Aarohi Srivastava and Abhinav Rastogi and Abhishek Rao and Abu Awal Md Shoeb and Abubakar Abid and Adam Fisch and Adam R. Brown and Adam Santoro and Aditya Gupta and Adri{\`a} Garriga-Alonso and Agnieszka Kluska and Aitor Lewkowycz and Akshat Agarwal and Alethea Power and Alex Ray and Alex Warstadt and Alexander W. Kocurek and Ali Safaya and Ali Tazarv and Alice Xiang and Alicia Parrish and Allen Nie and Aman Hussain and Amanda Askell and Amanda Dsouza and Ameet Annasaheb Rahane and Anantharaman S. Iyer and Anders Andreassen and Andrea Santilli and Andreas Stuhlmuller and Andrew M. Dai and Andrew D. La and Andrew Kyle Lampinen and Andy Zou and Angela Jiang and Angelica Chen and Anh Vuong and Animesh Gupta and Anna Gottardi and Antonio Norelli and Anu Venkatesh and Arash Gholamidavoodi and Arfa Tabassum and Arul Menezes and Arun Kirubarajan and Asher Mullokandov and Ashish Sabharwal and Austin Herrick and Avia Efrat and Aykut Erdem and Ayla Karakacs and Bridget R. Roberts and Bao Sheng Loe and Barret Zoph and Bartlomiej Bojanowski and Batuhan Ozyurt and Behnam Hedayatnia and Behnam Neyshabur and Benjamin Inden and Benno Stein and Berk Ekmekci and Bill Yuchen Lin and Blake Stephen Howald and Cameron Diao and Cameron Dour and Catherine Stinson and Cedrick Argueta and C'esar Ferri Ram'irez and Chandan Singh and Charles Rathkopf and Chenlin Meng and Chitta Baral and Chiyu Wu and Chris Callison-Burch and Chris Waites and Christian Voigt and Christopher D. Manning and Christopher Potts and Cindy Tatiana Ramirez and Clara Rivera and Clemencia Siro and Colin Raffel and Courtney Ashcraft and Cristina Garbacea and Damien Sileo and Daniel H Garrette and Dan Hendrycks and Dan Kilman and Dan Roth and Daniel Freeman and Daniel Khashabi and Daniel Levy and Daniel Gonz'alez and Danny Hernandez and Danqi Chen and Daphne Ippolito and Dar Gilboa and David Dohan and D. Drakard and David Jurgens and Debajyoti Datta and Deep Ganguli and Denis Emelin and Denis Kleyko and Deniz Yuret and Derek Chen and Derek Tam and Dieuwke Hupkes and Diganta Misra and Dilyar Buzan and Dimitri Coelho Mollo and Diyi Yang and Dong-Ho Lee and Ekaterina Shutova and Ekin Dogus Cubuk and Elad Segal and Eleanor Hagerman and Elizabeth Barnes and Elizabeth P. Donoway and Ellie Pavlick and Emanuele Rodol{\`a} and Emma FC Lam and Eric Chu and Eric Tang and Erkut Erdem and Ernie Chang and Ethan A. Chi and Ethan Dyer and Ethan J. Jerzak and Ethan Kim and Eunice Engefu Manyasi and Evgenii Zheltonozhskii and Fan Xia and Fatemeh Siar and Fernando Mart'inez-Plumed and Francesca Happ'e and François Chollet and Frieda Rong and Gaurav Mishra and Genta Indra Winata and Gerard de Melo and Germ{\'a}n Kruszewski and Giambattista Parascandolo and Giorgio Mariani and Gloria Wang and Gonzalo Jaimovitch-L'opez and Gregor Betz and Guy Gur-Ari and Hana Galijasevic and Han Sol Kim and Hannah Rashkin and Hanna Hajishirzi and Harsh Mehta and Hayden Bogar and Henry Shevlin and Hinrich Sch{\"u}tze and Hiromu Yakura and Hongming Zhang and Hubert Wong and Ian Aik-Soon Ng and Isaac Noble and Jaap Jumelet and Jack Geissinger and John Kernion and Jacob Hilton and Jaehoon Lee and Jaime Fern{\'a}ndez Fisac and J. Brooker Simon and James Koppel and James Zheng and James Zou and Jan Koco'n and Jana Thompson and Jared Kaplan and Jarema Radom and Jascha Narain Sohl-Dickstein and Jason Phang and Jason Wei and Jason Yosinski and Jekaterina Novikova and Jelle Bosscher and Jenni Marsh and Jeremy Kim and Jeroen Taal and Jesse Engel and Jesujoba Oluwadara Alabi and Jiacheng Xu and Jiaming Song and Jillian Tang and Jane W Waweru and John Burden and John Miller and John U. Balis and Jonathan Berant and Jorg Frohberg and Jos Rozen and Jos{\'e} Hern{\'a}ndez-Orallo and Joseph Boudeman and Joseph Jones and Joshua B. Tenenbaum and Joshua S. Rule and Joyce Chua and Kamil Kanclerz and Karen Livescu and Karl Krauth and Karthik Gopalakrishnan and Katerina Ignatyeva and Katja Markert and Kaustubh D. Dhole and Kevin Gimpel and Kevin Ochieng’ Omondi and Kory Wallace Mathewson and Kristen Chiafullo and Ksenia Shkaruta and Kumar Shridhar and Kyle McDonell and Kyle Richardson and Laria Reynolds and Leo Gao and Li Zhang and Liam Dugan and Lianhui Qin and Lidia Contreras-Ochando and Louis-Philippe Morency and Luca Moschella and Luca Lam and Lucy Noble and Ludwig Schmidt and Luheng He and Luis Oliveros Col'on and Luke Metz and Lutfi Kerem cSenel and Maarten Bosma and Maarten Sap and Maartje ter Hoeve and Madotto Andrea and Maheen Saleem Farooqi and Manaal Faruqui and Mantas Mazeika and Marco Baturan and Marco Marelli and Marco Maru and M Quintana and Marie Tolkiehn and Mario Giulianelli and Martha Lewis and Martin Potthast and Matthew Leavitt and Matthias Hagen and M'aty'as Schubert and Medina Baitemirova and Melissa Arnaud and Melvin Andrew McElrath and Michael A. Yee and Michael Cohen and Mi Gu and Michael I. Ivanitskiy and Michael Starritt and Michael Strube and Michal Swkedrowski and Michele Bevilacqua and Michihiro Yasunaga and Mihir Kale and Mike Cain and Mimee Xu and Mirac Suzgun and Monica Tiwari and Mohit Bansal and Moin Aminnaseri and Mor Geva and Mozhdeh Gheini and T MukundVarma and Nanyun Peng and Nathan Chi and Nayeon Lee and Neta Gur-Ari Krakover and Nicholas Cameron and Nicholas S. Roberts and Nicholas Doiron and Nikita Nangia and Niklas Deckers and Niklas Muennighoff and Nitish Shirish Keskar and Niveditha Iyer and Noah Constant and Noah Fiedel and Nuan Wen and Oliver Zhang and Omar Agha and Omar Elbaghdadi and Omer Levy and Owain Evans and Pablo Antonio Moreno Casares and Parth Doshi and Pascale Fung and Paul Pu Liang and Paul Vicol and Pegah Alipoormolabashi and Peiyuan Liao and Percy Liang and Peter W. Chang and Peter Eckersley and Phu Mon Htut and Pi-Bei Hwang and P. Milkowski and Piyush S. Patil and Pouya Pezeshkpour and Priti Oli and Qiaozhu Mei and QING LYU and Qinlang Chen and Rabin Banjade and Rachel Etta Rudolph and Raefer Gabriel and Rahel Habacker and Ram'on Risco Delgado and Rapha{\"e}l Milli{\`e}re and Rhythm Garg and Richard Barnes and Rif A. Saurous and Riku Arakawa and Robbe Raymaekers and Robert Frank and Rohan Sikand and Roman Novak and Roman Sitelew and Ronan Le Bras and Rosanne Liu and Rowan Jacobs and Rui Zhang and Ruslan Salakhutdinov and Ryan Chi and Ryan Lee and Ryan Stovall and Ryan Teehan and Rylan Yang and Sahib J. Singh and Saif M. Mohammad and Sajant Anand and Sam Dillavou and Sam Shleifer and Sam Wiseman and Samuel Gruetter and Sam Bowman and Samuel S. Schoenholz and Sanghyun Han and Sanjeev Kwatra and Sarah A. Rous and Sarik Ghazarian and Sayan Ghosh and Sean Casey and Sebastian Bischoff and Sebastian Gehrmann and Sebastian Schuster and Sepideh Sadeghi and Shadi S. Hamdan and Sharon Zhou and Shashank Srivastava and Sherry Shi and Shikhar Singh and Shima Asaadi and Shixiang Shane Gu and Shubh Pachchigar and Shubham Toshniwal and Shyam Upadhyay and Shyamolima Debnath and Siamak Shakeri and Simon Thormeyer and Simone Melzi and Siva Reddy and Sneha Priscilla Makini and Soo-hwan Lee and Spencer Bradley Torene and Sriharsha Hatwar and Stanislas Dehaene and Stefan Divic and Stefano Ermon and Stella Rose Biderman and Stephanie C. Lin and S. Prasad and Steven T. Piantadosi and Stuart M. Shieber and Summer Misherghi and Svetlana Kiritchenko and Swaroop Mishra and Tal Linzen and Tal Schuster and Tao Li and Tao Yu and Tariq A. Ali and Tatsuo Hashimoto and Te-Lin Wu and Theo Desbordes and Theodore Rothschild and Thomas Phan and Tianle Wang and Tiberius Nkinyili and Timo Schick and T. N. Kornev and Timothy Telleen-Lawton and Titus Tunduny and Tobias Gerstenberg and Trenton Chang and Trishala Neeraj and Tushar Khot and Tyler O’Brien Shultz and Uri Shaham and Vedant Misra and Vera Demberg and Victoria Nyamai and Vikas Raunak and Vinay Venkatesh Ramasesh and Vinay Uday Prabhu and Vishakh Padmakumar and Vivek Srikumar and William Fedus and William Saunders and William Zhang and W Vossen and Xiang Ren and Xiaoyu Tong and Xinyi Wu and Xudong Shen and Yadollah Yaghoobzadeh and Yair Lakretz and Yang Song and Yasaman Bahri and Ye Ji Choi and Yichi Yang and Yiding Hao and Yifu Chen and Yonatan Belinkov and Yu Hou and Yu Hou and Yuntao Bai and Zachary Seid and Zhao Xinran and Zhuoye Zhao and Zi Fu Wang and Zijie J. Wang and Zirui Wang and Ziyi Wu and Sahib Singh and Uri Shaham},
  journal={ArXiv},
  year={2022},
  volume={abs/2206.04615}
}

@inproceedings{malkiel-wolf-2021-maximal,
    title = "Maximal Multiverse Learning for Promoting Cross-Task Generalization of Fine-Tuned Language Models",
    author = "Malkiel, Itzik  and
      Wolf, Lior",
    booktitle = "Proceedings of the 16th Conference of the European Chapter of the Association for Computational Linguistics: Main Volume",
    month = apr,
    year = "2021",
    address = "Online",
    publisher = "Association for Computational Linguistics",
    url = "https://aclanthology.org/2021.eacl-main.14",
    doi = "10.18653/v1/2021.eacl-main.14",
    pages = "187--199",
}

@inproceedings{lin-2004-rouge,
    title = "{ROUGE}: A Package for Automatic Evaluation of Summaries",
    author = "Lin, Chin-Yew",
    booktitle = "Text Summarization Branches Out",
    month = jul,
    year = "2004",
    address = "Barcelona, Spain",
    publisher = "Association for Computational Linguistics",
    url = "https://aclanthology.org/W04-1013",
    pages = "74--81",
}

@inproceedings{Gupta2022InstructDialIZ,
  title={InstructDial: Improving Zero and Few-shot Generalization in Dialogue through Instruction Tuning},
  author={Prakhar Gupta and Cathy Jiao and Yi-Ting Yeh and Shikib Mehri and Maxine Esk{\'e}nazi and Jeffrey P. Bigham},
  booktitle={Conference on Empirical Methods in Natural Language Processing},
  year={2022}
}

@article{Xu2022MultiInstructIM,
  title={MultiInstruct: Improving Multi-Modal Zero-Shot Learning via Instruction Tuning},
  author={Zhiyang Xu and Ying Shen and Lifu Huang},
  journal={ArXiv},
  year={2022},
  volume={abs/2212.10773}
}

@article{Ivison2022HINTHI,
  title={HINT: Hypernetwork Instruction Tuning for Efficient Zero-Shot Generalisation},
  author={Hamish Ivison and Akshita Bhagia and Yizhong Wang and Hannaneh Hajishirzi and Matthew E. Peters},
  journal={ArXiv},
  year={2022},
  volume={abs/2212.10315}
}

@article{Jang2023ExploringTB,
  title={Exploring the Benefits of Training Expert Language Models over Instruction Tuning},
  author={Joel Jang and Seungone Kim and Seonghyeon Ye and Doyoung Kim and Lajanugen Logeswaran and Moontae Lee and Kyungjae Lee and Minjoon Seo},
  journal={ArXiv},
  year={2023},
  volume={abs/2302.03202}
}

@inproceedings{Zhang2023TowardsBT,
  title={Towards Building the Federated GPT: Federated Instruction Tuning},
  author={Jianyi Zhang and Saeed Vahidian and Martin Kuo and Chunyuan Li and Ruiyi Zhang and Guoyin Wang and Yiran Chen},
  year={2023}
}

@inproceedings{Yang2022TowardsAR,
  title={Towards Applicable Reinforcement Learning: Improving the Generalization and Sample Efficiency with Policy Ensemble},
  author={Zhengyu Yang and Kan Ren and Xufang Luo and Minghuan Liu and Weiqing Liu and J. Bian and Weinan Zhang and Dongsheng Li},
  booktitle={International Joint Conference on Artificial Intelligence},
  year={2022}
}

@article{Guo2022ImprovingTS,
  title={Improving the Sample Efficiency of Prompt Tuning with Domain Adaptation},
  author={Xu Guo and Boyang Albert Li and Han Yu},
  journal={ArXiv},
  year={2022},
  volume={abs/2210.02952}
}

@inproceedings{Zhang2021OnTC,
  title={On the Convergence and Sample Efficiency of Variance-Reduced Policy Gradient Method},
  author={Junyu Zhang and Chengzhuo Ni and Zheng Yu and Csaba Szepesvari and Mengdi Wang},
  booktitle={Neural Information Processing Systems},
  year={2021}
}

@inproceedings{Yarats2019ImprovingSE,
  title={Improving Sample Efficiency in Model-Free Reinforcement Learning from Images},
  author={Denis Yarats and Amy Zhang and Ilya Kostrikov and Brandon Amos and Joelle Pineau and Rob Fergus},
  booktitle={AAAI Conference on Artificial Intelligence},
  year={2019}
}

@article{Yang2022SampleEO,
  title={Sample Efficiency of Data Augmentation Consistency Regularization},
  author={Shuo Yang and Yijun Dong and Rachel A. Ward and Inderjit S. Dhillon and Sujay Sanghavi and Qi Lei},
  journal={ArXiv},
  year={2022},
  volume={abs/2202.12230}
}

@article{Lagani2021HebbianSL,
  title={Hebbian Semi-Supervised Learning in a Sample Efficiency Setting},
  author={Gabriele Lagani and F. Falchi and Claudio Gennaro and Giuseppe Amato},
  journal={Neural networks : the official journal of the International Neural Network Society},
  year={2021},
  volume={143},
  pages={
          719-731
        }
}

@article{Wei2022ChainOT,
  title={Chain of Thought Prompting Elicits Reasoning in Large Language Models},
  author={Jason Wei and Xuezhi Wang and Dale Schuurmans and Maarten Bosma and Ed Huai-hsin Chi and F. Xia and Quoc Le and Denny Zhou},
  journal={ArXiv},
  year={2022},
  volume={abs/2201.11903}
}

@inproceedings{parmar-etal-2022-boxbart,
    title = "In-{B}o{XBART}: Get Instructions into Biomedical Multi-Task Learning",
    author = "Parmar, Mihir  and
      Mishra, Swaroop  and
      Purohit, Mirali  and
      Luo, Man  and
      Mohammad, Murad  and
      Baral, Chitta",
    booktitle = "Findings of the Association for Computational Linguistics: NAACL 2022",
    month = jul,
    year = "2022",
    address = "Seattle, United States",
    publisher = "Association for Computational Linguistics",
    url = "https://aclanthology.org/2022.findings-naacl.10",
    doi = "10.18653/v1/2022.findings-naacl.10",
    pages = "112--128",
}

@inproceedings{mishra-etal-2022-reframing,
    title = "Reframing Instructional Prompts to {GPT}k{'}s Language",
    author = "Mishra, Swaroop  and
      Khashabi, Daniel  and
      Baral, Chitta  and
      Choi, Yejin  and
      Hajishirzi, Hannaneh",
    booktitle = "Findings of the Association for Computational Linguistics: ACL 2022",
    month = may,
    year = "2022",
    address = "Dublin, Ireland",
    publisher = "Association for Computational Linguistics",
    url = "https://aclanthology.org/2022.findings-acl.50",
    doi = "10.18653/v1/2022.findings-acl.50",
    pages = "589--612",
}

@article{luo2022biotabqa,
  title={BioTABQA: Instruction Learning for Biomedical Table Question Answering},
  author={Luo, Man and Saxena, Sharad and Mishra, Swaroop and Parmar, Mihir and Baral, Chitta},
  journal={arXiv preprint arXiv:2207.02419},
  year={2022}
}

@article{puri2022many,
  title={How Many Data Samples is an Additional Instruction Worth?},
  author={Puri, Ravsehaj Singh and Mishra, Swaroop and Parmar, Mihir and Baral, Chitta},
  journal={arXiv preprint arXiv:2203.09161},
  year={2022}
}

@inproceedings{
wei2022finetuned,
title={Finetuned Language Models are Zero-Shot Learners},
author={Jason Wei and Maarten Bosma and Vincent Zhao and Kelvin Guu and Adams Wei Yu and Brian Lester and Nan Du and Andrew M. Dai and Quoc V Le},
booktitle={International Conference on Learning Representations},
year={2022},
url={https://openreview.net/forum?id=gEZrGCozdqR}
}

@inproceedings{le2021many,
  title={How many data points is a prompt worth?},
  author={Le Scao, Teven and Rush, Alexander M},
  booktitle={Proceedings of the 2021 Conference of the North American Chapter of the Association for Computational Linguistics: Human Language Technologies},
  pages={2627--2636},
  year={2021}
}

@article{nye2021show,
  title={Show your work: Scratchpads for intermediate computation with language models},
  author={Nye, Maxwell and Andreassen, Anders Johan and Gur-Ari, Guy and Michalewski, Henryk and Austin, Jacob and Bieber, David and Dohan, David and Lewkowycz, Aitor and Bosma, Maarten and Luan, David and others},
  journal={arXiv preprint arXiv:2112.00114},
  year={2021}
}

@article{zhou2022least,
  title={Least-to-Most Prompting Enables Complex Reasoning in Large Language Models},
  author={Zhou, Denny and Sch{\"a}rli, Nathanael and Hou, Le and Wei, Jason and Scales, Nathan and Wang, Xuezhi and Schuurmans, Dale and Bousquet, Olivier and Le, Quoc and Chi, Ed},
  journal={arXiv preprint arXiv:2205.10625},
  year={2022}
}

@article{khot2020text,
  title={Text modular networks: Learning to decompose tasks in the language of existing models},
  author={Khot, Tushar and Khashabi, Daniel and Richardson, Kyle and Clark, Peter and Sabharwal, Ashish},
  journal={arXiv preprint arXiv:2009.00751},
  year={2020}
}

@article{patel2022question,
  title={Is a Question Decomposition Unit All We Need?},
  author={Patel, Pruthvi and Mishra, Swaroop and Parmar, Mihir and Baral, Chitta},
  journal={arXiv preprint arXiv:2205.12538},
  year={2022}
}

@article{wang2022instructionner,
  title={InstructionNER: A Multi-Task Instruction-Based Generative Framework for Few-shot NER},
  author={Wang, Liwen and Li, Rumei and Yan, Yang and Yan, Yuanmeng and Wang, Sirui and Wu, Wei and Xu, Weiran},
  journal={arXiv preprint arXiv:2203.03903},
  year={2022}
}

@article{kuznia2022less,
  title={Less is more: Summary of long instructions is better for program synthesis},
  author={Kuznia, Kirby and Mishra, Swaroop and Parmar, Mihir and Baral, Chitta},
  journal={arXiv preprint arXiv:2203.08597},
  year={2022}
}

@article{reif2021recipe,
  title={A recipe for arbitrary text style transfer with large language models},
  author={Reif, Emily and Ippolito, Daphne and Yuan, Ann and Coenen, Andy and Callison-Burch, Chris and Wei, Jason},
  journal={arXiv preprint arXiv:2109.03910},
  year={2021}
}

@article{chen2021adaprompt,
  title={Adaprompt: Adaptive prompt-based finetuning for relation extraction},
  author={Chen, Xiang and Xie, Xin and Zhang, Ningyu and Yan, Jiahuan and Deng, Shumin and Tan, Chuanqi and Huang, Fei and Si, Luo and Chen, Huajun},
  journal={arXiv e-prints},
  pages={arXiv--2104},
  year={2021}
}

\clearpage

%%%%%%%%%%%%%%%%%%%%%%%%%%%%%%%%%%%%%%%%%%%%%%%%%%%%%%%%%%%%

\appendix

\section*{Appendix}

\section{Task Descriptions}

\paragraph{Answerability Classification:} Answerability classification is a natural language processing (NLP) task that involves determining whether a given text contains a question that can be answered. This task can be useful in a variety of applications, such as chatbots or information retrieval systems, where it is important to know whether a user's input is a question that can be answered by the system. To perform answerability classification, an NLP model must first be trained on a dataset of texts labeled as either "answerable" or "unanswerable." The model can then be used to classify new texts as either answerable or unanswerable based on their similarity to the texts in the training dataset. Table \ref{tab:answerability_classification} gives an example of this category.

\paragraph{Cause Effect Classification:} Cause-effect classification is a natural language processing (NLP) task that involves determining the causal relationships between events or actions described in text. This task can be useful in a variety of applications, such as information extraction and text summarization, where it is important to understand the underlying causes and effects of events described in text. Example: Consider the following two sentences: "The car wouldn't start because the battery was dead." "The child was crying because he fell and skinned his knee." In the first sentence, the cause is "the battery was dead," and the effect is "the car wouldn't start." In the second sentence, the cause is "he fell and skinned his knee," and the effect is "the child was crying." 

\paragraph{Coreference Resolution:} Coreference resolution is a natural language processing (NLP) task that involves identifying and linking mentions of the same real-world entities in text. This task is important for understanding the meaning and context of text, as it allows a system to determine that multiple mentions of a word or phrase in a document refer to the same entity. For example, consider the following text: "John went to the store to buy some milk. He needed it for his cereal." In this text, the pronouns "he" and "his" refer to the same person, "John." A coreference resolution system would identify these pronouns as referring to the same entity and link them to the proper noun "John."
Table \ref{tab:coreference_resolution} gives an example of this category.

\paragraph{Data-to-text:} Data-to-text generation is a natural language processing (NLP) task that involves automatically generating human-readable text from structured data. This task can be useful in a variety of applications, such as automated report generation or data summarization, where it is important to present data in a clear and concise manner. An example of data-to-text generation is generating a weather report from data about the current temperature, humidity, and forecast for a particular location. The data might include the following: Temperature: 75 degrees Fahrenheit 
Humidity: 50\%
Forecast: sunny
A data-to-text generation system could use this data to generate the following text: "The current temperature is 75 degrees Fahrenheit and the humidity is 50\%. The forecast for today is sunny." Table \ref{tab:data_to_text} gives an example of this category.

\paragraph{Dialogue Act Recognition:} Dialogue act recognition is a natural language processing (NLP) task that involves identifying the purpose or intention behind a speaker's words in a conversation. This task can be useful in a variety of applications, such as chatbots or virtual assistants, where it is important to understand the intent behind a user's input in order to respond appropriately. An example of dialogue act recognition is identifying the intent behind the following statement: "Can you pass the salt?" The dialogue act in this statement might be classified as a request, as the speaker is asking the listener to perform an action.

\paragraph{Grammar Error Correction:} Grammar error correction is a natural language processing (NLP) task that involves identifying and correcting grammatical errors in a given text. An example of a sentence with a grammatical error that could be corrected as part of this task is: "I went to the stores to buy some food and clothes." This sentence contains the grammatical error of using the wrong form of the word "store." The correct form should be "store," which is singular, as in "I went to the store to buy some food and clothes."

\paragraph{Keyword Tagging:} Keyword tagging is the process of assigning specific keywords or labels to a piece of text or document. This task is often used in natural language processing (NLP) to help classify and organize large amounts of text data for various purposes, such as search engines, topic modeling, and sentiment analysis. For example, consider a news article about recent political events in the United States. Keyword tagging for this article might include labels such as "politics," "US politics," "election," "government," and "political parties." These tags can help identify the main themes and topics discussed in the article, making it easier for users to search for and find similar articles on the same topics.

\paragraph{Overlap Extraction:} Overlap extraction is a natural language processing (NLP) task that involves extracting overlapping text or data from multiple sources. This can be useful for a variety of purposes, such as identifying common themes in different documents, comparing and contrast information, or finding duplicates in a dataset. For example, consider a scenario where you have two news articles discussing the same topic. You might use overlap extraction to identify the common themes or ideas discussed in both articles, such as the main events, people involved, or key quotes. This could help you understand the overall coverage of the topic and identify any discrepancies or differences in the way it was presented by the two sources.

\paragraph{Question Rewriting:} Question Rewriting is a natural language processing (NLP) task that involves generating a new version of a given question that has the same meaning as the original, but is phrased differently. 
For example, given the question "What is the capital of France?", a question rewriting task might generate the following rephrased question: "Where is the seat of government for France located?". 
Table \ref{tab:question_rewriting} gives an example of this category.

\paragraph{Textual Entailment:} Textual entailment is a natural language processing task that involves determining the relationship between two text passages. Specifically, it involves determining whether one passage, called the "hypothesis," can be inferred from the other passage, called the "premise."

For example:

Premise: "The cat is sitting on the couch."
Hypothesis: "There is a cat on the couch."

In this case, the hypothesis can be inferred from the premise, so the textual entailment relationship is "entailment."
Table \ref{tab:textual_entailment} gives an example of this category.

\paragraph{Title Generation:} Title generation is a natural language processing (NLP) task that involves creating a title for a given text or topic. This task is often used in content creation and marketing, where an eye-catching title is essential for attracting attention and engaging readers. For example, a title generation task might involve creating a title for an article about the benefits of meditation. Some possible titles might be "5 Reasons Why Meditation is the Key to a Stress-Free Life," "Discover the Surprising Benefits of Meditation," or "Meditation: The Ultimate Tool for Relaxation and Mindfulness." The goal of the title generation task is to generate a title that accurately reflects the content of the article and is compelling enough to encourage readers to click and read more. Table \ref{tab:title_generation} gives an example of this category.

\paragraph{Word Analogy:} Word analogy is a natural language processing task that involves identifying relationships between words based on their meanings and contexts. The goal is to find a word that is similar to another word in a specific way, based on the relationship between the two words. For example, if the task is to find a word that is similar to "man" in the same way that "woman" is similar to "man," the correct answer would be "wife." The relationship between the words "man" and "wife" is that they are both terms for a specific type of spouse, with "man" being the term for a husband and "wife" being the term for a wife.

\section{Detailed Results}
\label{detailed_results}

\paragraph{Hyperparameters:} Train batch size: 1, Eval Batch size: 1, Gradient Accumulation Steps: 2, Max source length: 1024, Max target length 128, generation max length: 128, learning rate: 5e-05, number of epochs: 2, warmup steps: 0 

\paragraph{Other results and datasets:} Table 6 gives the task-wise statistics, the number of total samples, and their category. Table 5 is the pivot table of the same, giving a category-wise count of the tasks. Table 7 gives the total number of samples used for different baselines and STL MTL methods. Table 8 gives STL baseline scores. Table 9 gives STL Model results. Table 12 gives a category-wise average for the same. 
Table 10 gives MTL baselines, and 11 gives the MTL results, and Table 13 gives category-wise pivot for the same.

\begin{table*}[t!]
    \centering
    \small
    \resizebox{\linewidth}{!}
    {
        \begin{tabular}{lll|lll}
\hline
\textbf{\begin{tabular}[c]{@{}l@{}}Task\\ Number\end{tabular}} & \textbf{\begin{tabular}[c]{@{}l@{}}Task \\ Category\end{tabular}} & \textbf{\begin{tabular}[c]{@{}l@{}}Train\\ Samples\end{tabular}} & \textbf{\begin{tabular}[c]{@{}l@{}}Task\\ Number\end{tabular}} & \textbf{\begin{tabular}[c]{@{}l@{}}Task \\ Category\end{tabular}} & \textbf{\begin{tabular}[c]{@{}l@{}}Train\\ Samples\end{tabular}} \\ \hline
task020                                                        & Answerability Classification                                      & 271                                                              & task201                                                        & Textual Entailment                                                & 6390                                                             \\
task033                                                        & Coreference Resolution                                            & 6390                                                             & task202                                                        & Textual Entailment                                                & 6388                                                             \\
task034                                                        & Question Rewriting                                                & 6390                                                             & task219                                                        & Title Generation                                                  & 6390                                                             \\
task035                                                        & Question Rewriting                                                & 6394                                                             & task220                                                        & Title Generation                                                  & 6389                                                             \\
task036                                                        & Keyword Tagging                                                   & 812                                                              & task226                                                        & Answerability Classification                                      & 368                                                              \\
task039                                                        & Overlap Extraction                                                & 6390                                                             & task232                                                        & Answerability Classification                                      & 6390                                                             \\
task050                                                        & Answerability Classification                                      & 5802                                                             & task233                                                        & Answerability Classification                                      & 6389                                                             \\
task102                                                        & Data to Text                                                      & 5298                                                             & task242                                                        & Answerability Classification                                      & 5884                                                             \\
task1152                                                       & Word Analogy                                                      & 94                                                               & task249                                                        & Coreference Resolution                                            & 574                                                              \\
task1153                                                       & Word Analogy                                                      & 1908                                                             & task281                                                        & Overlap Extraction                                                & 1365                                                             \\
task1154                                                       & Word Analogy                                                      & 694                                                              & task288                                                        & Title Generation                                                  & 1823                                                             \\
task1155                                                       & Word Analogy                                                      & 436                                                              & task290                                                        & Answerability Classification                                      & 3372                                                             \\
task1156                                                       & Word Analogy                                                      & 548                                                              & task304                                                        & Coreference Resolution                                            & 6389                                                             \\
task1157                                                       & Word Analogy                                                      & 857                                                              & task329                                                        & Coreference Resolution                                            & 4339                                                             \\
task1158                                                       & Word Analogy                                                      & 266                                                              & task330                                                        & Coreference Resolution                                            & 3849                                                             \\
task1159                                                       & Word Analogy                                                      & 588                                                              & task349                                                        & Answerability Classification                                      & 6388                                                             \\
task1161                                                       & Title Generation                                                  & 6390                                                             & task362                                                        & Dialogue Act Recognition                                          & 6285                                                             \\
task1195                                                       & Question Rewriting                                                & 6384                                                             & task391                                                        & Cause Effect Classification                                       & 2173                                                             \\
task121                                                        & Question Rewriting                                                & 17                                                               & task392                                                        & Cause Effect Classification                                       & 2493                                                             \\
task133                                                        & Coreference Resolution                                            & 2745                                                             & task393                                                        & Cause Effect Classification                                       & 100                                                              \\
task1342                                                       & Title Generation                                                  & 6107                                                             & task401                                                        & Coreference Resolution                                            & 3636                                                             \\
task1344                                                       & Textual Entailment                                                & 2370                                                             & task402                                                        & Question Rewriting                                                & 2935                                                             \\
task1345                                                       & Question Rewriting                                                & 6390                                                             & task418                                                        & Title Generation                                                  & 3229                                                             \\
task1356                                                       & Title Generation                                                  & 6017                                                             & task442                                                        & Question Rewriting                                                & 1686                                                             \\
task1358                                                       & Title Generation                                                  & 6017                                                             & task500                                                        & Title Generation                                                  & 6390                                                             \\
task1385                                                       & Textual Entailment                                                & 880                                                              & task510                                                        & Title Generation                                                  & 6389                                                             \\
task1386                                                       & Textual Entailment                                                & 883                                                              & task520                                                        & Answerability Classification                                      & 890                                                              \\
task1387                                                       & Textual Entailment                                                & 1083                                                             & task569                                                        & Title Generation                                                  & 6379                                                             \\
task1388                                                       & Textual Entailment                                                & 191                                                              & task602                                                        & Title Generation                                                  & 84                                                               \\
task1390                                                       & Coreference Resolution                                            & 538                                                              & task613                                                        & Keyword Tagging                                                   & 6390                                                             \\
task1391                                                       & Coreference Resolution                                            & 6381                                                             & task614                                                        & Cause Effect Classification                                       & 6387                                                             \\
task1393                                                       & Cause Effect Classification                                       & 386                                                              & task619                                                        & Title Generation                                                  & 1149                                                             \\
task1394                                                       & Dialogue Act Recognition                                          & 114                                                              & task620                                                        & Keyword Tagging                                                   & 1151                                                             \\
task1407                                                       & Data to Text                                                      & 3019                                                             & task623                                                        & Keyword Tagging                                                   & 290                                                              \\
task1409                                                       & Data to Text                                                      & 6372                                                             & task640                                                        & Textual Entailment                                                & 100                                                              \\
task1439                                                       & Answerability Classification                                      & 2420                                                             & task641                                                        & Textual Entailment                                                & 88                                                               \\
task1442                                                       & Answerability Classification                                      & 1769                                                             & task642                                                        & Textual Entailment                                                & 190                                                              \\
task1516                                                       & Textual Entailment                                                & 599                                                              & task645                                                        & Keyword Tagging                                                   & 1890                                                             \\
task1529                                                       & Textual Entailment                                                & 4950                                                             & task648                                                        & Coreference Resolution                                            & 171                                                              \\
task1531                                                       & Dialogue Act Recognition                                          & 387                                                              & task670                                                        & Question Rewriting                                                & 4639                                                             \\
task1533                                                       & Dialogue Act Recognition                                          & 6052                                                             & task671                                                        & Question Rewriting                                                & 4639                                                             \\
task1534                                                       & Dialogue Act Recognition                                          & 6051                                                             & task677                                                        & Data to Text                                                      & 4400                                                             \\
task1540                                                       & Title Generation                                                  & 2895                                                             & task738                                                        & Textual Entailment                                                & 6291                                                             \\
task1554                                                       & Textual Entailment                                                & 6385                                                             & task743                                                        & Title Generation                                                  & 540                                                              \\
task1557                                                       & Grammar Error Correction                                          & 644                                                              & task760                                                        & Data to Text                                                      & 26                                                               \\
task1562                                                       & Question Rewriting                                                & 32                                                               & task769                                                        & Title Generation                                                  & 890                                                              \\
task1586                                                       & Title Generation                                                  & 4890                                                             & task827                                                        & Cause Effect Classification                                       & 886                                                              \\
task1598                                                       & Data to Text                                                      & 6384                                                             & task828                                                        & Cause Effect Classification                                       & 886                                                              \\
task1612                                                       & Textual Entailment                                                & 1690                                                             & task879                                                        & Dialogue Act Recognition                                          & 2195                                                             \\
task1615                                                       & Textual Entailment                                                & 1687                                                             & task880                                                        & Dialogue Act Recognition                                          & 1231                                                             \\
task1622                                                       & Question Rewriting                                                & 1885                                                             & task890                                                        & Textual Entailment                                                & 88                                                               \\
task1624                                                       & Answerability Classification                                      & 1390                                                             & task891                                                        & Coreference Resolution                                            & 87                                                               \\
task1631                                                       & Data to Text                                                      & 2880                                                             & task892                                                        & Coreference Resolution                                            & 83                                                               \\
task1640                                                       & Answerability Classification                                      & 2538                                                             & task893                                                        & Coreference Resolution                                            & 100                                                              \\
task1659                                                       & Title Generation                                                  & 6389                                                             & task935                                                        & Textual Entailment                                                & 6390                                                             \\
task1664                                                       & Coreference Resolution                                            & 278                                                              & task936                                                        & Textual Entailment                                                & 6390                                                             \\
task1728                                                       & Data to Text                                                      & 6385                                                             & task937                                                        & Textual Entailment                                                & 6389                                                             \\
task190                                                        & Textual Entailment                                                & 6390                                                             & task957                                                        & Data to Text                                                      & 2051                                                             \\
task199                                                        & Textual Entailment                                                & 6389                                                             & task970                                                        & Textual Entailment                                                & 2247                                                             \\
task200                                                        & Textual Entailment                                                & 4354                                                             &                                                                &                                                                   &                                                                  \\ \hline
\end{tabular}
    }
    \caption{Dataset Statistics which tell the max number of train samples available to each dataset present. }
    \label{tab:dataset_stats}
\end{table*}

\begin{table*}[t!]
    \centering
    \small
    \resizebox{\linewidth}{!}
    {
        \begin{tabular}{llllll|llllll}
\hline
\textbf{\begin{tabular}[c]{@{}l@{}}Task\\ Number\end{tabular}} & \textbf{\begin{tabular}[c]{@{}l@{}}Train \\ Samples \\ in 10\end{tabular}} & \textbf{\begin{tabular}[c]{@{}l@{}}Train \\ Samples \\ in 100\end{tabular}} & \textbf{\begin{tabular}[c]{@{}l@{}}Train \\ Samples \\ in 200\\ \\ STL Setup \\ Baseline 1\\ \\ MLT Setup\\ Baseline 2\end{tabular}} & \textbf{\begin{tabular}[c]{@{}l@{}}Train \\ Samples \\ in 1000\\ \\ STL Setup\\ Baseline 2\\ \\ MLT Setup\\ Baseline 1\end{tabular}} & \textbf{\begin{tabular}[c]{@{}l@{}}Train \\ Samples \\ in All\\ \\ STL Setup\\ Baseline 3\end{tabular}} & \textbf{\begin{tabular}[c]{@{}l@{}}Task\\ Number\end{tabular}} & \textbf{\begin{tabular}[c]{@{}l@{}}Train \\ Samples \\ in 10\end{tabular}} & \textbf{\begin{tabular}[c]{@{}l@{}}Train \\ Samples \\ in 100\end{tabular}} & \textbf{\begin{tabular}[c]{@{}l@{}}Train \\ Samples \\ in 200\\ \\ STL Setup \\ Baseline 1\\ \\ MLT Setup\\ Baseline 2\end{tabular}} & \textbf{\begin{tabular}[c]{@{}l@{}}Train \\ Samples \\ in 1000\\ \\ STL Setup\\ Baseline 2\\ \\ MLT Setup\\ Baseline 1\end{tabular}} & \textbf{\begin{tabular}[c]{@{}l@{}}Train \\ Samples \\ in All\\ \\ STL Setup\\ Baseline 3\end{tabular}} \\ \hline
task020                                                        & 10                                                                         & 100                                                                         & 200                                                                                                                                  & 271                                                                                                                                  & 271                                                                                                     & task201                                                        & 10                                                                         & 100                                                                         & 200                                                                                                                                  & 1000                                                                                                                                 & 6390                                                                                                    \\
task033                                                        & 10                                                                         & 100                                                                         & 200                                                                                                                                  & 1000                                                                                                                                 & 6390                                                                                                    & task202                                                        & 10                                                                         & 100                                                                         & 200                                                                                                                                  & 1000                                                                                                                                 & 6388                                                                                                    \\
task034                                                        & 10                                                                         & 100                                                                         & 200                                                                                                                                  & 1000                                                                                                                                 & 6390                                                                                                    & task219                                                        & 10                                                                         & 100                                                                         & 200                                                                                                                                  & 1000                                                                                                                                 & 6390                                                                                                    \\
task035                                                        & 10                                                                         & 100                                                                         & 200                                                                                                                                  & 1000                                                                                                                                 & 6394                                                                                                    & task220                                                        & 10                                                                         & 100                                                                         & 200                                                                                                                                  & 1000                                                                                                                                 & 6389                                                                                                    \\
task036                                                        & 10                                                                         & 100                                                                         & 200                                                                                                                                  & 812                                                                                                                                  & 812                                                                                                     & task226                                                        & 10                                                                         & 100                                                                         & 200                                                                                                                                  & 368                                                                                                                                  & 368                                                                                                     \\
task039                                                        & 10                                                                         & 100                                                                         & 200                                                                                                                                  & 1000                                                                                                                                 & 6390                                                                                                    & task232                                                        & 10                                                                         & 100                                                                         & 200                                                                                                                                  & 1000                                                                                                                                 & 6390                                                                                                    \\
task050                                                        & 10                                                                         & 100                                                                         & 200                                                                                                                                  & 1000                                                                                                                                 & 5802                                                                                                    & task233                                                        & 10                                                                         & 100                                                                         & 200                                                                                                                                  & 1000                                                                                                                                 & 6389                                                                                                    \\
task102                                                        & 10                                                                         & 100                                                                         & 200                                                                                                                                  & 1000                                                                                                                                 & 5298                                                                                                    & task242                                                        & 10                                                                         & 100                                                                         & 200                                                                                                                                  & 1000                                                                                                                                 & 5884                                                                                                    \\
task1152                                                       & 10                                                                         & 94                                                                          & 94                                                                                                                                   & 94                                                                                                                                   & 94                                                                                                      & task249                                                        & 10                                                                         & 100                                                                         & 200                                                                                                                                  & 574                                                                                                                                  & 574                                                                                                     \\
task1153                                                       & 10                                                                         & 100                                                                         & 200                                                                                                                                  & 1000                                                                                                                                 & 1908                                                                                                    & task281                                                        & 10                                                                         & 100                                                                         & 200                                                                                                                                  & 1000                                                                                                                                 & 1365                                                                                                    \\
task1154                                                       & 10                                                                         & 100                                                                         & 200                                                                                                                                  & 694                                                                                                                                  & 694                                                                                                     & task288                                                        & 10                                                                         & 100                                                                         & 200                                                                                                                                  & 1000                                                                                                                                 & 1823                                                                                                    \\
task1155                                                       & 10                                                                         & 100                                                                         & 200                                                                                                                                  & 436                                                                                                                                  & 436                                                                                                     & task290                                                        & 10                                                                         & 100                                                                         & 200                                                                                                                                  & 1000                                                                                                                                 & 3372                                                                                                    \\
task1156                                                       & 10                                                                         & 100                                                                         & 200                                                                                                                                  & 548                                                                                                                                  & 548                                                                                                     & task304                                                        & 10                                                                         & 100                                                                         & 200                                                                                                                                  & 1000                                                                                                                                 & 6389                                                                                                    \\
task1157                                                       & 10                                                                         & 100                                                                         & 200                                                                                                                                  & 857                                                                                                                                  & 857                                                                                                     & task329                                                        & 10                                                                         & 100                                                                         & 200                                                                                                                                  & 1000                                                                                                                                 & 4339                                                                                                    \\
task1158                                                       & 10                                                                         & 100                                                                         & 200                                                                                                                                  & 266                                                                                                                                  & 266                                                                                                     & task330                                                        & 10                                                                         & 100                                                                         & 200                                                                                                                                  & 1000                                                                                                                                 & 3849                                                                                                    \\
task1159                                                       & 10                                                                         & 100                                                                         & 200                                                                                                                                  & 588                                                                                                                                  & 588                                                                                                     & task349                                                        & 10                                                                         & 100                                                                         & 200                                                                                                                                  & 1000                                                                                                                                 & 6388                                                                                                    \\
task1161                                                       & 10                                                                         & 100                                                                         & 200                                                                                                                                  & 1000                                                                                                                                 & 6390                                                                                                    & task362                                                        & 10                                                                         & 100                                                                         & 200                                                                                                                                  & 1000                                                                                                                                 & 6285                                                                                                    \\
task1195                                                       & 10                                                                         & 100                                                                         & 200                                                                                                                                  & 1000                                                                                                                                 & 6384                                                                                                    & task391                                                        & 10                                                                         & 100                                                                         & 200                                                                                                                                  & 1000                                                                                                                                 & 2173                                                                                                    \\
task121                                                        & 10                                                                         & 17                                                                          & 17                                                                                                                                   & 17                                                                                                                                   & 17                                                                                                      & task392                                                        & 10                                                                         & 100                                                                         & 200                                                                                                                                  & 1000                                                                                                                                 & 2493                                                                                                    \\
task133                                                        & 10                                                                         & 100                                                                         & 200                                                                                                                                  & 1000                                                                                                                                 & 2745                                                                                                    & task393                                                        & 10                                                                         & 100                                                                         & 100                                                                                                                                  & 100                                                                                                                                  & 100                                                                                                     \\
task1342                                                       & 10                                                                         & 100                                                                         & 200                                                                                                                                  & 1000                                                                                                                                 & 6107                                                                                                    & task401                                                        & 10                                                                         & 100                                                                         & 200                                                                                                                                  & 1000                                                                                                                                 & 3636                                                                                                    \\
task1344                                                       & 10                                                                         & 100                                                                         & 200                                                                                                                                  & 1000                                                                                                                                 & 2370                                                                                                    & task402                                                        & 10                                                                         & 100                                                                         & 200                                                                                                                                  & 1000                                                                                                                                 & 2935                                                                                                    \\
task1345                                                       & 10                                                                         & 100                                                                         & 200                                                                                                                                  & 1000                                                                                                                                 & 6390                                                                                                    & task418                                                        & 10                                                                         & 100                                                                         & 200                                                                                                                                  & 1000                                                                                                                                 & 3229                                                                                                    \\
task1356                                                       & 10                                                                         & 100                                                                         & 200                                                                                                                                  & 1000                                                                                                                                 & 6017                                                                                                    & task442                                                        & 10                                                                         & 100                                                                         & 200                                                                                                                                  & 1000                                                                                                                                 & 1686                                                                                                    \\
task1358                                                       & 10                                                                         & 100                                                                         & 200                                                                                                                                  & 1000                                                                                                                                 & 6017                                                                                                    & task500                                                        & 10                                                                         & 100                                                                         & 200                                                                                                                                  & 1000                                                                                                                                 & 6390                                                                                                    \\
task1385                                                       & 10                                                                         & 100                                                                         & 200                                                                                                                                  & 880                                                                                                                                  & 880                                                                                                     & task510                                                        & 10                                                                         & 100                                                                         & 200                                                                                                                                  & 1000                                                                                                                                 & 6389                                                                                                    \\
task1386                                                       & 10                                                                         & 100                                                                         & 200                                                                                                                                  & 883                                                                                                                                  & 883                                                                                                     & task520                                                        & 10                                                                         & 100                                                                         & 200                                                                                                                                  & 890                                                                                                                                  & 890                                                                                                     \\
task1387                                                       & 10                                                                         & 100                                                                         & 200                                                                                                                                  & 1000                                                                                                                                 & 1083                                                                                                    & task569                                                        & 10                                                                         & 100                                                                         & 200                                                                                                                                  & 1000                                                                                                                                 & 6379                                                                                                    \\
task1388                                                       & 10                                                                         & 100                                                                         & 191                                                                                                                                  & 191                                                                                                                                  & 191                                                                                                     & task602                                                        & 10                                                                         & 84                                                                          & 84                                                                                                                                   & 84                                                                                                                                   & 84                                                                                                      \\
task1390                                                       & 10                                                                         & 100                                                                         & 200                                                                                                                                  & 538                                                                                                                                  & 538                                                                                                     & task613                                                        & 10                                                                         & 100                                                                         & 200                                                                                                                                  & 1000                                                                                                                                 & 6390                                                                                                    \\
task1391                                                       & 10                                                                         & 100                                                                         & 200                                                                                                                                  & 1000                                                                                                                                 & 6381                                                                                                    & task614                                                        & 10                                                                         & 100                                                                         & 200                                                                                                                                  & 1000                                                                                                                                 & 6387                                                                                                    \\
task1393                                                       & 10                                                                         & 100                                                                         & 200                                                                                                                                  & 386                                                                                                                                  & 386                                                                                                     & task619                                                        & 10                                                                         & 100                                                                         & 200                                                                                                                                  & 1000                                                                                                                                 & 1149                                                                                                    \\
task1394                                                       & 10                                                                         & 100                                                                         & 114                                                                                                                                  & 114                                                                                                                                  & 114                                                                                                     & task620                                                        & 10                                                                         & 100                                                                         & 200                                                                                                                                  & 1000                                                                                                                                 & 1151                                                                                                    \\
task1407                                                       & 10                                                                         & 100                                                                         & 200                                                                                                                                  & 1000                                                                                                                                 & 3019                                                                                                    & task623                                                        & 10                                                                         & 100                                                                         & 200                                                                                                                                  & 290                                                                                                                                  & 290                                                                                                     \\
task1409                                                       & 10                                                                         & 100                                                                         & 200                                                                                                                                  & 1000                                                                                                                                 & 6372                                                                                                    & task640                                                        & 10                                                                         & 100                                                                         & 100                                                                                                                                  & 100                                                                                                                                  & 100                                                                                                     \\
task1439                                                       & 10                                                                         & 100                                                                         & 200                                                                                                                                  & 1000                                                                                                                                 & 2420                                                                                                    & task641                                                        & 10                                                                         & 88                                                                          & 88                                                                                                                                   & 88                                                                                                                                   & 88                                                                                                      \\
task1442                                                       & 10                                                                         & 100                                                                         & 200                                                                                                                                  & 1000                                                                                                                                 & 1769                                                                                                    & task642                                                        & 10                                                                         & 100                                                                         & 190                                                                                                                                  & 190                                                                                                                                  & 190                                                                                                     \\
task1516                                                       & 10                                                                         & 100                                                                         & 200                                                                                                                                  & 599                                                                                                                                  & 599                                                                                                     & task645                                                        & 10                                                                         & 100                                                                         & 200                                                                                                                                  & 1000                                                                                                                                 & 1890                                                                                                    \\
task1529                                                       & 10                                                                         & 100                                                                         & 200                                                                                                                                  & 1000                                                                                                                                 & 4950                                                                                                    & task648                                                        & 10                                                                         & 100                                                                         & 171                                                                                                                                  & 171                                                                                                                                  & 171                                                                                                     \\
task1531                                                       & 10                                                                         & 100                                                                         & 200                                                                                                                                  & 387                                                                                                                                  & 387                                                                                                     & task670                                                        & 10                                                                         & 100                                                                         & 200                                                                                                                                  & 1000                                                                                                                                 & 4639                                                                                                    \\
task1533                                                       & 10                                                                         & 100                                                                         & 200                                                                                                                                  & 1000                                                                                                                                 & 6052                                                                                                    & task671                                                        & 10                                                                         & 100                                                                         & 200                                                                                                                                  & 1000                                                                                                                                 & 4639                                                                                                    \\
task1534                                                       & 10                                                                         & 100                                                                         & 200                                                                                                                                  & 1000                                                                                                                                 & 6051                                                                                                    & task677                                                        & 10                                                                         & 100                                                                         & 200                                                                                                                                  & 1000                                                                                                                                 & 4400                                                                                                    \\
task1540                                                       & 10                                                                         & 100                                                                         & 200                                                                                                                                  & 1000                                                                                                                                 & 2895                                                                                                    & task738                                                        & 10                                                                         & 100                                                                         & 200                                                                                                                                  & 1000                                                                                                                                 & 6291                                                                                                    \\
task1554                                                       & 10                                                                         & 100                                                                         & 200                                                                                                                                  & 1000                                                                                                                                 & 6385                                                                                                    & task743                                                        & 10                                                                         & 100                                                                         & 200                                                                                                                                  & 540                                                                                                                                  & 540                                                                                                     \\
task1557                                                       & 10                                                                         & 100                                                                         & 200                                                                                                                                  & 644                                                                                                                                  & 644                                                                                                     & task760                                                        & 10                                                                         & 26                                                                          & 26                                                                                                                                   & 26                                                                                                                                   & 26                                                                                                      \\
task1562                                                       & 10                                                                         & 32                                                                          & 32                                                                                                                                   & 32                                                                                                                                   & 32                                                                                                      & task769                                                        & 10                                                                         & 100                                                                         & 200                                                                                                                                  & 890                                                                                                                                  & 890                                                                                                     \\
task1586                                                       & 10                                                                         & 100                                                                         & 200                                                                                                                                  & 1000                                                                                                                                 & 4890                                                                                                    & task827                                                        & 10                                                                         & 100                                                                         & 200                                                                                                                                  & 886                                                                                                                                  & 886                                                                                                     \\
task1598                                                       & 10                                                                         & 100                                                                         & 200                                                                                                                                  & 1000                                                                                                                                 & 6384                                                                                                    & task828                                                        & 10                                                                         & 100                                                                         & 200                                                                                                                                  & 886                                                                                                                                  & 886                                                                                                     \\
task1612                                                       & 10                                                                         & 100                                                                         & 200                                                                                                                                  & 1000                                                                                                                                 & 1690                                                                                                    & task879                                                        & 10                                                                         & 100                                                                         & 200                                                                                                                                  & 1000                                                                                                                                 & 2195                                                                                                    \\
task1615                                                       & 10                                                                         & 100                                                                         & 200                                                                                                                                  & 1000                                                                                                                                 & 1687                                                                                                    & task880                                                        & 10                                                                         & 100                                                                         & 200                                                                                                                                  & 1000                                                                                                                                 & 1231                                                                                                    \\
task1622                                                       & 10                                                                         & 100                                                                         & 200                                                                                                                                  & 1000                                                                                                                                 & 1885                                                                                                    & task890                                                        & 10                                                                         & 88                                                                          & 88                                                                                                                                   & 88                                                                                                                                   & 88                                                                                                      \\
task1624                                                       & 10                                                                         & 100                                                                         & 200                                                                                                                                  & 1000                                                                                                                                 & 1390                                                                                                    & task891                                                        & 10                                                                         & 87                                                                          & 87                                                                                                                                   & 87                                                                                                                                   & 87                                                                                                      \\
task1631                                                       & 10                                                                         & 100                                                                         & 200                                                                                                                                  & 1000                                                                                                                                 & 2880                                                                                                    & task892                                                        & 10                                                                         & 83                                                                          & 83                                                                                                                                   & 83                                                                                                                                   & 83                                                                                                      \\
task1640                                                       & 10                                                                         & 100                                                                         & 200                                                                                                                                  & 1000                                                                                                                                 & 2538                                                                                                    & task893                                                        & 10                                                                         & 100                                                                         & 100                                                                                                                                  & 100                                                                                                                                  & 100                                                                                                     \\
task1659                                                       & 10                                                                         & 100                                                                         & 200                                                                                                                                  & 1000                                                                                                                                 & 6389                                                                                                    & task935                                                        & 10                                                                         & 100                                                                         & 200                                                                                                                                  & 1000                                                                                                                                 & 6390                                                                                                    \\
task1664                                                       & 10                                                                         & 100                                                                         & 200                                                                                                                                  & 278                                                                                                                                  & 278                                                                                                     & task936                                                        & 10                                                                         & 100                                                                         & 200                                                                                                                                  & 1000                                                                                                                                 & 6390                                                                                                    \\
task1728                                                       & 10                                                                         & 100                                                                         & 200                                                                                                                                  & 1000                                                                                                                                 & 6385                                                                                                    & task937                                                        & 10                                                                         & 100                                                                         & 200                                                                                                                                  & 1000                                                                                                                                 & 6389                                                                                                    \\
task190                                                        & 10                                                                         & 100                                                                         & 200                                                                                                                                  & 1000                                                                                                                                 & 6390                                                                                                    & task957                                                        & 10                                                                         & 100                                                                         & 200                                                                                                                                  & 1000                                                                                                                                 & 2051                                                                                                    \\
task199                                                        & 10                                                                         & 100                                                                         & 200                                                                                                                                  & 1000                                                                                                                                 & 6389                                                                                                    & task970                                                        & 10                                                                         & 100                                                                         & 200                                                                                                                                  & 1000                                                                                                                                 & 2247                                                                                                    \\
task200                                                        & 10                                                                         & 100                                                                         & 200                                                                                                                                  & 1000                                                                                                                                 & 4354                                                                                                    &                                                                &                                                                            &                                                                             &                                                                                                                                      &                                                                                                                                      &                                                                                                         \\ \hline
\end{tabular}
    }
    \caption{Training sample statistics that showcase a number of samples chosen when training with a certain percentage of the data. There are multiple names in a single column highlighting that all of those setups/baselines used that number of samples.}
    \label{tab:train_statistics}
\end{table*}

\begin{table*}[t!]
    \centering
    \small
    \resizebox{\linewidth}{!}
    {
        \begin{tabular}{llll|llll}
\hline
\textbf{\begin{tabular}[c]{@{}l@{}}Task\\ Number\end{tabular}} & \textbf{Baseline 1} & \textbf{Baseline 2} & \textbf{Baseline 3} & \textbf{\begin{tabular}[c]{@{}l@{}}Task\\ Number\end{tabular}} & \textbf{Baseline 1} & \textbf{Baseline 2} & \textbf{Baseline 3} \\ \hline
task020                                                        & 0.50                & 0.50                & 0.50                & task201                                                        & 0.56                & 0.88                & 0.93                \\
task033                                                        & 0.69                & 0.73                & 0.81                & task202                                                        & 0.91                & 0.97                & 0.96                \\
task034                                                        & 0.94                & 0.94                & 0.95                & task219                                                        & 0.34                & 0.37                & 0.36                \\
task035                                                        & 0.92                & 0.91                & 0.92                & task220                                                        & 1.00                & 1.00                & 1.00                \\
task036                                                        & 0.33                & 0.38                & 0.38                & task226                                                        & 0.56                & 0.59                & 0.59                \\
task039                                                        & 0.65                & 0.67                & 0.76                & task232                                                        & 0.50                & 0.83                & 0.90                \\
task050                                                        & 0.82                & 0.80                & 0.82                & task233                                                        & 0.50                & 0.60                & 0.70                \\
task102                                                        & 0.65                & 0.67                & 0.68                & task242                                                        & 0.99                & 0.99                & 0.99                \\
task1152                                                       & 0.80                & 0.02                & 0.02                & task249                                                        & 0.64                & 0.53                & 0.53                \\
task1153                                                       & 0.95                & 1.00                & 1.00                & task281                                                        & 0.63                & 0.69                & 0.68                \\
task1154                                                       & 1.00                & 1.00                & 1.00                & task288                                                        & 0.34                & 0.36                & 0.35                \\
task1155                                                       & 1.00                & 1.00                & 1.00                & task290                                                        & 0.84                & 0.94                & 0.91                \\
task1156                                                       & 1.00                & 1.00                & 1.00                & task304                                                        & 0.38                & 0.77                & 0.79                \\
task1157                                                       & 0.99                & 1.00                & 1.00                & task329                                                        & 0.73                & 0.79                & 0.90                \\
task1158                                                       & 1.00                & 1.00                & 1.00                & task330                                                        & 0.85                & 0.89                & 0.91                \\
task1159                                                       & 0.99                & 1.00                & 1.00                & task349                                                        & 0.76                & 0.81                & 0.92                \\
task1161                                                       & 0.38                & 0.38                & 0.41                & task362                                                        & 0.92                & 0.94                & 0.95                \\
task1195                                                       & 0.92                & 0.97                & 0.98                & task391                                                        & 0.86                & 0.94                & 0.93                \\
task121                                                        & 0.22                & 0.50                & 0.50                & task392                                                        & 0.86                & 0.93                & 0.93                \\
task133                                                        & 0.72                & 0.76                & 0.83                & task393                                                        & 0.31                & 0.21                & 0.21                \\
task1342                                                       & 0.18                & 0.17                & 0.19                & task401                                                        & 0.85                & 0.84                & 0.92                \\
task1344                                                       & 0.87                & 0.90                & 0.93                & task402                                                        & 0.56                & 0.57                & 0.54                \\
task1345                                                       & 0.40                & 0.39                & 0.40                & task418                                                        & 0.30                & 0.34                & 0.34                \\
task1356                                                       & 0.31                & 0.31                & 0.33                & task442                                                        & 0.66                & 0.67                & 0.68                \\
task1358                                                       & 0.35                & 0.38                & 0.39                & task500                                                        & 0.41                & 0.41                & 0.46                \\
task1385                                                       & 0.45                & 0.73                & 0.73                & task510                                                        & 0.49                & 0.49                & 0.52                \\
task1386                                                       & 0.35                & 0.54                & 0.54                & task520                                                        & 1.00                & 1.00                & 1.00                \\
task1387                                                       & 0.34                & 0.48                & 0.52                & task569                                                        & 0.46                & 0.45                & 0.47                \\
task1388                                                       & 0.78                & 0.67                & 0.67                & task602                                                        & 0.31                & 0.39                & 0.39                \\
task1390                                                       & 0.49                & 0.50                & 0.50                & task613                                                        & 0.18                & 0.26                & 0.24                \\
task1391                                                       & 0.77                & 0.50                & 0.92                & task614                                                        & 0.61                & 0.60                & 0.64                \\
task1393                                                       & 0.85                & 0.77                & 0.77                & task619                                                        & 0.45                & 0.44                & 0.47                \\
task1394                                                       & 0.54                & 0.46                & 0.46                & task620                                                        & 0.01                & 0.05                & 0.04                \\
task1407                                                       & 0.43                & 0.45                & 0.48                & task623                                                        & 0.98                & 1.00                & 1.00                \\
task1409                                                       & 0.51                & 0.52                & 0.55                & task640                                                        & 0.89                & 0.86                & 0.86                \\
task1439                                                       & 0.75                & 0.76                & 0.71                & task641                                                        & 0.36                & 0.37                & 0.37                \\
task1442                                                       & 0.62                & 0.65                & 0.58                & task642                                                        & 0.50                & 0.50                & 0.50                \\
task1516                                                       & 1.00                & 1.00                & 1.00                & task645                                                        & 0.96                & 0.97                & 0.97                \\
task1529                                                       & 0.89                & 0.96                & 0.97                & task648                                                        & 0.67                & 0.69                & 0.69                \\
task1531                                                       & 0.60                & 0.58                & 0.58                & task670                                                        & 0.78                & 0.75                & 0.75                \\
task1533                                                       & 0.50                & 0.64                & 0.79                & task671                                                        & 0.65                & 0.63                & 0.63                \\
task1534                                                       & 0.51                & 0.97                & 0.99                & task677                                                        & 0.33                & 0.31                & 0.34                \\
task1540                                                       & 0.39                & 0.42                & 0.43                & task738                                                        & 0.87                & 0.91                & 0.98                \\
task1554                                                       & 0.89                & 0.98                & 0.98                & task743                                                        & 0.52                & 0.54                & 0.54                \\
task1557                                                       & 0.88                & 0.87                & 0.87                & task760                                                        & 0.02                & 0.09                & 0.09                \\
task1562                                                       & 0.52                & 0.51                & 0.51                & task769                                                        & 0.95                & 0.96                & 0.96                \\
task1586                                                       & 0.36                & 0.39                & 0.40                & task827                                                        & 0.87                & 0.91                & 0.91                \\
task1598                                                       & 0.49                & 0.47                & 0.51                & task828                                                        & 0.89                & 0.95                & 0.95                \\
task1612                                                       & 0.82                & 0.91                & 0.91                & task879                                                        & 0.50                & 0.88                & 1.00                \\
task1615                                                       & 0.92                & 0.97                & 0.97                & task880                                                        & 0.51                & 0.99                & 0.97                \\
task1622                                                       & 0.93                & 0.96                & 0.96                & task890                                                        & 0.53                & 0.33                & 0.33                \\
task1624                                                       & 0.53                & 0.91                & 0.94                & task891                                                        & 0.72                & 0.73                & 0.73                \\
task1631                                                       & 0.99                & 0.99                & 0.99                & task892                                                        & 0.39                & 0.30                & 0.30                \\
task1640                                                       & 0.78                & 0.86                & 0.89                & task893                                                        & 1.00                & 0.67                & 0.67                \\
task1659                                                       & 0.49                & 0.62                & 0.71                & task935                                                        & 0.64                & 0.80                & 0.84                \\
task1664                                                       & 0.90                & 0.90                & 0.90                & task936                                                        & 0.72                & 0.81                & 0.84                \\
task1728                                                       & 0.57                & 0.56                & 0.62                & task937                                                        & 0.75                & 0.83                & 0.89                \\
task190                                                        & 0.77                & 0.64                & 0.85                & task957                                                        & 0.50                & 0.52                & 0.51                \\
task199                                                        & 0.52                & 0.87                & 0.92                & task970                                                        & 0.75                & 0.81                & 0.84                \\
task200                                                        & 0.96                & 0.95                & 0.97                &                                                                &                     &                     &                     \\ \hline
\end{tabular}
    }
    \caption{ 1 model Setup. Scores of baselines for each task.}
    \label{tab:1m_baselines}
\end{table*}
\begin{table*}[t!]
    \centering
    \small
    \resizebox{\linewidth}{!}
    {
        \begin{tabular}{llllll|llllll}
\hline
\textbf{\begin{tabular}[c]{@{}l@{}}Task\\ Number\end{tabular}} & \textbf{\begin{tabular}[c]{@{}l@{}}Score\\ 10\end{tabular}} & \textbf{\begin{tabular}[c]{@{}l@{}}Score\\ 100\end{tabular}} & \textbf{\begin{tabular}[c]{@{}l@{}}Score\\ 200\end{tabular}} & \textbf{\begin{tabular}[c]{@{}l@{}}Score\\ 1000\end{tabular}} & \textbf{\begin{tabular}[c]{@{}l@{}}Score\\ All\end{tabular}} & \textbf{\begin{tabular}[c]{@{}l@{}}Task\\ Number\end{tabular}} & \textbf{\begin{tabular}[c]{@{}l@{}}Score\\ 10\end{tabular}} & \textbf{\begin{tabular}[c]{@{}l@{}}Score\\ 100\end{tabular}} & \textbf{\begin{tabular}[c]{@{}l@{}}Score\\ 200\end{tabular}} & \textbf{\begin{tabular}[c]{@{}l@{}}Score\\ 1000\end{tabular}} & \textbf{\begin{tabular}[c]{@{}l@{}}Score\\ All\end{tabular}} \\ \hline
task020                                                        & 0.52                                                        & 0.52                                                         & 0.50                                                         & 0.50                                                          & 0.50                                                         & task201                                                        & 0.34                                                        & 0.34                                                         & 0.33                                                         & 0.73                                                          & 0.88                                                         \\
task033                                                        & 0.67                                                        & 0.67                                                         & 0.66                                                         & 0.74                                                          & 0.79                                                         & task202                                                        & 0.94                                                        & 0.94                                                         & 0.93                                                         & 0.97                                                          & 0.96                                                         \\
task034                                                        & 0.94                                                        & 0.94                                                         & 0.94                                                         & 0.94                                                          & 0.95                                                         & task219                                                        & 0.34                                                        & 0.34                                                         & 0.36                                                         & 0.33                                                          & 0.34                                                         \\
task035                                                        & 0.93                                                        & 0.93                                                         & 0.93                                                         & 0.92                                                          & 0.92                                                         & task220                                                        & 0.98                                                        & 0.98                                                         & 1.00                                                         & 1.00                                                          & 1.00                                                         \\
task036                                                        & 0.31                                                        & 0.31                                                         & 0.32                                                         & 0.37                                                          & 0.34                                                         & task226                                                        & 0.50                                                        & 0.50                                                         & 0.50                                                         & 0.50                                                          & 0.50                                                         \\
task039                                                        & 0.74                                                        & 0.74                                                         & 0.65                                                         & 0.61                                                          & 0.81                                                         & task232                                                        & 0.69                                                        & 0.69                                                         & 0.77                                                         & 0.88                                                          & 0.87                                                         \\
task050                                                        & 0.73                                                        & 0.73                                                         & 0.79                                                         & 0.85                                                          & 0.83                                                         & task233                                                        & 0.50                                                        & 0.50                                                         & 0.50                                                         & 0.53                                                          & 0.50                                                         \\
task102                                                        & 0.65                                                        & 0.65                                                         & 0.66                                                         & 0.68                                                          & 0.69                                                         & task242                                                        & 0.98                                                        & 0.98                                                         & 0.99                                                         & 0.99                                                          & 0.63                                                         \\
task1152                                                       & 1.00                                                        & 1.00                                                         & 0.98                                                         & 0.98                                                          & 1.00                                                         & task249                                                        & 0.71                                                        & 0.71                                                         & 0.65                                                         & 0.63                                                          & 0.71                                                         \\
task1153                                                       & 0.99                                                        & 0.99                                                         & 0.98                                                         & 1.00                                                          & 1.00                                                         & task281                                                        & 0.65                                                        & 0.65                                                         & 0.62                                                         & 0.67                                                          & 0.69                                                         \\
task1154                                                       & 1.00                                                        & 1.00                                                         & 0.99                                                         & 1.00                                                          & 1.00                                                         & task288                                                        & 0.30                                                        & 0.30                                                         & 0.33                                                         & 0.33                                                          & 0.33                                                         \\
task1155                                                       & 1.00                                                        & 1.00                                                         & 1.00                                                         & 1.00                                                          & 1.00                                                         & task290                                                        & 0.91                                                        & 0.91                                                         & 0.92                                                         & 0.93                                                          & 0.93                                                         \\
task1156                                                       & 1.00                                                        & 1.00                                                         & 1.00                                                         & 1.00                                                          & 1.00                                                         & task304                                                        & 0.53                                                        & 0.53                                                         & 0.45                                                         & 0.67                                                          & 0.85                                                         \\
task1157                                                       & 0.99                                                        & 0.99                                                         & 1.00                                                         & 1.00                                                          & 1.00                                                         & task329                                                        & 0.76                                                        & 0.76                                                         & 0.78                                                         & 0.85                                                          & 0.85                                                         \\
task1158                                                       & 1.00                                                        & 1.00                                                         & 1.00                                                         & 1.00                                                          & 1.00                                                         & task330                                                        & 0.86                                                        & 0.86                                                         & 0.87                                                         & 0.89                                                          & 0.89                                                         \\
task1159                                                       & 1.00                                                        & 1.00                                                         & 1.00                                                         & 1.00                                                          & 1.00                                                         & task349                                                        & 0.70                                                        & 0.70                                                         & 0.73                                                         & 0.81                                                          & 0.93                                                         \\
task1161                                                       & 0.34                                                        & 0.34                                                         & 0.36                                                         & 0.37                                                          & 0.38                                                         & task362                                                        & 0.92                                                        & 0.92                                                         & 0.93                                                         & 0.96                                                          & 0.95                                                         \\
task1195                                                       & 0.96                                                        & 0.96                                                         & 0.97                                                         & 0.97                                                          & 0.98                                                         & task391                                                        & 0.90                                                        & 0.90                                                         & 0.93                                                         & 0.92                                                          & 0.92                                                         \\
task121                                                        & 0.50                                                        & 0.50                                                         & 0.47                                                         & 0.47                                                          & 0.50                                                         & task392                                                        & 0.88                                                        & 0.88                                                         & 0.91                                                         & 0.94                                                          & 0.95                                                         \\
task133                                                        & 0.76                                                        & 0.76                                                         & 0.76                                                         & 0.84                                                          & 0.85                                                         & task393                                                        & 0.47                                                        & 0.47                                                         & 0.52                                                         & 0.52                                                          & 0.52                                                         \\
task1342                                                       & 0.13                                                        & 0.13                                                         & 0.12                                                         & 0.14                                                          & 0.20                                                         & task401                                                        & 0.82                                                        & 0.82                                                         & 0.83                                                         & 0.85                                                          & 0.92                                                         \\
task1344                                                       & 0.82                                                        & 0.82                                                         & 0.82                                                         & 0.90                                                          & 0.91                                                         & task402                                                        & 0.48                                                        & 0.48                                                         & 0.55                                                         & 0.55                                                          & 0.57                                                         \\
task1345                                                       & 0.39                                                        & 0.39                                                         & 0.39                                                         & 0.39                                                          & 0.40                                                         & task418                                                        & 0.29                                                        & 0.29                                                         & 0.29                                                         & 0.31                                                          & 0.33                                                         \\
task1356                                                       & 0.04                                                        & 0.04                                                         & 0.03                                                         & 0.01                                                          & 0.03                                                         & task442                                                        & 0.63                                                        & 0.63                                                         & 0.63                                                         & 0.66                                                          & 0.63                                                         \\
task1358                                                       & 0.36                                                        & 0.36                                                         & 0.36                                                         & 0.36                                                          & 0.38                                                         & task500                                                        & 0.36                                                        & 0.36                                                         & 0.39                                                         & 0.37                                                          & 0.39                                                         \\
task1385                                                       & 0.34                                                        & 0.34                                                         & 0.39                                                         & 0.66                                                          & 0.62                                                         & task510                                                        & 0.46                                                        & 0.46                                                         & 0.47                                                         & 0.47                                                          & 0.51                                                         \\
task1386                                                       & 0.31                                                        & 0.31                                                         & 0.42                                                         & 0.48                                                          & 0.53                                                         & task520                                                        & 1.00                                                        & 1.00                                                         & 1.00                                                         & 1.00                                                          & 1.00                                                         \\
task1387                                                       & 0.33                                                        & 0.33                                                         & 0.35                                                         & 0.44                                                          & 0.46                                                         & task569                                                        & 0.43                                                        & 0.43                                                         & 0.46                                                         & 0.44                                                          & 0.46                                                         \\
task1388                                                       & 0.74                                                        & 0.74                                                         & 0.80                                                         & 0.80                                                          & 0.80                                                         & task602                                                        & 0.06                                                        & 0.06                                                         & 0.41                                                         & 0.41                                                          & 0.41                                                         \\
task1390                                                       & 0.48                                                        & 0.48                                                         & 0.54                                                         & 0.50                                                          & 0.56                                                         & task613                                                        & 0.23                                                        & 0.23                                                         & 0.21                                                         & 0.29                                                          & 0.40                                                         \\
task1391                                                       & 0.76                                                        & 0.76                                                         & 0.83                                                         & 0.85                                                          & 0.94                                                         & task614                                                        & 0.62                                                        & 0.62                                                         & 0.65                                                         & 0.63                                                          & 0.66                                                         \\
task1393                                                       & 0.94                                                        & 0.94                                                         & 0.87                                                         & 0.92                                                          & 0.92                                                         & task619                                                        & 0.44                                                        & 0.44                                                         & 0.45                                                         & 0.45                                                          & 0.42                                                         \\
task1394                                                       & 0.84                                                        & 0.84                                                         & 0.87                                                         & 0.87                                                          & 0.88                                                         & task620                                                        & 0.02                                                        & 0.02                                                         & 0.03                                                         & 0.06                                                          & 0.10                                                         \\
task1407                                                       & 0.41                                                        & 0.41                                                         & 0.43                                                         & 0.43                                                          & 0.48                                                         & task623                                                        & 0.50                                                        & 0.50                                                         & 0.96                                                         & 0.98                                                          & 0.96                                                         \\
task1409                                                       & 0.47                                                        & 0.47                                                         & 0.49                                                         & 0.52                                                          & 0.53                                                         & task640                                                        & 0.90                                                        & 0.90                                                         & 0.85                                                         & 0.85                                                          & 0.85                                                         \\
task1439                                                       & 0.58                                                        & 0.58                                                         & 0.60                                                         & 0.49                                                          & 0.63                                                         & task641                                                        & 0.79                                                        & 0.79                                                         & 0.63                                                         & 0.63                                                          & 0.81                                                         \\
task1442                                                       & 0.66                                                        & 0.66                                                         & 0.60                                                         & 0.66                                                          & 0.68                                                         & task642                                                        & 0.50                                                        & 0.50                                                         & 0.50                                                         & 0.50                                                          & 0.50                                                         \\
task1516                                                       & 0.99                                                        & 0.99                                                         & 1.00                                                         & 0.98                                                          & 1.00                                                         & task645                                                        & 0.97                                                        & 0.97                                                         & 0.96                                                         & 0.97                                                          & 0.98                                                         \\
task1529                                                       & 0.84                                                        & 0.84                                                         & 0.89                                                         & 0.92                                                          & 0.96                                                         & task648                                                        & 0.76                                                        & 0.76                                                         & 0.77                                                         & 0.77                                                          & 0.71                                                         \\
task1531                                                       & 0.58                                                        & 0.58                                                         & 0.67                                                         & 0.50                                                          & 0.65                                                         & task670                                                        & 0.77                                                        & 0.77                                                         & 0.76                                                         & 0.74                                                          & 0.73                                                         \\
task1533                                                       & 0.50                                                        & 0.50                                                         & 0.50                                                         & 0.77                                                          & 0.72                                                         & task671                                                        & 0.64                                                        & 0.64                                                         & 0.64                                                         & 0.62                                                          & 0.61                                                         \\
task1534                                                       & 0.54                                                        & 0.54                                                         & 0.89                                                         & 0.97                                                          & 0.50                                                         & task677                                                        & 0.30                                                        & 0.30                                                         & 0.33                                                         & 0.34                                                          & 0.33                                                         \\
task1540                                                       & 0.37                                                        & 0.37                                                         & 0.41                                                         & 0.40                                                          & 0.40                                                         & task738                                                        & 0.89                                                        & 0.89                                                         & 0.90                                                         & 0.90                                                          & 0.96                                                         \\
task1554                                                       & 0.83                                                        & 0.83                                                         & 0.93                                                         & 0.98                                                          & 0.95                                                         & task743                                                        & 0.47                                                        & 0.47                                                         & 0.48                                                         & 0.55                                                          & 0.54                                                         \\
task1557                                                       & 0.88                                                        & 0.88                                                         & 0.88                                                         & 0.87                                                          & 0.88                                                         & task760                                                        & 0.08                                                        & 0.08                                                         & 0.68                                                         & 0.68                                                          & 0.68                                                         \\
task1562                                                       & 0.52                                                        & 0.52                                                         & 0.53                                                         & 0.53                                                          & 0.51                                                         & task769                                                        & 0.96                                                        & 0.96                                                         & 0.96                                                         & 0.97                                                          & 0.96                                                         \\
task1586                                                       & 0.29                                                        & 0.29                                                         & 0.29                                                         & 0.28                                                          & 0.27                                                         & task827                                                        & 0.87                                                        & 0.87                                                         & 0.90                                                         & 0.91                                                          & 0.92                                                         \\
task1598                                                       & 0.49                                                        & 0.49                                                         & 0.49                                                         & 0.49                                                          & 0.52                                                         & task828                                                        & 0.90                                                        & 0.90                                                         & 0.94                                                         & 0.98                                                          & 0.96                                                         \\
task1612                                                       & 0.88                                                        & 0.88                                                         & 0.85                                                         & 0.86                                                          & 0.91                                                         & task879                                                        & 0.50                                                        & 0.50                                                         & 0.50                                                         & 0.96                                                          & 1.00                                                         \\
task1615                                                       & 0.96                                                        & 0.96                                                         & 0.93                                                         & 0.96                                                          & 0.97                                                         & task880                                                        & 0.33                                                        & 0.33                                                         & 0.63                                                         & 0.99                                                          & 0.99                                                         \\
task1622                                                       & 0.95                                                        & 0.95                                                         & 0.93                                                         & 0.96                                                          & 0.96                                                         & task890                                                        & 0.65                                                        & 0.65                                                         & 0.68                                                         & 0.68                                                          & 0.63                                                         \\
task1624                                                       & 0.65                                                        & 0.65                                                         & 0.75                                                         & 0.79                                                          & 0.83                                                         & task891                                                        & 0.85                                                        & 0.85                                                         & 0.82                                                         & 0.82                                                          & 0.85                                                         \\
task1631                                                       & 0.99                                                        & 0.99                                                         & 0.99                                                         & 0.99                                                          & 0.98                                                         & task892                                                        & 0.69                                                        & 0.69                                                         & 0.48                                                         & 0.48                                                          & 0.69                                                         \\
task1640                                                       & 0.74                                                        & 0.74                                                         & 0.81                                                         & 0.84                                                          & 0.84                                                         & task893                                                        & 1.00                                                        & 1.00                                                         & 1.00                                                         & 1.00                                                          & 1.00                                                         \\
task1659                                                       & 0.50                                                        & 0.50                                                         & 0.48                                                         & 0.53                                                          & 0.67                                                         & task935                                                        & 0.72                                                        & 0.72                                                         & 0.85                                                         & 0.89                                                          & 0.86                                                         \\
task1664                                                       & 0.95                                                        & 0.95                                                         & 0.96                                                         & 0.97                                                          & 0.98                                                         & task936                                                        & 0.83                                                        & 0.83                                                         & 0.83                                                         & 0.83                                                          & 0.84                                                         \\
task1728                                                       & 0.54                                                        & 0.54                                                         & 0.54                                                         & 0.58                                                          & 0.62                                                         & task937                                                        & 0.75                                                        & 0.75                                                         & 0.77                                                         & 0.86                                                          & 0.85                                                         \\
task190                                                        & 0.75                                                        & 0.75                                                         & 0.86                                                         & 0.86                                                          & 0.87                                                         & task957                                                        & 0.50                                                        & 0.50                                                         & 0.50                                                         & 0.50                                                          & 0.49                                                         \\
task199                                                        & 0.60                                                        & 0.60                                                         & 0.56                                                         & 0.84                                                          & 0.73                                                         & task970                                                        & 0.74                                                        & 0.74                                                         & 0.75                                                         & 0.80                                                          & 0.88                                                         \\
task200                                                        & 0.95                                                        & 0.95                                                         & 0.94                                                         & 0.94                                                          & 0.95                                                         &                                                                &                                                             &                                                              &                                                              &                                                               &                                                              \\ \hline
\end{tabular}
    }
    \caption{ 1 model Setup. Scores when 1 model is used for 1 task using x samples. }
    \label{tab:1m_scores}
\end{table*}

\begin{table*}[t!]
\vspace{-4.5mm}
    \centering
    \small
    \resizebox{0.95\linewidth}{!}
    {
        \begin{tabular}{lll|lll}
\hline
\textbf{\begin{tabular}[c]{@{}l@{}}Task\\ Number\end{tabular}} & \textbf{MLT Baseline 1} & \textbf{MLT Baseline 2} & \textbf{\begin{tabular}[c]{@{}l@{}}Task\\ Number\end{tabular}} & \textbf{MLT Baseline 1} & \textbf{MLT Baseline 2} \\ \hline
task020                                                        & 0.60                    & 0.56                    & task201                                                        & 0.33                    & 0.29                    \\
task033                                                        & 0.79                    & 0.75                    & task202                                                        & 0.40                    & 0.37                    \\
task034                                                        & 0.99                    & 0.95                    & task219                                                        & 0.38                    & 0.53                    \\
task035                                                        & 0.97                    & 0.93                    & task220                                                        & 0.98                    & 0.98                    \\
task036                                                        & 0.38                    & 0.35                    & task226                                                        & 0.56                    & 0.53                    \\
task039                                                        & 0.76                    & 0.72                    & task232                                                        & 0.70                    & 0.68                    \\
task050                                                        & 0.85                    & 0.81                    & task233                                                        & 0.59                    & 0.56                    \\
task102                                                        & 0.60                    & 0.56                    & task242                                                        & 1.00                    & 0.99                    \\
task1152                                                       & 1.00                    & 1.00                    & task249                                                        & 0.70                    & 0.65                    \\
task1153                                                       & 0.96                    & 0.92                    & task281                                                        & 0.70                    & 0.68                    \\
task1154                                                       & 1.00                    & 1.00                    & task288                                                        & 0.39                    & 0.35                    \\
task1155                                                       & 1.00                    & 1.00                    & task290                                                        & 0.90                    & 0.88                    \\
task1156                                                       & 0.99                    & 0.98                    & task304                                                        & 0.59                    & 0.55                    \\
task1157                                                       & 1.00                    & 1.00                    & task329                                                        & 0.90                    & 0.85                    \\
task1158                                                       & 1.00                    & 1.00                    & task330                                                        & 0.93                    & 0.92                    \\
task1159                                                       & 1.00                    & 1.00                    & task349                                                        & 0.82                    & 0.80                    \\
task1161                                                       & 0.28                    & 0.25                    & task362                                                        & 0.94                    & 0.92                    \\
task1195                                                       & 0.99                    & 0.95                    & task391                                                        & 0.96                    & 0.91                    \\
task121                                                        & 0.46                    & 0.42                    & task392                                                        & 0.95                    & 0.93                    \\
task133                                                        & 0.79                    & 0.76                    & task393                                                        & 0.57                    & 0.52                    \\
task1342                                                       & 0.57                    & 0.53                    & task401                                                        & 0.88                    & 0.82                    \\
task1344                                                       & 0.76                    & 0.75                    & task402                                                        & 0.54                    & 0.51                    \\
task1345                                                       & 0.34                    & 0.32                    & task418                                                        & 0.27                    & 0.25                    \\
task1356                                                       & 0.40                    & 0.56                    & task442                                                        & 0.63                    & 0.61                    \\
task1358                                                       & 0.34                    & 0.31                    & task500                                                        & 0.41                    & 0.38                    \\
task1385                                                       & 0.59                    & 0.56                    & task510                                                        & 0.47                    & 0.45                    \\
task1386                                                       & 0.64                    & 0.60                    & task520                                                        & 1.00                    & 0.96                    \\
task1387                                                       & 0.55                    & 0.51                    & task569                                                        & 0.45                    & 0.43                    \\
task1388                                                       & 0.85                    & 0.80                    & task602                                                        & 0.57                    & 0.53                    \\
task1390                                                       & 0.62                    & 0.57                    & task613                                                        & 0.28                    & 0.47                    \\
task1391                                                       & 0.87                    & 0.84                    & task614                                                        & 0.62                    & 0.59                    \\
task1393                                                       & 0.97                    & 0.94                    & task619                                                        & 0.07                    & 0.05                    \\
task1394                                                       & 0.70                    & 0.67                    & task620                                                        & 0.19                    & 0.06                    \\
task1407                                                       & 0.51                    & 0.50                    & task623                                                        & 1.00                    & 1.00                    \\
task1409                                                       & 0.50                    & 0.44                    & task640                                                        & 0.37                    & 0.34                    \\
task1439                                                       & 0.80                    & 0.76                    & task641                                                        & 0.31                    & 0.27                    \\
task1442                                                       & 0.77                    & 0.74                    & task642                                                        & 0.52                    & 0.47                    \\
task1516                                                       & 1.00                    & 1.00                    & task645                                                        & 0.98                    & 0.95                    \\
task1529                                                       & 0.94                    & 0.90                    & task648                                                        & 0.70                    & 0.68                    \\
task1531                                                       & 0.61                    & 0.59                    & task670                                                        & 0.66                    & 0.61                    \\
task1533                                                       & 0.57                    & 0.54                    & task671                                                        & 0.57                    & 0.55                    \\
task1534                                                       & 0.73                    & 0.71                    & task677                                                        & 0.31                    & 0.29                    \\
task1540                                                       & 0.34                    & 0.31                    & task738                                                        & 0.94                    & 0.91                    \\
task1554                                                       & 0.83                    & 0.78                    & task743                                                        & 0.54                    & 0.52                    \\
task1557                                                       & 0.68                    & 0.64                    & task760                                                        & 0.58                    & 0.55                    \\
task1562                                                       & 0.48                    & 0.44                    & task769                                                        & 1.00                    & 0.96                    \\
task1586                                                       & 0.26                    & 0.24                    & task827                                                        & 0.95                    & 0.91                    \\
task1598                                                       & 0.62                    & 0.57                    & task828                                                        & 0.98                    & 0.93                    \\
task1612                                                       & 0.91                    & 0.86                    & task879                                                        & 0.53                    & 0.49                    \\
task1615                                                       & 0.06                    & 0.02                    & task880                                                        & 0.32                    & 0.28                    \\
task1622                                                       & 1.00                    & 0.96                    & task890                                                        & 0.69                    & 0.65                    \\
task1624                                                       & 0.80                    & 0.77                    & task891                                                        & 0.88                    & 0.87                    \\
task1631                                                       & 1.00                    & 0.99                    & task892                                                        & 0.48                    & 0.45                    \\
task1640                                                       & 0.79                    & 0.75                    & task893                                                        & 1.00                    & 0.99                    \\
task1659                                                       & 0.54                    & 0.51                    & task935                                                        & 0.81                    & 0.78                    \\
task1664                                                       & 0.95                    & 0.90                    & task936                                                        & 0.78                    & 0.75                    \\
task1728                                                       & 0.63                    & 0.59                    & task937                                                        & 0.80                    & 0.77                    \\
task190                                                        & 0.76                    & 0.72                    & task957                                                        & 0.56                    & 0.54                    \\
task199                                                        & 0.70                    & 0.67                    & task970                                                        & 0.82                    & 0.77                    \\
task200                                                        & 0.41                    & 0.37                    &                                                                &                         &                         \\ \hline
\end{tabular}
    }
    \caption{ MLT Setup. Scores of baselines for each task. The baselines were trained in a combined fashion.}
    \label{tab:mlt_baselines}
    % \vspace{-6mm}
\end{table*}
\begin{table*}[t!]
    \centering
    \small
    \resizebox{\linewidth}{!}
    {
        \begin{tabular}{llllll|llllll}
\hline
\textbf{\begin{tabular}[c]{@{}l@{}}Task\\ Number\end{tabular}} & \textbf{\begin{tabular}[c]{@{}l@{}}Score\\ 10\end{tabular}} & \textbf{\begin{tabular}[c]{@{}l@{}}Score\\ 100\end{tabular}} & \textbf{\begin{tabular}[c]{@{}l@{}}Score \\ 200\end{tabular}} & \textbf{\begin{tabular}[c]{@{}l@{}}Score\\ 1000\end{tabular}} & \textbf{\begin{tabular}[c]{@{}l@{}}Score\\ All\end{tabular}} & \textbf{\begin{tabular}[c]{@{}l@{}}Task\\ Number\end{tabular}} & \textbf{\begin{tabular}[c]{@{}l@{}}Score\\ 10\end{tabular}} & \textbf{\begin{tabular}[c]{@{}l@{}}Score\\ 100\end{tabular}} & \textbf{\begin{tabular}[c]{@{}l@{}}Score \\ 200\end{tabular}} & \textbf{\begin{tabular}[c]{@{}l@{}}Score\\ 1000\end{tabular}} & \textbf{\begin{tabular}[c]{@{}l@{}}Score\\ All\end{tabular}} \\ \hline
task020                                                        & 0.54                                                        & 0.57                                                         & 0.60                                                          & 0.53                                                          & 0.53                                                         & task201                                                        & 0.13                                                        & 0.39                                                         & 0.58                                                          & 0.87                                                          & 0.88                                                         \\
task033                                                        & 0.64                                                        & 0.71                                                         & 0.69                                                          & 0.73                                                          & 0.76                                                         & task202                                                        & 0.83                                                        & 0.90                                                         & 0.94                                                          & 0.98                                                          & 1.00                                                         \\
task034                                                        & 0.93                                                        & 0.95                                                         & 0.95                                                          & 0.96                                                          & 0.97                                                         & task219                                                        & 0.20                                                        & 0.27                                                         & 0.28                                                          & 0.27                                                          & 0.30                                                         \\
task035                                                        & 0.91                                                        & 0.93                                                         & 0.93                                                          & 0.93                                                          & 0.95                                                         & task220                                                        & 0.98                                                        & 0.98                                                         & 0.99                                                          & 0.99                                                          & 1.00                                                         \\
task036                                                        & 0.20                                                        & 0.23                                                         & 0.28                                                          & 0.31                                                          & 0.33                                                         & task226                                                        & 0.51                                                        & 0.51                                                         & 0.51                                                          & 0.51                                                          & 0.51                                                         \\
task039                                                        & 0.38                                                        & 0.59                                                         & 0.68                                                          & 0.72                                                          & 0.73                                                         & task232                                                        & 0.51                                                        & 0.51                                                         & 0.77                                                          & 0.75                                                          & 0.76                                                         \\
task050                                                        & 0.73                                                        & 0.80                                                         & 0.78                                                          & 0.81                                                          & 0.83                                                         & task233                                                        & 0.51                                                        & 0.51                                                         & 0.57                                                          & 0.53                                                          & 0.53                                                         \\
task102                                                        & 0.50                                                        & 0.56                                                         & 0.55                                                          & 0.58                                                          & 0.60                                                         & task242                                                        & 0.98                                                        & 0.99                                                         & 0.99                                                          & 0.99                                                          & 1.00                                                         \\
task1152                                                       & 0.86                                                        & 1.00                                                         & 1.00                                                          & 1.00                                                          & 1.00                                                         & task249                                                        & 0.63                                                        & 0.79                                                         & 0.81                                                          & 0.77                                                          & 0.78                                                         \\
task1153                                                       & 0.59                                                        & 0.96                                                         & 1.00                                                          & 1.00                                                          & 1.00                                                         & task281                                                        & 0.55                                                        & 0.64                                                         & 0.65                                                          & 0.71                                                          & 0.74                                                         \\
task1154                                                       & 0.61                                                        & 0.87                                                         & 0.99                                                          & 1.00                                                          & 1.00                                                         & task288                                                        & 0.30                                                        & 0.32                                                         & 0.33                                                          & 0.34                                                          & 0.37                                                         \\
task1155                                                       & 0.93                                                        & 1.00                                                         & 1.00                                                          & 1.00                                                          & 1.00                                                         & task290                                                        & 0.84                                                        & 0.90                                                         & 0.92                                                          & 0.94                                                          & 0.95                                                         \\
task1156                                                       & 0.67                                                        & 1.00                                                         & 1.00                                                          & 1.00                                                          & 1.00                                                         & task304                                                        & 0.21                                                        & 0.77                                                         & 0.83                                                          & 0.86                                                          & 0.88                                                         \\
task1157                                                       & 0.53                                                        & 0.98                                                         & 1.00                                                          & 1.00                                                          & 1.00                                                         & task329                                                        & 0.51                                                        & 0.64                                                         & 0.77                                                          & 0.79                                                          & 0.82                                                         \\
task1158                                                       & 0.78                                                        & 1.00                                                         & 1.00                                                          & 0.95                                                          & 0.96                                                         & task330                                                        & 0.80                                                        & 0.90                                                         & 0.93                                                          & 0.91                                                          & 0.94                                                         \\
task1159                                                       & 0.42                                                        & 0.93                                                         & 1.00                                                          & 1.00                                                          & 1.00                                                         & task349                                                        & 0.54                                                        & 0.72                                                         & 0.76                                                          & 0.82                                                          & 0.85                                                         \\
task1161                                                       & 0.25                                                        & 0.27                                                         & 0.27                                                          & 0.26                                                          & 0.28                                                         & task362                                                        & 0.83                                                        & 0.91                                                         & 0.91                                                          & 0.94                                                          & 0.96                                                         \\
task1195                                                       & 0.88                                                        & 0.94                                                         & 0.96                                                          & 0.97                                                          & 0.98                                                         & task391                                                        & 0.89                                                        & 0.88                                                         & 0.92                                                          & 0.94                                                          & 0.95                                                         \\
task121                                                        & 0.46                                                        & 0.46                                                         & 0.48                                                          & 0.50                                                          & 0.51                                                         & task392                                                        & 0.90                                                        & 0.90                                                         & 0.93                                                          & 0.95                                                          & 0.97                                                         \\
task133                                                        & 0.69                                                        & 0.65                                                         & 0.77                                                          & 0.80                                                          & 0.81                                                         & task393                                                        & 0.24                                                        & 0.32                                                         & 0.54                                                          & 0.39                                                          & 0.41                                                         \\
task1342                                                       & 0.05                                                        & 0.07                                                         & 0.10                                                          & 0.12                                                          & 0.62                                                         & task401                                                        & 0.63                                                        & 0.80                                                         & 0.84                                                          & 0.88                                                          & 0.92                                                         \\
task1344                                                       & 0.74                                                        & 0.78                                                         & 0.82                                                          & 0.91                                                          & 0.92                                                         & task402                                                        & 0.49                                                        & 0.40                                                         & 0.53                                                          & 0.52                                                          & 0.55                                                         \\
task1345                                                       & 0.36                                                        & 0.37                                                         & 0.37                                                          & 0.36                                                          & 0.37                                                         & task418                                                        & 0.22                                                        & 0.25                                                         & 0.24                                                          & 0.26                                                          & 0.28                                                         \\
task1356                                                       & 0.02                                                        & 0.04                                                         & 0.05                                                          & 0.04                                                          & 0.27                                                         & task442                                                        & 0.57                                                        & 0.61                                                         & 0.57                                                          & 0.61                                                          & 0.62                                                         \\
task1358                                                       & 0.29                                                        & 0.29                                                         & 0.31                                                          & 0.31                                                          & 0.32                                                         & task500                                                        & 0.33                                                        & 0.38                                                         & 0.37                                                          & 0.38                                                          & 0.42                                                         \\
task1385                                                       & 0.34                                                        & 0.49                                                         & 0.56                                                          & 0.65                                                          & 0.68                                                         & task510                                                        & 0.42                                                        & 0.44                                                         & 0.44                                                          & 0.45                                                          & 0.45                                                         \\
task1386                                                       & 0.43                                                        & 0.48                                                         & 0.54                                                          & 0.58                                                          & 0.61                                                         & task520                                                        & 0.99                                                        & 1.00                                                         & 0.99                                                          & 1.00                                                          & 1.00                                                         \\
task1387                                                       & 0.48                                                        & 0.38                                                         & 0.46                                                          & 0.51                                                          & 0.52                                                         & task569                                                        & 0.45                                                        & 0.43                                                         & 0.38                                                          & 0.44                                                          & 0.45                                                         \\
task1388                                                       & 0.71                                                        & 0.80                                                         & 0.79                                                          & 0.77                                                          & 0.81                                                         & task602                                                        & 0.03                                                        & 0.08                                                         & 0.04                                                          & 0.02                                                          & 0.36                                                         \\
task1390                                                       & 0.51                                                        & 0.58                                                         & 0.75                                                          & 0.80                                                          & 0.83                                                         & task613                                                        & 0.15                                                        & 0.17                                                         & 0.17                                                          & 0.22                                                          & 0.23                                                         \\
task1391                                                       & 0.76                                                        & 0.83                                                         & 0.82                                                          & 0.90                                                          & 0.93                                                         & task614                                                        & 0.56                                                        & 0.59                                                         & 0.61                                                          & 0.62                                                          & 0.64                                                         \\
task1393                                                       & 0.85                                                        & 0.88                                                         & 0.95                                                          & 0.98                                                          & 1.00                                                         & task619                                                        & 0.34                                                        & 0.37                                                         & 0.37                                                          & 0.40                                                          & 0.43                                                         \\
task1394                                                       & 0.69                                                        & 0.84                                                         & 0.89                                                          & 0.86                                                          & 0.89                                                         & task620                                                        & 0.09                                                        & 0.05                                                         & 0.06                                                          & 0.12                                                          & 0.15                                                         \\
task1407                                                       & 0.41                                                        & 0.48                                                         & 0.51                                                          & 0.51                                                          & 0.51                                                         & task623                                                        & 0.52                                                        & 0.55                                                         & 0.97                                                          & 0.99                                                          & 1.00                                                         \\
task1409                                                       & 0.46                                                        & 0.51                                                         & 0.53                                                          & 0.56                                                          & 0.58                                                         & task640                                                        & 0.74                                                        & 0.86                                                         & 0.93                                                          & 0.93                                                          & 0.94                                                         \\
task1439                                                       & 0.58                                                        & 0.59                                                         & 0.66                                                          & 0.63                                                          & 0.64                                                         & task641                                                        & 0.64                                                        & 0.80                                                         & 0.83                                                          & 0.81                                                          & 0.83                                                         \\
task1442                                                       & 0.60                                                        & 0.67                                                         & 0.73                                                          & 0.72                                                          & 0.74                                                         & task642                                                        & 0.51                                                        & 0.50                                                         & 0.83                                                          & 0.85                                                          & 0.86                                                         \\
task1516                                                       & 0.84                                                        & 0.98                                                         & 1.00                                                          & 1.00                                                          & 1.00                                                         & task645                                                        & 0.95                                                        & 0.96                                                         & 0.96                                                          & 0.98                                                          & 1.00                                                         \\
task1529                                                       & 0.84                                                        & 0.91                                                         & 0.94                                                          & 0.91                                                          & 0.94                                                         & task648                                                        & 0.50                                                        & 0.75                                                         & 0.78                                                          & 0.86                                                          & 0.87                                                         \\
task1531                                                       & 0.53                                                        & 0.61                                                         & 0.56                                                          & 0.55                                                          & 0.57                                                         & task670                                                        & 0.63                                                        & 0.64                                                         & 0.63                                                          & 0.61                                                          & 0.61                                                         \\
task1533                                                       & 0.61                                                        & 0.76                                                         & 0.74                                                          & 0.79                                                          & 0.82                                                         & task671                                                        & 0.54                                                        & 0.54                                                         & 0.52                                                          & 0.51                                                          & 0.55                                                         \\
task1534                                                       & 0.62                                                        & 0.73                                                         & 0.81                                                          & 0.98                                                          & 1.00                                                         & task677                                                        & 0.24                                                        & 0.26                                                         & 0.29                                                          & 0.29                                                          & 0.29                                                         \\
task1540                                                       & 0.32                                                        & 0.34                                                         & 0.29                                                          & 0.33                                                          & 0.36                                                         & task738                                                        & 0.82                                                        & 0.90                                                         & 0.93                                                          & 0.92                                                          & 0.94                                                         \\
task1554                                                       & 0.83                                                        & 0.84                                                         & 0.87                                                          & 0.91                                                          & 0.94                                                         & task743                                                        & 0.45                                                        & 0.50                                                         & 0.52                                                          & 0.55                                                          & 0.58                                                         \\
task1557                                                       & 0.76                                                        & 0.78                                                         & 0.77                                                          & 0.78                                                          & 0.81                                                         & task760                                                        & 0.03                                                        & 0.04                                                         & 0.11                                                          & 0.12                                                          & 0.65                                                         \\
task1562                                                       & 0.52                                                        & 0.53                                                         & 0.51                                                          & 0.50                                                          & 0.52                                                         & task769                                                        & 0.93                                                        & 0.95                                                         & 0.97                                                          & 0.97                                                          & 1.00                                                         \\
task1586                                                       & 0.22                                                        & 0.23                                                         & 0.21                                                          & 0.22                                                          & 0.23                                                         & task827                                                        & 0.85                                                        & 0.86                                                         & 0.92                                                          & 0.93                                                          & 0.95                                                         \\
task1598                                                       & 0.49                                                        & 0.55                                                         & 0.57                                                          & 0.58                                                          & 0.58                                                         & task828                                                        & 0.70                                                        & 0.95                                                         & 0.93                                                          & 0.96                                                          & 0.98                                                         \\
task1612                                                       & 0.52                                                        & 0.73                                                         & 0.78                                                          & 0.88                                                          & 0.91                                                         & task879                                                        & 0.68                                                        & 0.69                                                         & 0.76                                                          & 0.98                                                          & 0.99                                                         \\
task1615                                                       & 0.54                                                        & 0.82                                                         & 0.81                                                          & 0.89                                                          & 0.91                                                         & task880                                                        & 0.27                                                        & 0.23                                                         & 0.37                                                          & 0.95                                                          & 0.96                                                         \\
task1622                                                       & 0.88                                                        & 0.92                                                         & 0.95                                                          & 0.97                                                          & 0.97                                                         & task890                                                        & 0.63                                                        & 0.68                                                         & 0.75                                                          & 0.71                                                          & 0.72                                                         \\
task1624                                                       & 0.73                                                        & 0.80                                                         & 0.77                                                          & 0.86                                                          & 0.88                                                         & task891                                                        & 0.76                                                        & 0.84                                                         & 0.84                                                          & 0.84                                                          & 0.84                                                         \\
task1631                                                       & 0.99                                                        & 0.99                                                         & 0.99                                                          & 0.99                                                          & 1.00                                                         & task892                                                        & 0.25                                                        & 0.39                                                         & 0.31                                                          & 0.71                                                          & 0.74                                                         \\
task1640                                                       & 0.74                                                        & 0.81                                                         & 0.79                                                          & 0.82                                                          & 0.84                                                         & task893                                                        & 0.75                                                        & 0.98                                                         & 1.00                                                          & 0.95                                                          & 0.96                                                         \\
task1659                                                       & 0.39                                                        & 0.44                                                         & 0.49                                                          & 0.53                                                          & 0.55                                                         & task935                                                        & 0.60                                                        & 0.83                                                         & 0.81                                                          & 0.89                                                          & 0.91                                                         \\
task1664                                                       & 0.68                                                        & 0.90                                                         & 0.92                                                          & 0.95                                                          & 0.96                                                         & task936                                                        & 0.76                                                        & 0.84                                                         & 0.81                                                          & 0.89                                                          & 0.93                                                         \\
task1728                                                       & 0.52                                                        & 0.56                                                         & 0.58                                                          & 0.61                                                          & 0.64                                                         & task937                                                        & 0.54                                                        & 0.74                                                         & 0.82                                                          & 0.87                                                          & 0.89                                                         \\
task190                                                        & 0.55                                                        & 0.73                                                         & 0.80                                                          & 0.85                                                          & 0.88                                                         & task957                                                        & 0.52                                                        & 0.55                                                         & 0.56                                                          & 0.56                                                          & 0.57                                                         \\
task199                                                        & 0.51                                                        & 0.51                                                         & 0.80                                                          & 0.85                                                          & 0.87                                                         & task970                                                        & 0.69                                                        & 0.74                                                         & 0.75                                                          & 0.79                                                          & 0.80                                                         \\
task200                                                        & 0.92                                                        & 0.94                                                         & 0.94                                                          & 0.94                                                          & 0.97                                                         &                                                                &                                                             &                                                              &                                                               &                                                               &                                                              \\ \hline
\end{tabular}
    }
    \caption{MLT Setup. Scores when 1 model is used for MLT using x samples.}
    \label{tab:mlt_scores}
\end{table*}

\begin{table*}[ht!]
    \centering
    \small
    \resizebox{\linewidth}{!}
    {
        \begin{tabular}{l|llllllll}
\hline
\textbf{Row Labels}          & \multicolumn{1}{r}{\textbf{10}} & \multicolumn{1}{r}{{\color[HTML]{000000} \textbf{100}}} & \multicolumn{1}{r}{{\color[HTML]{000000} \textbf{200}}} & \multicolumn{1}{r}{{\color[HTML]{000000} \textbf{1000}}} & \multicolumn{1}{r}{{\color[HTML]{000000} \textbf{All}}} & \multicolumn{1}{r}{\textbf{STL Baseline 1}} & \multicolumn{1}{r}{\textbf{STL Baseline 2}} & \textbf{STL Baseline 3} \\ \hline
Answerability Classification & 0.670                           & 0.705                                                   & 0.728                                                   & 0.752                                                    & 0.744                                                   & 0.704                                       & 0.788                                       & 0.804                   \\
Cause Effect Classification  & 0.669                           & 0.797                                                   & 0.818                                                   & 0.830                                                    & 0.835                                                   & 0.751                                       & 0.760                                       & 0.763                   \\
Coreference Resolution       & 0.614                           & 0.757                                                   & 0.743                                                   & 0.776                                                    & 0.828                                                   & 0.700                                       & 0.686                                       & 0.744                   \\
Data to Text                 & 0.460                           & 0.493                                                   & 0.567                                                   & 0.578                                                    & 0.591                                                   & 0.497                                       & 0.509                                       & 0.528                   \\
Dialogue Act Recognition     & 0.629                           & 0.600                                                   & 0.711                                                   & 0.858                                                    & 0.813                                                   & 0.584                                       & 0.778                                       & 0.819                   \\
Grammar Error Correction     & 0.871                           & 0.878                                                   & 0.878                                                   & 0.871                                                    & 0.882                                                   & 0.883                                       & 0.873                                       & 0.873                   \\
Keyword Tagging              & 0.414                           & 0.405                                                   & 0.495                                                   & 0.533                                                    & 0.556                                                   & 0.493                                       & 0.533                                       & 0.527                   \\
Overlap Extraction           & 0.426                           & 0.691                                                   & 0.632                                                   & 0.638                                                    & 0.748                                                   & 0.637                                       & 0.680                                       & 0.722                   \\
Question Rewriting           & 0.670                           & 0.700                                                   & 0.704                                                   & 0.706                                                    & 0.706                                                   & 0.681                                       & 0.710                                       & 0.711                   \\
Textual Entailment           & 0.608                           & 0.724                                                   & 0.740                                                   & 0.803                                                    & 0.820                                                   & 0.710                                       & 0.778                                       & 0.804                   \\
Title Generation             & 0.364                           & 0.395                                                   & 0.425                                                   & 0.430                                                    & 0.446                                                   & 0.447                                       & 0.468                                       & 0.484                   \\
Word Analogy                 & 0.536                           & 0.998                                                   & 0.994                                                   & 0.998                                                    & 1.000                                                   & 0.966                                       & 0.878                                       & 0.878                   \\ \hline
\end{tabular}
    }
    \caption{ Scores of all baselines and STL setups according to each category. }
    \label{tab:1m_categories}
\end{table*}
\begin{table*}[ht]
    \centering
    \small
    \resizebox{\linewidth}{!}
    {
        \begin{tabular}{l|rrrrrrr}
\hline
\textbf{Row Labels}          & \textbf{10} & {\color[HTML]{000000} \textbf{100}} & {\color[HTML]{000000} \textbf{200}} & {\color[HTML]{000000} \textbf{1000}} & {\color[HTML]{000000} \textbf{All}} & \textbf{MTL Baseline 1} & \textbf{MTL Baseline 2} \\ \hline
Answerability Classification & 0.676       & 0.722                               & 0.757                               & 0.762                                & 0.762                               & 0.769                   & 0.753                   \\
Cause Effect Classification  & 0.713       & 0.769                               & 0.829                               & 0.825                                & 0.825                               & 0.851                   & 0.819                   \\
Coreference Resolution       & 0.616       & 0.753                               & 0.791                               & 0.840                                & 0.840                               & 0.819                   & 0.759                   \\
Data to Text                 & 0.516       & 0.557                               & 0.578                               & 0.588                                & 0.533                               & 0.547                   & 0.558                   \\
Dialogue Act Recognition     & 0.602       & 0.681                               & 0.720                               & 0.865                                & 0.865                               & 0.759                   & 0.600                   \\
Grammar Error Correction     & 0.763       & 0.779                               & 0.772                               & 0.778                                & 0.778                               & 0.678                   & 0.640                   \\
Keyword Tagging              & 0.409       & 0.432                               & 0.534                               & 0.523                                & 0.523                               & 0.520                   & 0.565                   \\
Overlap Extraction           & 0.467       & 0.613                               & 0.664                               & 0.711                                & 0.711                               & 0.698                   & 0.696                   \\
Question Rewriting           & 0.654       & 0.665                               & 0.672                               & 0.676                                & 0.676                               & 0.679                   & 0.659                   \\
Textual Entailment           & 0.630       & 0.733                               & 0.797                               & 0.842                                & 0.842                               & 0.672                   & 0.622                   \\
Title Generation             & 0.404       & 0.433                               & 0.440                               & 0.439                                & 0.382                               & 0.422                   & 0.452                   \\
Word Analogy                 & 0.674       & 0.968                               & 0.999                               & 0.994                                & 0.994                               & 1.000                   & 0.988                   \\ \hline
\end{tabular}
    }
    \caption{Scores of all baselines and MTL setups according to each category.}
    \label{tab:mtl_categories}
\end{table*}

\begin{table*}[h!]
    \centering
    \small
    \fontsize{9pt}{\baselineskip}\selectfont % font size
    \renewcommand\tabcolsep{1pt} % column space
    \renewcommand\arraystretch{1}
    \resizebox{\linewidth}{!}
    {
       \begin{tabular}{ll}
        \hline
        \textbf{Task}      & \textbf{Textual Entailment}                                                       \\ \hline
        \textbf{Definition} & Definition: In this task, you're given two sentences. Indicate if the first    \\
        \textbf{}          & sentence clearly entails the second sentence (i.e., one can conclude the          \\
        \textbf{}          & 2nd sentence by reading the 1st one)Indicate your answer with '1'                 \\
                           & if the first sentence entails the second sentence, otherwise answer with '0'      \\ \hline
        \textbf{Example 1} & Input: Sentence 1: No Weapons of Mass Destruction Found in Iraq Yet.              \\
                           & Sentence 2: Weapons of Mass Destruction Found in Iraq.                            \\
        \textbf{}          & Output: 0                                                                         \\ \hline
        \textbf{Example 2}  & Input: Sentence 1: A place of sorrow, after Pope John Paul II died, became     \\
        \textbf{}          & a place of celebration, as Roman Catholic faithful gathered in downtown           \\
        \textbf{}          & Chicago to mark the installation of new Pope Benedict XVI.                        \\
        \textbf{}          & Sentence 2: Pope Benedict XVI is the new leader of the Roman Catholic Church.     \\
                           & Output: 1                                                                         \\ \hline
        \textbf{Input}     & Now complete the following example-                                               \\
        \textbf{}           & Input: Sentence 1: Since 1987, however, Brazil has taken steps to dramatically \\
                           & reduce the destruction, including stepped-up enforcement and the elimination      \\
                           & of tax incentives that led to large-scale land clearing. Sentence 2: In the early \\
                           & 1990s Brazil began to take action to save the rainforest.                         \\
        \textbf{}          & Output: 0                                                                         \\ \hline
        \end{tabular}
    }
    \caption{Example of the category Textual Entailment by showcasing an example from Task 1344 of SuperNI dataset.    }
    \label{tab:textual_entailment}
\end{table*}

\begin{table*}[h!]
    \centering
    \small
    \fontsize{9pt}{\baselineskip}\selectfont % font size
    \renewcommand\tabcolsep{1pt} % column space
    \renewcommand\arraystretch{1}
    \resizebox{\linewidth}{!}
    {
        \begin{tabular}{ll}
\hline
\textbf{Task} &
  \textbf{Answerability Classification} \\ \hline
\textbf{Definition} &
  \begin{tabular}[c]{@{}l@{}}The answer will be 'yes' if the provided sentence contains an explicit mention \\ that answers the given question. Otherwise, the answer should be 'no'. \\ Instances where the answer is implied from the sentence using ""instinct"" \\ or ""common sense"" (as opposed to being written explicitly in the sentence) \\ should be labeled as 'no'.\end{tabular} \\ \hline
\textbf{Example 1} &
  \begin{tabular}[c]{@{}l@{}}Input: Sentence: Jack played basketball for an hour after school, after which \\ he was very tired\\ Question: How long did Jack play basketball?\end{tabular} \\
 &
  Output: Yes \\ \hline
\textbf{Example 2} &
  \begin{tabular}[c]{@{}l@{}}Input: Sentence: He was born in China, so he went to the Embassy at 1 pm to \\ apply for a U.S. Visa.\\ Question: When did he go to Embassy?\end{tabular} \\
 &
  Output: Yes \\ \hline
\textbf{Input} &
  Now complete the following example- \\
\textbf{} &
  \begin{tabular}[c]{@{}l@{}}Input: Sentence: The Vice President's guidance was we need to take them out.\\ Question: Has he always wanted to take them out?\end{tabular} \\
\textbf{} &
  Output: No \\ \hline
\end{tabular}
    }
    \caption{Example of the category Answerability Classification by showcasing an example from Task 020 of SuperNI dataset. }
    \label{tab:answerability_classification}
\end{table*}
\begin{table*}[h!]
    \centering
    \small
    \fontsize{9pt}{\baselineskip}\selectfont % font size
    \renewcommand\tabcolsep{1pt} % column space
    \renewcommand\arraystretch{1}
    \resizebox{\linewidth}{!}
    {
        \begin{tabular}{ll}
\hline
\textbf{Task}      & \textbf{Coreference Resolution}                                                          \\ \hline
\textbf{Definition} &
  \begin{tabular}[c]{@{}l@{}}You need to answer a given question containing a blank (\_). Your answer must be one of the two \\ objects mentioned in the question, for example ""trophy"" and ""suitcase"". Your answer must \\ not contain a word that is not present in the question. Please don't use articles (e.g., the, a) \\ before the answer.\end{tabular} \\ \hline
\textbf{Example 1} & Input: The trophy doesn't fit into the brown suitcase because \_ is too large.           \\
                   & Output: trophy                                                                           \\ \hline
\textbf{Example 2} & Grace was happy to trade me her sweater for my jacket. She thinks \_ looks dowdy on her. \\
                   & Output: sweater                                                                          \\ \hline
\textbf{Input}     & Now complete the following example-                                                      \\
\textbf{} &
  \begin{tabular}[c]{@{}l@{}}Input: The goldfish were finally removed from the bag and transferred into the tank, as \\ the \_ was a temporary home for them.\end{tabular} \\
\textbf{}          & Output: bag                                                                              \\ \hline
\end{tabular}
    }
    \caption{ Example of the category Coreference Resolution by showcasing an example from Task 033 of SuperNI dataset.   }
    \label{tab:coreference_resolution}
\end{table*}

\begin{table*}[h!]
    \centering
    \small
    \fontsize{8pt}{\baselineskip}\selectfont % font size
    \renewcommand\tabcolsep{1pt} % column space
    \renewcommand\arraystretch{1}
    \resizebox{\linewidth}{!}
    {
        \begin{tabular}{ll}
\hline
\textbf{Task}      & \textbf{Data to Text}                                              \\ \hline
\textbf{Definition} &
  \begin{tabular}[c]{@{}l@{}}In this task, you are given concept set (with 3 to 5 concepts) that contain \\ mentions of names of people, places, activities, or things. These concept \\ sets reflect reasonable concept co-occurrences in everyday situations. \\ All concepts given as input are separated by ""\#"". Your job is to generate \\ a sentence describing a day-to-day scene using all concepts from a \\ given concept set.\end{tabular} \\ \hline
\textbf{Example 1} & Input: mountain\#ski\#skier                                              \\
                   & Output: Skier skis down the mountain                                     \\ \hline
\textbf{Example 2} & Input: call\#character\#contain\#wallpaper                               \\
                   & Output: queen of wallpaper containing a portrait called film character. \\ \hline
\textbf{Input}     & Now complete the following example-                                      \\
\textbf{}          & Input: lake\#shore\#walk                                                 \\
\textbf{}          & Output: Men walk along the shore of the lake.                            \\ \hline
\end{tabular}
    }
    \caption{Example of the category Data to text by showcasing an example from Task 102 of SuperNI dataset.}
    \label{tab:data_to_text}
\end{table*}

\begin{table*}[h!]
    \centering
    \small
    \fontsize{9pt}{\baselineskip}\selectfont % font size
    \renewcommand\tabcolsep{1pt} % column space
    \renewcommand\arraystretch{1}
    \resizebox{\linewidth}{!}
    {
        \begin{tabular}{ll}
\hline
\textbf{Task}      & \textbf{Question Rewriting}                                                                                                    \\ \hline
\textbf{Definition} &
  \begin{tabular}[c]{@{}l@{}}Given a disfluent sentence, modify the sentence to it to its equivalent fluent \\ form, preserving the meaning of the sentence.\end{tabular} \\ \hline
\textbf{Example 1} & \begin{tabular}[c]{@{}l@{}}Input: Who did the Han Chinese want to help the Khitan no I mean the \\ Mongols fight?\end{tabular} \\
                   & Output: Who did the Han Chinese want to help the Mongols fight?                                                                \\ \hline
\textbf{Example 2} & \begin{tabular}[c]{@{}l@{}}Input: What part did no I meant how many chapters have coordinating \\ lead authors?\end{tabular}   \\
                   & Output: How many chapters have coordinating lead authors?                                                                      \\ \hline
\textbf{Input}     & Now complete the following example-                                                                                            \\
\textbf{}          & Input: What year did a plague-ridden ship land in Norway?                                                                      \\
\textbf{}          & Output: When did a plague-ridden ship land in Norway?                                                                          \\ \hline
\end{tabular}
    }
    \caption{Example of the category Question Rewriting by showcasing an example from Task 1195 of SuperNI dataset.}
    \label{tab:question_rewriting}
\end{table*}

\begin{table*}[h!]
    \centering
    \small
    \fontsize{9pt}{\baselineskip}\selectfont % font size
    \renewcommand\tabcolsep{1pt} % column space
    \renewcommand\arraystretch{1}
    \resizebox{\linewidth}{!}
    {
        \begin{tabular}{ll}
\hline
\textbf{Task} &
  \textbf{Title Generation} \\ \hline
\textbf{Definition} &
  \begin{tabular}[c]{@{}l@{}}In this task, you're given a paragraph from the research paper and your task is to generate a suitable title\\ for the research paper based on the given paper. Under 100 words is a good title length.\end{tabular} \\ \hline
\textbf{Example 1} &
  \begin{tabular}[c]{@{}l@{}}Input: The severe acute respiratory syndrome (SARS) epidemic originating from China in 2002 was\\ caused by a previously uncharacterized coronavirus that could be identified by specific RT-PCR\\ amplification. Efforts to control future SARS outbreaks depend on the accurate and early identification\\ of SARS-CoV infected patients. A real-time fluorogenic RT-PCR assay based on the 3 -noncoding\\ region (3 -NCR) of SARS-CoV genome was developed as a quantitative SARS diagnostic tool.\\ The ideal amplification efficiency of a sensitive SARS-CoV RT-PCR assay should yield an\\ E value (PCR product concentration increase per amplification cycle) equal to 2.0. It was\\ demonstrated that the 3 -NCR SARS-CoV based RT-PCR reactions could be formulated to reach\\ excellent E values of 1.81, or 91\% amplification efficacy. The SARS-CoV cDNA preparations \\ derived from viral RNA extract and the cloned recombinant plasmid both exhibit the identical \\ amplification characteristics, i.e. amplification efficacy using the same PCR formulation developed \\ in this study. The viral genomic copy (or genomic equivalences, GE) per infectious unit (GE/pfu) \\ of SARS-CoV used in this study was also established to be approximate 1200-1600:1. \\ The assay's detection sensitivity could reach 0.005 pfu or 6-8 GE per assay. It was preliminarily \\ demonstrated that the assay could efficiently detect SARS-CoV from clinical specimens of SARS \\ probable and suspected patients identified in Taiwan. The 3 -NCR based SARS-CoV assay demonstrated \\ 100\% diagnostic specificity testing samples of patients with acute respiratory disease from a \\ non-SARS epidemic region.\end{tabular} \\
 &
  Output: NHS Wales: Court action if trade deals affect service? \\ \hline
\textbf{Example 2} &
  \begin{tabular}[c]{@{}l@{}}Input: By Jon Welch and Paul MoseleyBBC News Details of health problems, family bereavements and \\ personal issues were sent by the University of East Anglia (UEA) in Norwich to 298 students. Megan \\ Baynes, 23, said she felt ""sick and horrified"" when she realised her details had been shared. The UEA \\ apologised ""unreservedly"" and said an inquiry had begun. The email contained a spreadsheet listing \\ 172 names and details extenuating circumstances in which extensions and other academic concessions \\ were granted to 42 students. 'Felt sick' It was sent to nearly 300 undergraduates, including Ms Baynes, \\ a former editor of student newspaper Concrete. She is currently awaiting the results of her American \\ Literature and Creative Writing degree, and had been granted extensions for coursework because of an \\ illness suffered by a family member. ""I felt sick at seeing my personal situation written in a spreadsheet, \\ and then seemingly sent to everyone on my course,"" she said. ""My situation was not the worst on there \\ but there are some on there that are so personal. There are people I know and I feel so awful for them and \\ can't imagine how they are feeling."" Theo Antoniou Phillips, UEA Students' Union undergraduate education \\ officer, said: ""This is a shocking and utterly unacceptable data breach that should never have happened."" \\ Jo Swo, the union's welfare, community and diversity officer, said: ""Given the university is supposed to be \\ making mental health a priority, this is a real slap in the face to students who have sought support."" \\ In a statement, a UEA spokeswoman said: ""An email was mistakenly sent to 298 American Studies \\ undergraduates this morning containing details of 42 students with extenuating circumstances. ""\\ This clearly should not have happened and the university apologises unreservedly. The university has launched \\ an urgent enquiry and is contacting all affected students to offer support. ""Anyone needing support should call \\ 01603 592761. The university is informing the ICO (Information Commissioner's Office)."" \\ The ICO has been contacted for comment.\end{tabular} \\
 &
  Output: University of East Anglia in students' personal data breach \\ \hline
\textbf{Input} &
  Now complete the following example- \\
\textbf{} &
  \begin{tabular}[c]{@{}l@{}}Input: President Donald Trump said Mr Mnuchin had spent his career making money in the private sector and \\ would now work for the taxpayer. Mr Mnuchin pledged to create jobs and combat terrorist financing. \\ Democrats had argued that Mr Mnuchin had made a fortune foreclosing on families during the financial crisis. \\ The top Democrat on the House Financial Services Committee, Maxine Waters of California, called Mr Mnuchin \\ ""the foreclosure king"". His critics have also questioned whether he is too close to the Wall Street banking \\ community, which he will be responsible for regulating. Democrats also complained that Mr Mnuchin had failed \\ to disclose nearly \$100m (£79m) in assets when he filed with the Senate Finance Committee. Mr Mnuchin spent \\ 17 years at Goldman Sachs before becoming a hedge fund manager. He later founded a film production company \\ that was behind such box office hits as the X-Men franchise and American Sniper. Mr Trump said Mr Mnuchin \\ would help make the US a ""jobs magnet"". ""He'll work 24 hours a day, I know him. He'll work 28 hours a day\\  if they give him the extra four hours,"" he said. Another former Goldman executive, Gary Cohn, is the director \\ of President Trump's National Economic Council. What do we know about the new treasury secretary's policy \\ plans? Mr Mnuchin hasn't announced a fully fledged plan, but his responses in media interviews and during the \\ Senate debate over his appointment make clear some of his priorities: There are still many policy areas that have\\  not been addressed, including how he will approach trading relations with China, Mexico and other partners.\end{tabular} \\
\textbf{} &
  Output: Trump says Mnuchin will fight for tax cuts and jobs \\ \hline
\end{tabular}
    }
    \caption{Example of the category Title Generation by showcasing an example from Task 1356 of SuperNI dataset.    }
    \label{tab:title_generation}
\end{table*}

\end{document}